\documentclass[twoside]{article}
\usepackage[utf8]{inputenc}
\usepackage{amsmath}
\usepackage{amsthm}
\usepackage{amsfonts}
\usepackage{amssymb}
\usepackage{float}
\usepackage{graphicx}
\usepackage{subfigure}
\usepackage[left=2cm,right=2cm,top=2cm,bottom=2cm]{geometry}
\usepackage{bm}

\newtheorem{theorem}{Theorem}[section]

\newtheorem{definition}{Definition}[section]

\newtheorem{uda}{Example}[section]
\newtheorem{rem}{Remark}[section]

\def\bhag#1{\noindent
\setcounter{equation}{0}
\section{#1}
}

\def\RR{{\mathbb R}}
\def\CC{{\mathbb C}}
\def\ZZ{{\mathbb Z}}

\def\PPI{{{\rm I}\kern-1pt\Pi}}

\def\TT{\mathbb T}
\def\T{\mathcal{T}}
\def\a{\alpha}
\def\b #1;{{\bf #1}}
\def\x{{\bf x}}

\def\e{\epsilon}

\def\O{{\cal O}}

\def\be{\begin{equation}}
\def\ee{\end{equation}}
\def\bea{\begin{eqnarray}}
\def\eea{\end{eqnarray}}
\def\eref#1{(\ref{#1})}
\def\disp{\displaystyle}

\def\donchitre#1#2{\vskip 6.5cm\noindent
\parbox[t]{1in}{\special{eps:#1.eps x=6.5cm y=5.5cm}}
\hbox to 7cm{}\parbox[t]{0.0cm}{\special{eps:#2.eps x=6.5cm y=5.5cm}}}

\def\dist{\mbox{\textsf{ dist }}}

\title{Theory inspired deep network for instantaneous-frequency extraction and signal components recovery from discrete blind-source data}
\author{Charles K. Chui\thanks{Department of Mathematics, Hong Kong Baptist University. This author is also associated with Department of Statistics, Stanford University, CA 94305, U.S.A. \textsf{email:} ckchui@stanford.edu. He is a Life Fellow of IEEE and his research is partially supported by the Hong Kong Research Council, under Projects $\sharp$ 12300917 and $\sharp$ 12303218, and HKBU Grants $\sharp$ RC-ICRS/16-17/03 and $\sharp$ RC-FNRA-IG/18-19/SCI/01. },
Ningning Han \thanks{Department of Mathematics, Hong Kong Baptist University, Hong Kong. \textsf{email:} ningninghan@hkbu.edu.hk. His research is supported by the Hong Kong Research Council, under Projects $\sharp$ 12300917 and $\sharp$ 12303218.}
 and
H.~N.~Mhaskar\thanks{Institute of Mathematical Sciences, Claremont Graduate University, Claremont, CA 91711, U.S.A. 
\textsf{email:} hrushikesh.mhaskar@cgu.edu }}
\date{}
\begin{document}
\maketitle
\maketitle
 \begin{abstract}
 This paper is concerned with the inverse problem of recovering the unknown signal components, along with extraction of their instantaneous frequencies (IFs), governed by the adaptive harmonic model (AHM), from discrete (and possibly non-uniform) samples of the blind-source composite signal. 
 None of the existing decomposition methods and algorithms, including the most popular empirical mode decomposition (EMD) computational scheme and its current modifications, is capable of solving this inverse problem. 
 In order to meet the AHM formulation and to extract the IFs of the decomposed components, called intrinsic mode functions (IMFs), each IMF of EMD is extended to an analytic function in the upper half of the complex plane via the Hilbert transform, followed by taking the real part of the polar form of the analytic extension. 
 Unfortunately, this approach most often fails to resolve the inverse problem satisfactorily. 
 More recently, to resolve the inverse problem, the notion of synchrosqueezed wavelet transform (SST) was proposed by Daubechies and Maes, and further developed in many other papers, while a more direct method, called signal separation operation (SSO), was proposed and developed in our previous work published in the journal, Applied and Computational Harmonic Analysis, vol. 30(2):243-261, 2016. 
 In the present paper, we propose a synthesis of SSO using a deep neural network, based directly on a discrete sample set, that may be non-uniformly sampled, of the blind-source signal. 
 Our method is localized, as illustrated by a number of numerical examples, including components with different signal arrival and departure times. 
 It also yields short-term prediction of the signal components, along with their IFs.
 Our neural networks are inspired by theory, designed so that they do not require any training in the traditional sense.
 \end{abstract}
\noindent{\textbf{Keywords:} Separation of components, non-stationary signals, deep networks, super-resolution.
\bhag{Introduction and Results}\label{intsect}

Many problems that arise from sensor arrays, stochastic control, mobile communication, and signal processing in general, are modeled by a linear combination of damped sinusoids, with stationary frequencies. 
However, in the current age of big data, particularly in time series analysis, the sinusoids are mostly non-stationary, in that the phase functions are not necessarily linear in time. A general mathematical model for such non-stationary signals or time series may be formulated by 
\be\label{ahm_model}
\sum_{j=1}^{K}A_j(t)\cos\phi_j(t)+A_0(t),
\ee
or more generally by the complex variant 
\be\label{hilbert_model}
f_I(t)=\sum_{j=1}^{K}A_j(t)\exp(i\phi_j(t))+A_0(t),
\ee
where the phase functions $\phi_j(t)$ are differentiable, the amplitude functions $A_j(t)$ are complex-valued and continuous, and $A_0(t)$ is a minimally oscillatory real-valued function, called the trend of $f_I(t)$. 
Here, the subscript $I$ of $f_I(t)$ is used to indicate that noise is not attached to the model. 
This is only a convenience of notation.
Our methods do work well in the presence of noise, as will be proved theoretically in Theorem~\ref{ssotheo} and experimentally in Section~\ref{exptsect}.
Without the trend function $A_0(t)$ in \eref{ahm_model}, the model is called the \emph{amplitude modulation-frequency modulation (AM-FM)} model in signal processing, and the \emph{adaptive harmonic model (AHM)} in the current mathematics literature (see, for example,  \cite{wuthakur, ingrid2011, hautieng_thesis2012, bspaper, cdw, chuilinwu2014}). In our recent paper \cite{chuimhasmaryke16}, the model \eref{hilbert_model} was called the \emph{Hilbert spectrum model (HSM)}. 
The objective of this paper is to introduce and develop an efficient and effective construction of deep networks for solving the inverse problem of recovering the number $K$ of terms in \eref{hilbert_model}, the instantaneous frequencies (IFs) defined by $\phi'_j(t)$, the amplitude functions $A_j(t)$, and the signal components $f_j(t):=A_j(t)\cos\phi_j(t)$, for $j = 1, \cdots, K$, as well as the trend $A_0(t)$, from a sufficiently ``dense'' finite data set $f_I(t_k): k= 0, \cdots, N$, that are allowed to be non-uniformly sampled on a bounded time interval $[b , c]$.
 Of course our constructions and theoretical development apply to infinite non-uniform data samples on an infinite interval as well, simply by ignoring the need of meeting the  requirement of certain boundary conditions.

It is important to point out that our objective of solving an inverse problem is different from that of the popular empirical mode decomposition (EMD) method, introduced in \cite{huang1998empirical}. Indeed, EMD is an ad hoc  computational scheme for the decomposition of a non-stationary signal or time series $g(t)$ into its intrinsic mode functions ``IMFs'' with residue, called the ``trend'', without the concern of recovering the actual IMFs and trend that constitute the source signal or time series $g(t)$. 
To emphasize the difference between EMD and our inverse problem, we recall the pioneering work \cite{prony_original} of Gaspard de Prony in resolving an exponential sum $H(t)$ with $K$ exponents and $K$ corresponding constant coefficients, from $2K$ samples $H(0), H(1), \cdots, H(2K-1)$, where $K$ is known, by first computing the $K$ exponents before computing the $K$ coefficients. 
In particular, if the exponents are $e^{it\omega_j}$, with $\omega_1 < \omega_2 <\cdots < \omega_K$, Prony’s method can be applied to solving the inverse problem \eref{hilbert_model}, at least in theory, for stationary signals without the trend $A_{0}(t)$. 
Unfortunately, Prony’s method fails when some of the frequencies $\omega_j$ are very close to one another. 
The reason is that when these unknown exponents are considered as zeros of a polynomial $p(z)$, with coefficients to be computed from Hankel matrix inversion, computation of these coefficients and the zeros from solving some corresponding eigenvalue problem is highly unstable.
 In any case, the procedure of first recovering the instantaneous frequencies before computing the signal components, as pioneered by Daubechies in \cite{daubechies1996nonlinear, ingrid2011}, and just about all later development in mathematics, including \cite{wuthakur, wu2011one, chuilinwu2014, bspaper, hautieng_thesis2012, cdw, chuimhasmaryke16},  is  opposite to that of EMD, which first computes the signal components, called intrinsic mode functions (IMFs) by repeated applications of the ``sifting process'', before computing the instantaneous frequencies by applying the Hilbert transform to each IMF.

On the other hand, the approach in our present paper is substantially different from those in the published literature, including \cite{hautieng_thesis2012, wuthakur, daubechies1996nonlinear, ingrid2011, wu2011one, chuilinwu2014, bspaper,  cdw, chuimhasmaryke16}. 
First, unlike these papers, we work with non-uniform samples of $f_I$. 
We then approximate $f_I$ using a spline quasi-interpolant \cite{chen1988construction}. 
This quasi-interpolant can be implemented using deep networks as suggested in \cite{multilayer}. 
The \emph{signal separation operator}, (SSO), proposed in \cite{bspaper} can then be evaluated using uniform samples of this quasi-interpolant. 
The SSO operator evaluates a trigonometric polynomial, which can also be synthesized using a further neural network 
as proved in \cite{mhaskar1995degree}, and its thresholding by another layer of a network computing the rectified linear units ReLU activation functions. 
As in \cite{bspaper}, clustering and finding local maxima then lead to the determination of the instantaneous frequencies.
Importantly, in contrast to \cite{ingrid2011}, we obtain the (complex) amplitudes simply by evaluating the SSO at these frequencies.

One interesting aspect of our construction is that the deep networks need not be ``trained'' in the usual sense, but are \emph{theory inspired}, in that their parameters are prescribed by the theory without training.
Our constructions are presented schematically in Figure~\ref{dnnfig} below.

\begin{figure}[ht]
\begin{center}
\includegraphics[width=\textwidth, keepaspectratio]{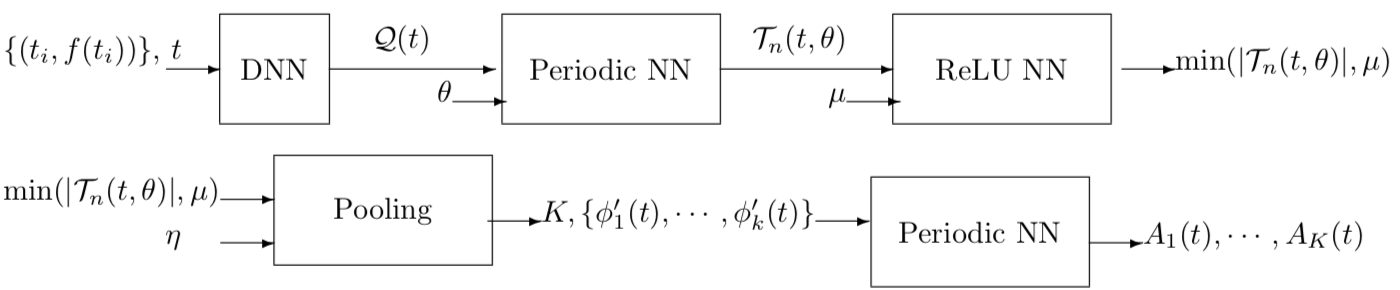} 
\end{center}
\caption{Theory inspired deep network to compute the thresholded values of SSO given non-uniform samples of the signal $f_I$.}
\label{dnnfig}
\end{figure}

After describing our Signal Separation Operator defined in \cite{bspaper} in Section~\ref{ssosect}, we develop the spline approximation in Section~\ref{splinesect}. The actual construction of deep networks as in Figure~\ref{dnnfig} is described in detail in Section~\ref{nnsect}, and illustrated in Section~\ref{exptsect}.

\bhag{Signal separation operator}\label{ssosect}
In this section, we review our construction of the signal separation operator (SSO) and its properties in \cite{bspaper}.
In \cite{bspaper}, trend extraction was done separately; the operator SSO is designed to separate from the Hilbert spectrum model 
\be\label{hilbert_model_bis}
f(t)=\sum_{j=1}^K f_j(t), \qquad f_j(t)=A_j(t)\exp(i\phi_j(t)), \quad j=1,\cdots, K,
\ee
the components $f_k(t)$  and the instantaneous frequencies $\phi_k'(t)$, finding the number $K$ of these quantities in the process.

The following definition summarizes the conditions assumed on the signal.
\begin{definition}\label{cosuniclassdef}
For each $t\in\RR$, let $\mathcal{H}(t)$ denote the collection of functions $f$ of the form \eref{hilbert_model_bis} where each component $f_j(t)=A_j(t)\exp(i\phi_j(t))$, $j=1,\cdots, K$, satisfies each of the following conditions. 
\begin{enumerate}
\item  $A_j :\RR\to \CC$ is continuous, and $\phi_j :\RR\to\RR$ is continuously differentiable. 
\item With
\be\label{cosmaxfreq}
B=B(t):=\max_{1\le j\le K}|\phi_j'(t)|,
\ee
there exists $\a=\a(t)>0$ with the following property:
For any $u$ with $|u| \le \a^{-1}(8\pi B)^{-1/2}$, and $j=1,\cdots, K$,
\be\label{cosakphikcond}
| A_j(t+u)-A_j(t) |\le \a^3|u||A_j(t)|, \qquad | \phi_j'(t+u)-\phi_j'(t)|\le \a^3|u||\phi_j'(t)|.
\ee
\end{enumerate}
\end{definition}

A crucial ingredient in the definition of SSO is the notion of a  lowpass window function defined below.

\begin{definition} \label{low_pass_def} \ A real-valued function $h(u)$, defined for all $u \in \RR$, is said to be an \textbf{admissible window function}, if $0 \le h \in C^3 (\RR)$ is an even function with support  $\mathsf{supp}(h) \subseteq [-1,1]$, such that $h(u_0) > 0$ for some $u_0 \in \mathsf{supp}(h)$. 
\end{definition}
Observe that since $h$ is continuous, $h(u_0) > 0$ implies that $h(u) > 0$ in some neighborhood of $u_0$, so that
\be\label{hbardef}
\hbar_n := \sum_{j \in \ZZ} h \left( \frac{j}{n}\right) > 0
\ee
for all sufficiently large values of $n > 0$.
In the sequel, we assume $h$ to be a fixed low pass window function that is at least $3$ times continuously differentiable.

In the definition of the operator SSO, we allow a perturbation of the original signal. 
Thus, we write
\be\label{noisetosignal}
F(t)=f(t)+\epsilon(t),
\ee
where $f$ is as in \eref{hilbert_model_bis}, and $\epsilon$ is a perturbation.
For each fixed $t\in\RR$, the operator SSO works with equidistant samples $\{F(t-j\delta)\}_{j\in\ZZ : |j|<n}$ for some $\delta>0$. 
In the remainder of this paper, let $\TT$ denote the quotient space of $\RR$ with equivalence relation  $u \approx v$ defined by $(u - v) \in 2 \pi \ZZ$, so that $|u - v| = |(u - v)\; {\rm mod}\; 2\pi|$.

\begin{definition} \label{sso_def} \ (\textbf{Signal separation operator, SSO})
For $\theta \in \TT$ and $u \in \RR$, the signal separation operator $\mathcal{T}_{n,\delta}$, applied to functions $F$ in  \eref{noisetosignal}, is defined by
\be\label{ssodef}
\mathcal{T}_{n,\delta}(F )\,(u,\theta) := \frac{1}{\hbar_n} \sum_{j \in \ZZ}\,h\left(\frac{j}{n}\right)\,e^{ij \theta}\,F(u - j \delta),
\ee
where $h$ is an admissible window function and $\delta, n >0$ are parameters, with $n$  chosen to be an integer so that $\hbar_n$, as defined in  \eref{hbardef}, is positive.
\end{definition}

The statement of our main theorem \cite[Theorem~2.4]{bspaper} given below in Theorem~\ref{ssotheo} below requires some further notation. 
We will consider $t\in\RR$ to be fixed, and denote for brevity
\be\label{cosMdef}
M = M(t) := \sum^K_{j=1} |A_j (t)|, \qquad \mu=\mu(t)=\min_{1\le j\le K}|A_j(t)|>0.
\ee
Also, with $\phi_0'(t):= 0$, we will assume that $\eta:=\eta(t)$ is chosen so that
\be\label{cosminsepcond}
\min_{0\le j\not=\ell \le K}
|\phi_j'(t)-\phi_\ell'(t)|=:2B\eta/\pi >0.
\ee
Necessarily, 
\be\label{etalimit}
0<\eta\le\pi/2.
\ee
We will further use the following abbreviated notation
\be\label{cosshorthand}
f^*_k := A_k (t)\,\exp (i\phi_k (t)), \ \omega^*_k :=  \delta \phi_k' (t),
\ee
to facilitate the statement of the theorem.
Observe that if the parameter $\delta$ of the SSO $\T_{n,\delta}$ is  chosen to satisfy
\be \label{deltacond}
0 < \delta \le \frac{1}{4B},
\ee
where $B = B(t)$ is defined by \eref{cosmaxfreq}, it follows from  \eref{cosmaxfreq} and \eref{cosminsepcond}  that $\omega_k^*\in (0,\pi/2]$ for each $k=1,\cdots, K$, and 
\be\label{cosomegaminsep}
\min_{0 \le k \ne \ell \le K} |\omega^*_k - \omega^*_\ell | = 4 B \eta \delta.
\ee
Here, in view of \eref{etalimit}, the distance $|\omega^*_k - \omega^*_\ell|$ may be interpreted as the distance between points on $\TT$. 

\begin{theorem}\label{ssotheo}\ Let $t \in \RR$ be fixed, and  $F(u) = f(u) + \e(u)$ as defined in  \eref{noisetosignal}, with $f \in {\cal H}(t)$ and 
\be\label{epscond}
|\e(u) | \le E\alpha, \ \ u \in \RR,
\ee
for some constant $E>0$, where $\alpha$ is as in Definition~\ref{cosuniclassdef}. Also, let $n$ be the smallest integer satisfying
\be \label{adef}
n\ge \left(\alpha \delta \sqrt{8 \pi B} \right)^{-1}.
\ee
Then the following statements hold for all    sufficiently small $\a>0$. 
\begin{enumerate}
\item[(a)] The set $\{u\in [0,\pi] : |\T_{n,\delta}(F)(t,u)| \ge \mu/2\}$ can be expressed as a disjoint union of exactly $K$ non--empty sets $\mathcal{G}_\ell$, $\ell=1,\cdots,K$, where $K$ is the number of signal components $f_1, \cdots, f_K$ of $f$ in \eref{ahm_model},  with the following properties:
\begin{enumerate}
\item[(i)] Each $\mathcal{G}_\ell$ contains  a unique $\omega_\ell^*$.
\item[(ii)] 
\be\label{gelldiam}
\mathsf{diam }(\mathcal{G}_\ell) \le B\eta\delta, \qquad 1\le \ell\le K.
\ee
\item[(iii)] 
\be\label{gellsep}
\dist(\mathcal{G}_\ell, \mathcal{G}_j)\ge B\eta\delta, \qquad 1\le \ell\not=j\le K.
\ee
\end{enumerate} 
\item[(b)] There exists $\gamma>0$,  such that if
\be\label{hatomegadef}
\widehat{\omega}_\ell=\arg\max_{\theta\in\mathcal{G}_\ell}|\T_{n,\delta}(F)(t,\theta)|, \qquad \ell=1,\cdots, K,  
\ee
then
\be\label{freqfound}
|\widehat{\omega}_\ell-\omega_\ell^*| \le \gamma\alpha\delta,
\ee
\item[(c)] Let   the kernel
\be\label{phi_kern_def}
\Phi_n(u):=\sum_{k\in\ZZ}h\left(\frac{k}{n}\right)e^{iku}, \qquad u\in\TT,
\ee
 be non--negative valued function.  Then for sufficiently small $\alpha$,
\be\label{ampfound}
\left|2|\T_{n,\delta}(F)(t,\widehat \omega_\ell)| - |A_\ell(t)|\right|\le 2(2M+E)\a,
\ee
and
\be\label{compfound}
|\T_{n,\delta}(F)(t,\widehat{\omega}_\ell)-f_\ell^*|\le 2\left(1+\frac{5M}{\mu}\right)(2M+E)\a, \qquad \ell=1,\cdots, K.
\ee
\end{enumerate}
\end{theorem}

\bhag{Piecewise polynomial approximation}\label{splinesect}
We note that the application of SSO requires equidistant samples around each $t\in\RR$. 
When the signal is given only as a set of samples at non-uniform nodes, our idea is to use a spline function to approximate the signal, and then use this approximation in place of $F$ in the application of SSO. The purpose of this section is to describe this construction.

In this section, we fix an integer $m\ge 1$, and  assume that the values $\{F(t_j)\}$ are known for a set of points $a=t_0<t_1<\cdots<t_{Mm+r}=b$. 
We write $I_j=[t_{jm+1}, t_{(j+1)m}]$, $j=0,\cdots, M-2$, $I_{M-1}=[t_{(M-1)m}, t_{Mm+r-1}]$, and 
\be\label{meshnorm}
\Delta=\max_{0\le k\le Mm+r-2}(t_{k+1}-t_k).
\ee
We denote the class of all algebraic polynomials of degree $< m$ by $\Pi_m$.
For any $g :[a,b]\to\RR$, we now define an approximation operator.
For $j=0,\cdots, M-1$, we define the polynomial $R_j(g)\in \Pi_m$ to be the unique polynomial that interpolates $g$ at $t_{jm+1},\cdots, t_{(j+1)m}$; i.e., $R_j(g)(t_k)=g(t_k)$ for $k=jm+1,\cdots, t_{(j+1)m}$.

The approximation to $g$ is then defined by
\be\label{ppapprox}
Q(g)(x)= R_j(g)(x), \quad x\in \begin{cases}
[t_0,t_{m+1}], &\mbox{ if $j=0$,}\\
(t_{jm+1}, t_{(j+1)m}], & \mbox{ if $j=1,\cdots, M-2$,}\\
(t_{(M-1)m}, t_{Mm+r-1}], & \mbox{if $j=M-1$,}\\
0, &\mbox{ if $x\not\in [a,b]$.}
\end{cases}
\ee

\noindent\textbf{Constatnt convention}\\
\emph{In the sequel, the symbols $c, c_1, \cdots$ will denote generic positive constants depending only the fixed parameters in the discussion, such as $m$, $h$, etc. Their values may be different at different occurrences, even within a single formula. 
The notation $A\sim B$ means $c_1A\le B\le c_2A$.}\\

Obviously,
\be\label{fepsdecomposition}
Q(F)=\sum_{j=1}^K Q(f_j) +Q(\epsilon).
\ee
It is clear that
\be\label{epsbd}
|Q(\epsilon)(x)|\le c\max_{t\in [a,b]}|\epsilon(t)|.
\ee
If each $f_j$, and hence, $f$ is $m$-times continuously differentiable on $[a,b]$, then it is well known that
\be\label{festimate}
|f(x)-Q(f)(x)|\le c\Delta^m\max_{t\in [a,b]}|f^{(m)}(t)|.
\ee

\begin{theorem}\label{perturb_sso_theo}
Let $f$ be $m$ times continuously differentiable on $[a,b]$, $f$ be supported on $[a,b]$ (in particular, $f^{(\ell)}(a)=f^{(\ell)}(b)=0$ for $\ell=0,\cdots, m-1$), and \eref{epscond} be replaced by
\be\label{bisepscond}
c\Delta^m\max_{t\in [a,b]}|f^{(m)}(t)|+|\e(u) | \le E\alpha, \ \ u \in \RR.
\ee
Then the conclusions of Theorem~\ref{ssotheo} hold if $\T_{n,\delta}(F)(t,u)$ is replaced by $\T_{n,\delta}(Q(F))(t,u)$.
\end{theorem}

\begin{rem}\label{trendrmk}
{\rm 
It is clear that there exists a polynomial $P\in \Pi_{2m}$ such that $f^{(\ell)}(a)=P^{(\ell)}(a)$ and $f^{(\ell)}(b)=P^{(\ell)}(b)$ for $\ell=0,\cdots,m-1$. Therefore, the assumption in Theorem~\ref{perturb_sso_theo} holds without the apparently extra condition about $f$ being supported on $[a,b]$. 
In this case, we need to deal with the function $F-P$ instead of $F$ . 
In \cite{bspaper}, we have derived several algorithms for removing the ``polynomial trend'' without any knowledge of the polynomial in advance.
\qed}
\end{rem}

\begin{rem}\label{quasiintrem}
{\rm
Although is stated with the piecewise polynomial interpolant as defined above, for the purposes of actual computation, there are many other methods available in the spline literature, which we use for the actual computations in Section~\ref{exptsect}.
In particular, spline quasi-interpolation, introduced in \cite{de1973spline} and discussed in some details in (\cite[pages 178 and 194]{deboorbk}), is attractive, since it is a local computational scheme that assures optimal order of approximation.  
To avoid using derivative data required in \cite{deboorbk}, the spline quasi-interpolation scheme constructed in \cite{chen1988construction} can be applied for real-time applications by using only data samples. 
On the other hand, if the data samples are valuable for certain applications, the quasi-interpolation scheme must be modified to possess both the interpolation and  quasi-interpolation properties.  
A “prediction-correction” formulation of such schemes,  called “local blending spline interpolation,” was introduced in \cite{chuidiamond90}. In combination with the real-time spline quasi-interpolation scheme in \cite{chen1988construction}, the local blending spline interpolation scheme for non-uniform knots is developed in \cite{chuilinwu2014} and adopted to satisfy the derivative boundary conditions in \cite{maryke_thesis2015, cdw}, by considering the knot sequence:
\be \label{knot}
\x: b= x_{-m+1} = \cdots = x_{-1} = x_0 < x_1 < \cdots< x_{2N} = \dots  = x_{2N+m-1} =c.
\ee
For example, with $m=4$ stacked knots at the boundary to interpolate the first and second divided differences at the end points $b$ and $c$ for cubic spline interpolation at $x_{2k} = t_k$ for $k=0,\dots, N$, where $x_{2k+1}$ may be chosen by taking the average of $x_{2k}$ and $x_{2k+2}$ for $k=0,\dots, N-1$. 
We remark that interpolation of  derivatives or divided differences at the boundary not only minimizes boundary artifact, but also allows the ``extrapolation'' capability of our computational scheme for each IMF and the trend, at least for $t<b-d$ and $t>c+d$ for some positive value $d$, with the larger extrapolation interval $[b-d,b]$ and $[c,c+d]$ by using higher $m$-th order B-splines to interpolate up to the $(m-2)$-th order derivatives of divided differences at $b$ and $c$ by any even $m>4$. 
To end this section, we point out that if the data ${F(t_k)}$ are available for equally spaced time samples, at $t_k = b + k\delta$, then local blending spline interpolation can be computed in real-time, simply by up-sampling, followed by moving averaging with weights derived in (\cite[pp.115-117]{chuiwaveletbk}). (See also \cite{chui1988multivariate} for the bivariate setting in terms of box splines).
}
\end{rem}

\bhag{Deep networks}\label{nnsect}
In this section, we describe the implementation of our method in terms of deep neural networks.

\subsection{Piecewise polynomials as deep networks}\label{splineddnnsect}
The first step in our algorithm sketched in Figure~\ref{dnnfig} is to implement the quasi-interpolant as a deep network. 
We observe that a piecewise polynomial $Q$ with knots $\{t_k\}$ is a linear combination of the form
\be\label{spline_pp}
Q(t)=\sum_k a_k(t-t_k)_+^m.
\ee

Next, let $J$ be an integer such that $2^J\ge m$, and $\sigma_2(t)=(t_+)^2$. 
The function $t\mapsto (t_+)^{2^J}$ can be implemented exactly as $((\cdots((t_+^2)^2)^2\cdots)^2$; i.e., as a neural network with $J$ layers with one neuron each, evaluating the activation function $\sigma_2$.
We denote this network by $\mathcal{P}_J$.
Finally, since the function $t\mapsto t^m$ can be expressed as a linear combination of the functions $t\mapsto (t-\ell)^{2^J}$ by taking the divided difference,  the function $t\mapsto (t_+)^m$ can be implemented as the same linear combination of the networks $t\mapsto \mathcal{P}_J(t-\ell)$.
A linear combination of these in turn implement $Q$ exactly.
See \cite{multilayer} for further details.

\subsection{Trigonometric polynomials as neural networks}\label{trigasnnsect}
The next step is to implement the function $\mathcal{T}_{n,\delta}(Q(F))$ as a neural network.
In some sense, since $\mathcal{T}_{n,\delta}(Q(F))$ is a trigonometric polynomial of degree $<n$, it is already a neural network with the activation function $t\mapsto\cos t$.
We have demonstrated in \cite{mhaskar1995degree} how any trigonometric polynomial can be implemented approximately using other sufficiently smooth activation functions.
For example, we consider  the smooth ReLU function $t\mapsto \log(1+e^t)=t_++\O(e^{-|t|})$. Then the function
$\disp\psi(t)=\log\left(\frac{(1+e^{t+\pi})(1+e^{t-\pi})}{(1+e^t)^2}\right)$
is integrable on $\RR$. 
The periodization 
\be\label{smoothreludef}
\phi(t)=\sum_{j\in\ZZ}\psi(t+2\pi j), \qquad t\in\RR,
\ee
is an analytic function on $\TT$.
With $N\ge 1$, we construct the network
\be\label{ssonet}
\mathbb{G}_{N,n,\delta}(F)(u,\theta)=\frac{1}{(2N+1)\hat{\phi}(1)\hbar_n}\sum_{k=0}^{2N}\exp\left(\frac{2\pi i k}{2N+1}\right)\left(\sum_{j\in\ZZ}
h\left(\frac{j}{n}\right)\,e^{ij \theta}\,F(u - j \delta) \phi\left(j\cdot(\circ)-\frac{2\pi k}{2N+1}\right)\right).
\ee
Then our results in \cite{mhaskar1995degree} imply that
\be\label{ssoapprox}
\max_{\theta\in\TT, u\in\RR}\left|\mathcal{T}_{n,\delta}(Q(F))(u,\theta)-\mathbb{G}_{N,n,\delta}(F)(u,\theta)\right|
\le cM\rho^N
\ee
for some constant $c>0$ and $\rho\in (0,1)$.

\subsection{Thresholding as a deep network}\label{thresholdsect}
The next step is to implement $\min(|\mathcal{T}_n(t,\theta)|,\mu)$. 
This is easy using ReLU networks; i.e., a network that uses the activation function $\sigma_1(t)=t_+$, equivalently, $t\mapsto |t|=t_++(-t)_+$.  
Indeed, for any real numbers $a, b$,
$$
\max(a,b)=(1/2)\left((a-b)_+ - (b-a)_+ +(a+b)_+ +(-a-b)_+\right),
$$
and
$$
\min(a,b)=(a+b)_++(-a-b)_+-\max(a,b).
$$
Thus, $\min(|\mathcal{T}_n(t,\theta)|,\mu)$ can be implemented as network with two layers, receiving the inputs $\mathcal{T}_n(t,\theta)$ and $\mu$ as shown in Figure~\ref{dnnfig}.

\section{Experimentation and Examples}\label{exptsect}

In this section, we demonstrate the effectiveness of the theory introduced in this paper,  by solving the inverse problem of recovering all instantaneous frequencies (IFs), amplitude functions, and IMFs, as well as the trend from the blind source signal $f_{I}$, plus an additive noise.
We will implement our deep networks as spline quasi-interpolants as described in Remark~\ref{quasiintrem}, followed by the SSO, thresholding, and the evaluation of another SSO as indicated in Figure~\ref{dnnfig}.
In each synthetic experiment,we consider white noise with zero mean and variance $\sigma^{2}$, and the Signal-to-Noise Ratio (SNR), is defined by
$$
{\rm SNR}[{\rm dB}]=10\log_{10}\frac{{\rm var}(f_{I})}{\sigma^{2}}.
$$
The mean square error (MSE) is used as a performance measure, defined by
$$
{\rm MSE}=\|f-\widetilde{f}\|_{2}^{2}/\|f\|_{2}^{2},
$$
where $f$ is the original signal and $\widetilde{f}$ denotes the recovered signal. Since the choice of the non-uniform samples as well as the noise is random, we repeat each experiment $50$ times. 
The accuracy of the reconstructed results of IFs and IMFs as well as the trend is measured by means of the normalized mean square error (NMSE), calculated by averaging MSE of 50 independent trials, and the standard deviation (STD).

\begin{uda}\label{uda:closefreq}
{\rm \textbf{(Close-by IFs)}
 The first example is a signal consisting of four IMFs with very close-by frequencies and a non-monotone trend, given by
\begin{equation}\label{experiments1}
\begin{aligned}
f_{I,1}(t)=f_{1,1}(t)+f_{1,2}(t)+f_{1,3}(t)+f_{1,4}(t)+A_{1,0}(t), 0<t<200,\\
\end{aligned}
\end{equation}
where
\begin{equation}\label{expt1components}
\begin{aligned}
&f_{1,1}(t)=(t/20 + 30)\cos(2\pi(0.95t)),\\
&f_{1,2}(t)=32\cos(2\pi(0.99t)),\\
&f_{1,3}(t)=40\exp(-(t/141-7/10)^{2})\cos(2\pi t),\\
&f_{1,4}(t)=(t^{2}/4000 +t/40 + 25)\cos(2\pi1.02t),\\
&A_{1,0}(t)= (2/625)(3t/20-16)^{4}-(32/45)(3t/20-10)^{2} + 3t/2+40.\\
\end{aligned}
\end{equation}
 In Figure \ref{fig:closebyfreq}, we plot the original signal $f_{I,1}(t)$ and the observed signal (SNR$=20$dB), together with the reconstructed results of IMFs ($0<t<50$) and trend ($0<t<200$). 
 \begin{figure}[ht]
\centering
\begin{tabular}{ccc}
\includegraphics[width=6cm]{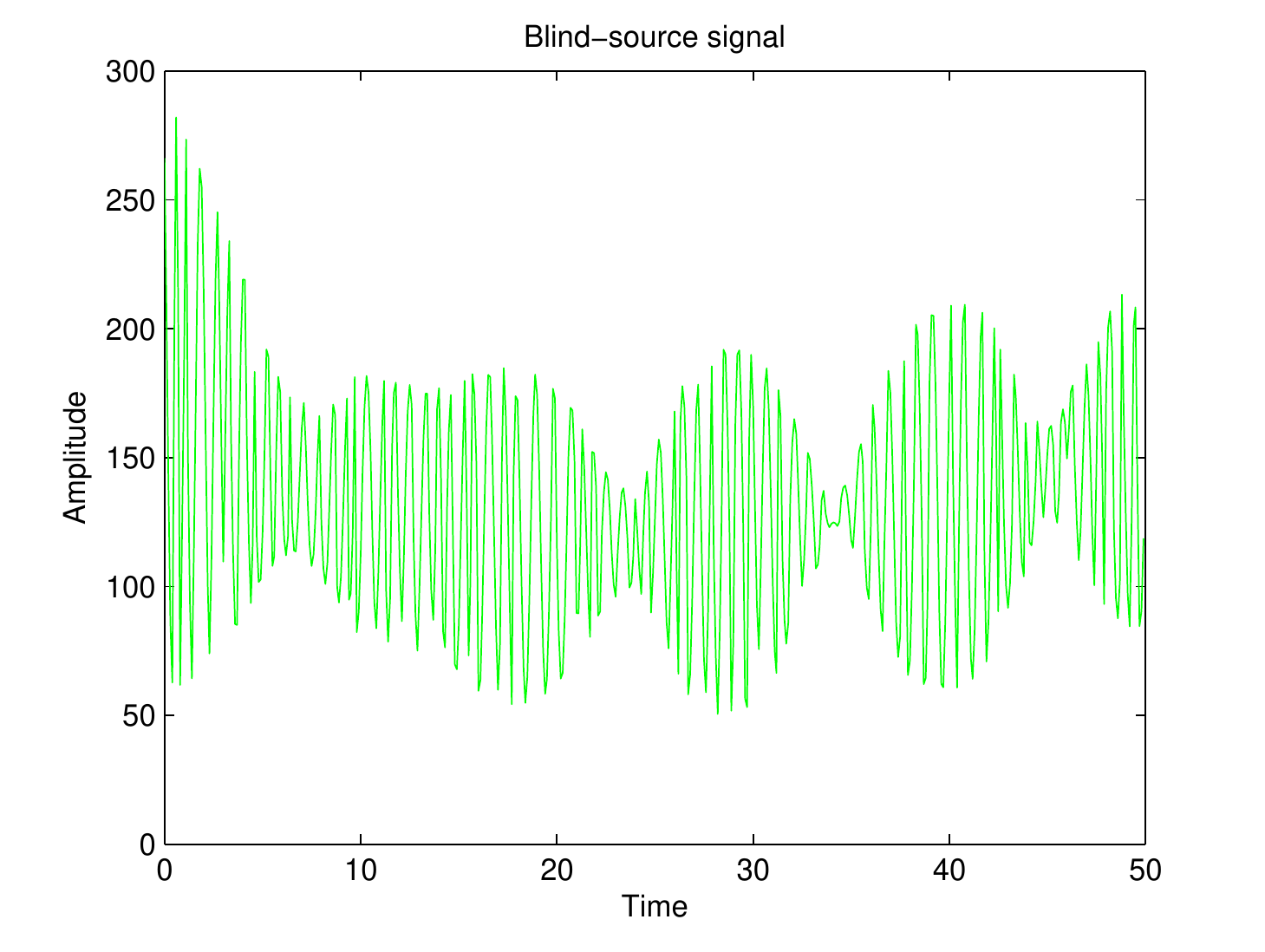}
 & \includegraphics[width=6cm]{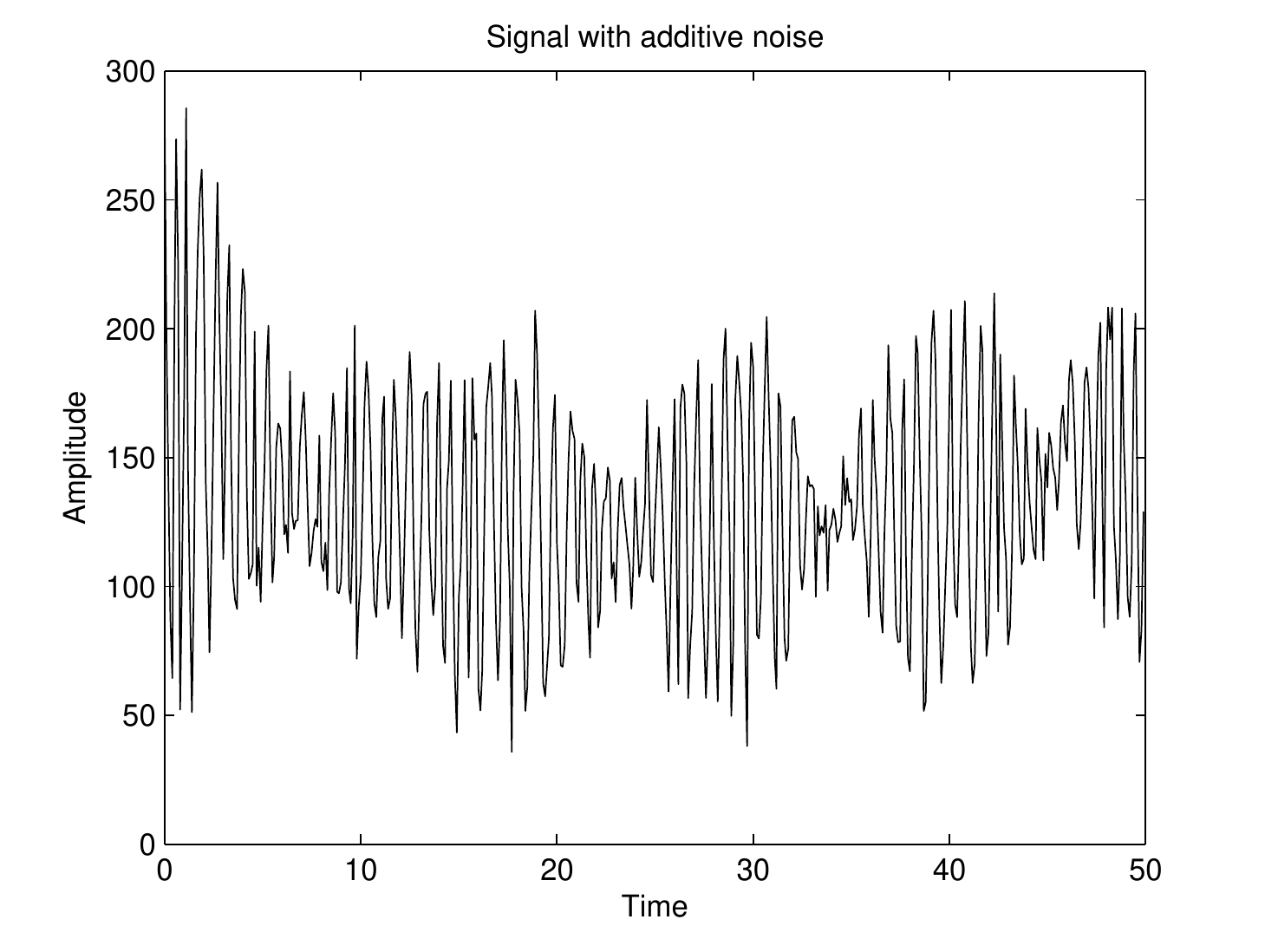}
 & \includegraphics[width=6cm]{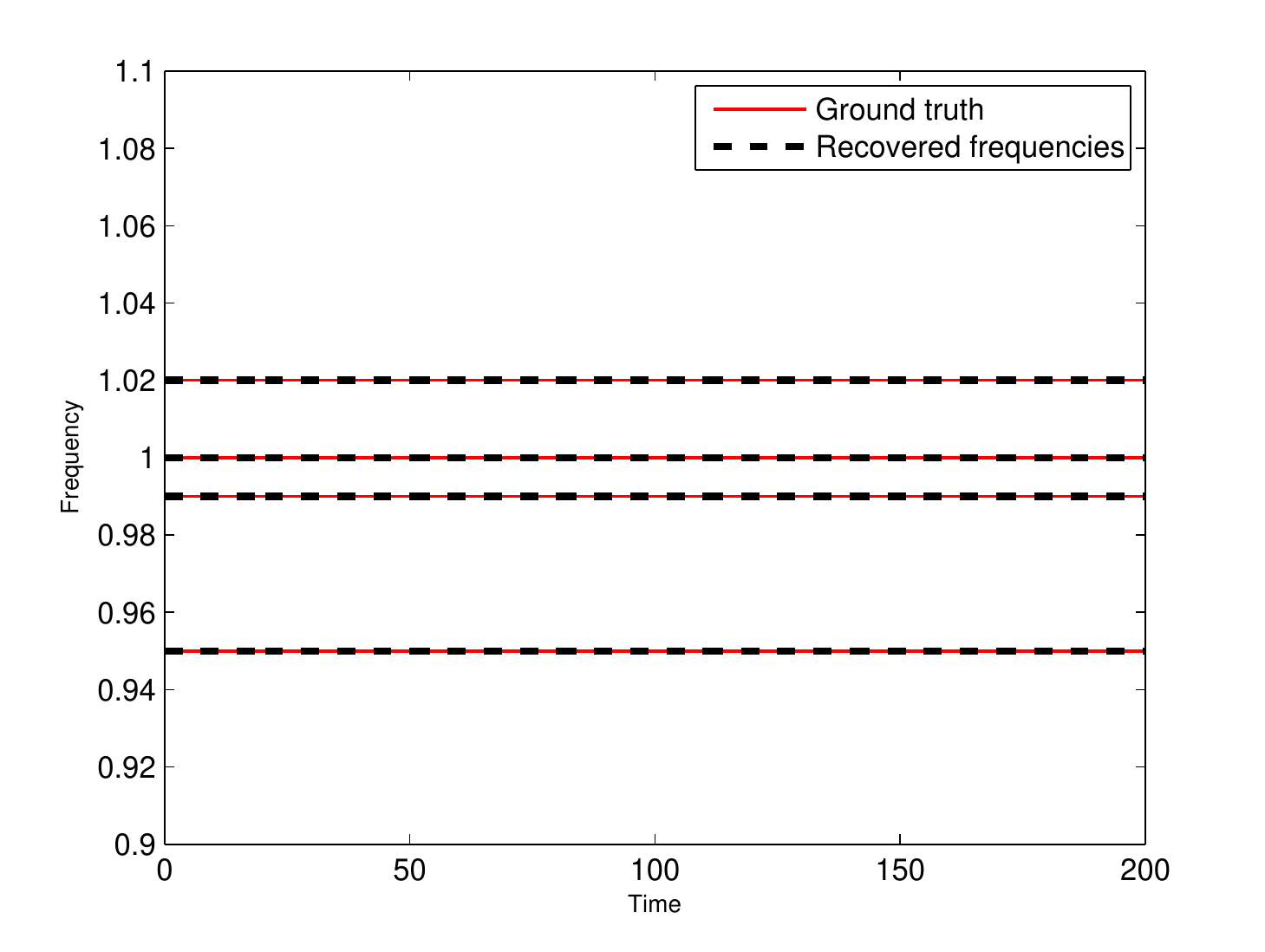}\\
  \includegraphics[width=6cm]{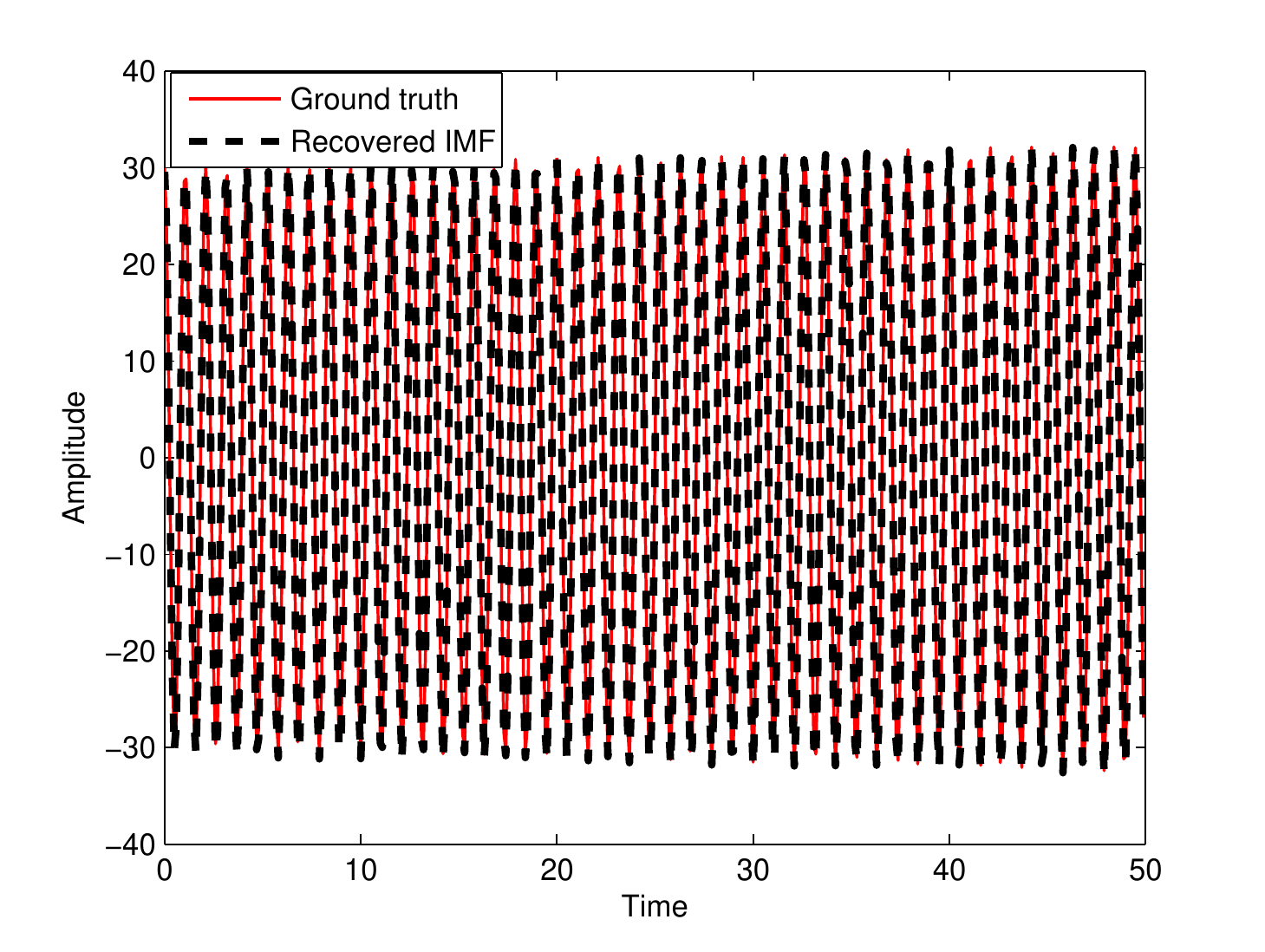}
&\includegraphics[width=6cm]{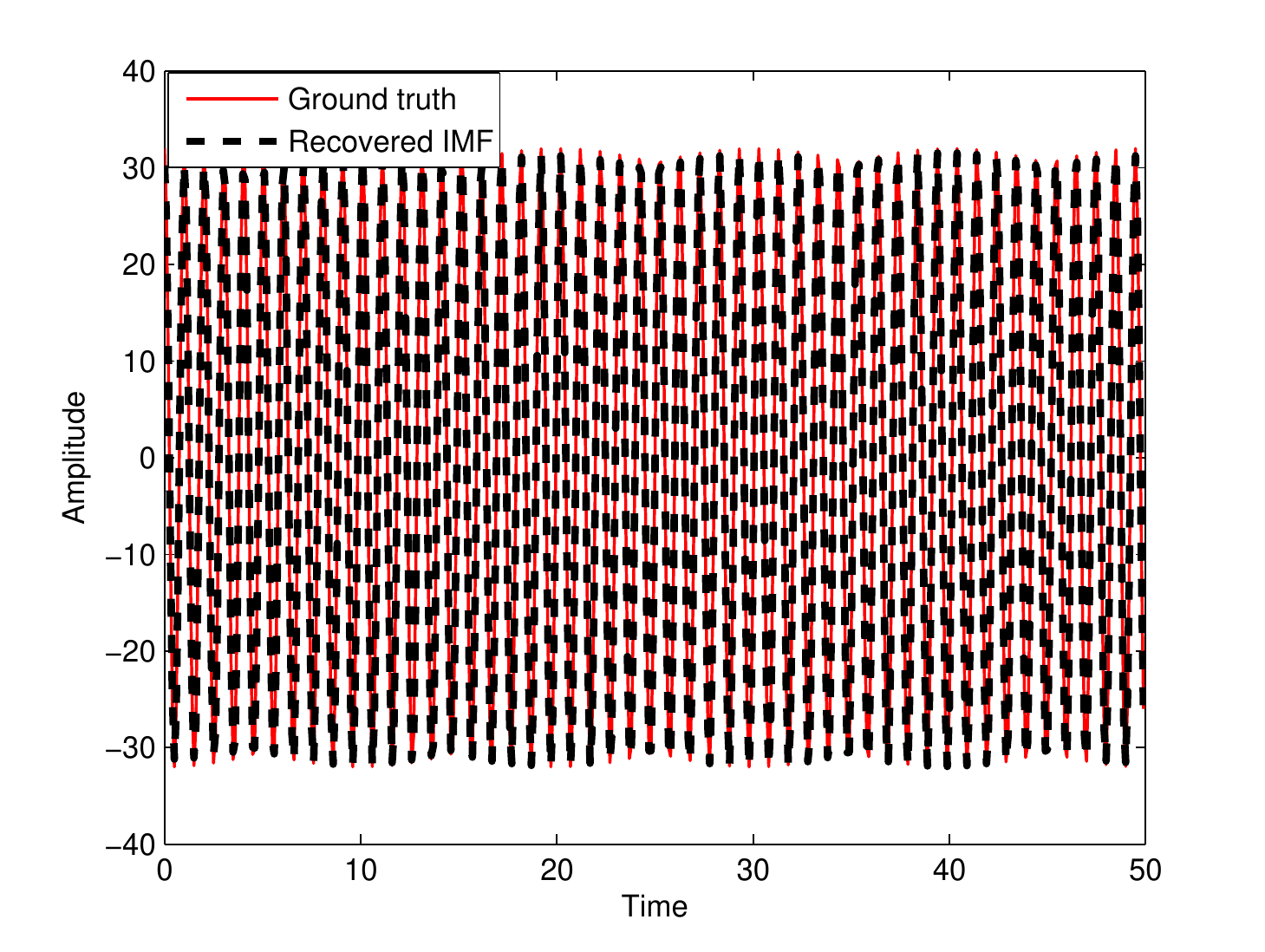}
 &\includegraphics[width=6cm]{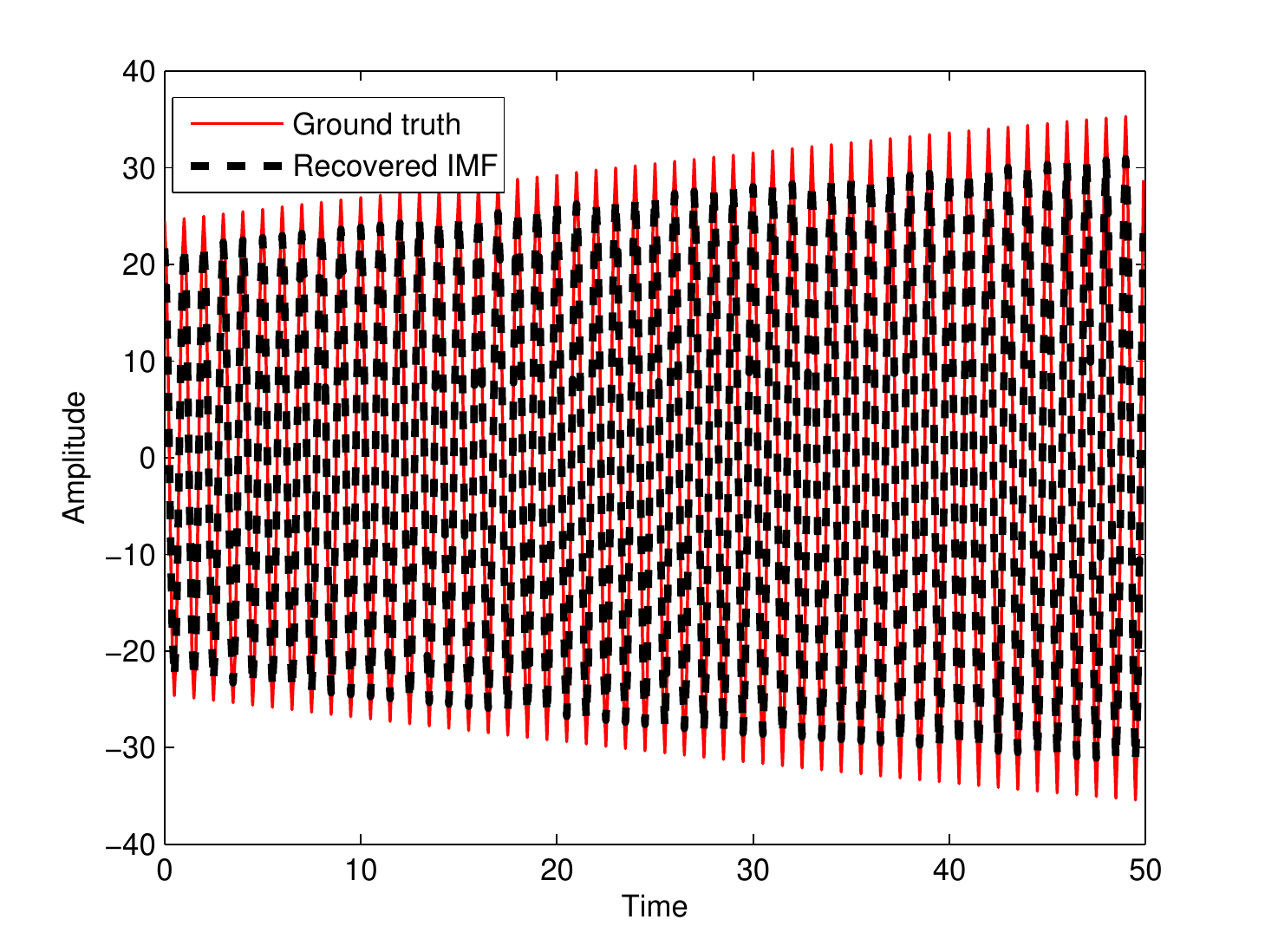}\\
  \includegraphics[width=6cm]{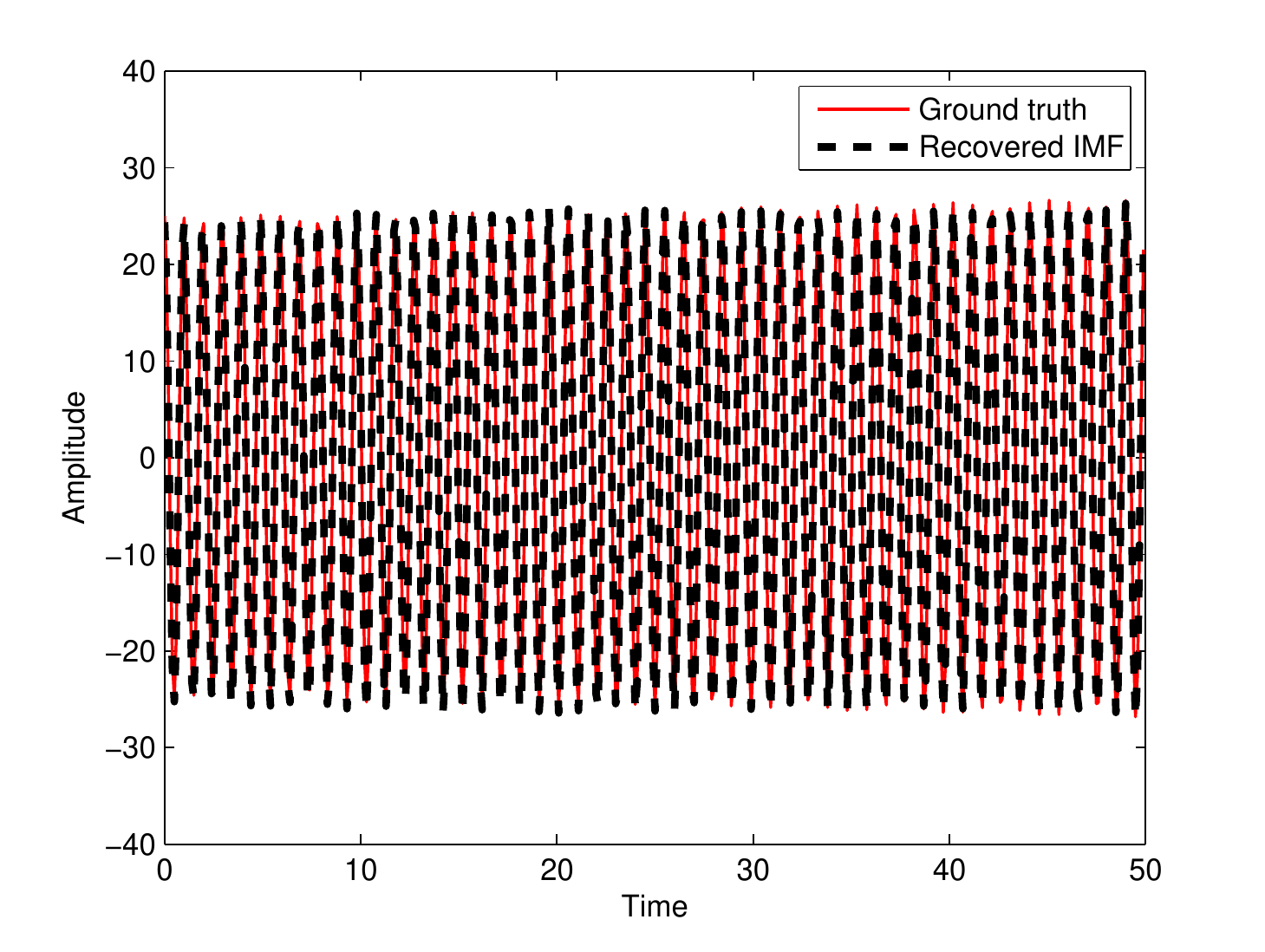}
  &\includegraphics[width=6cm]{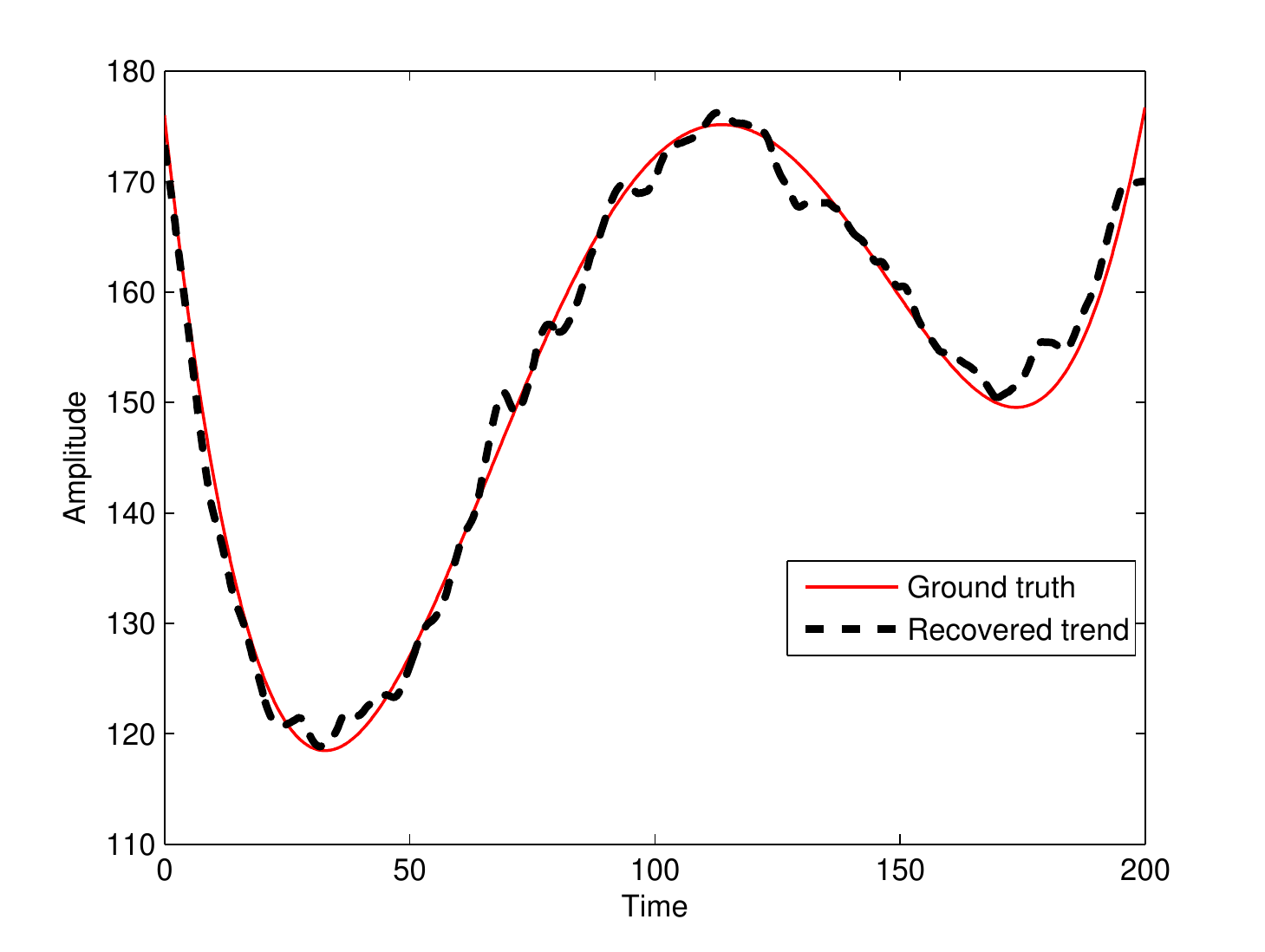}\\
\end{tabular}
\caption{Top (left to right): blind-source signal $f_{I,1}(t)$, $f_{I,1}(t)$ with additive noise, recovered frequencies. Middle (left to right): recovered $1$st, $2$nd, $3$rd IMFs. Bottom (left to right): recovered $4$th IMF, recovered trend.}
\label{fig:closebyfreq}
\end{figure}
The numerical errors in the recovery are listed in Table~\ref{table1}.
\begin{table}
\centering
\scriptsize
\caption{Reconstructed results of Examples $1\sim3$}
\label{table1}       
\begin{tabular}{|l|l|l|l|}
\hline
& Size of non-uniform samples (test data) & NMSE (STD) of reconstructed IFs &NMSE (STD) of reconstructed IMFs\\
\hline
$f_{1,1}(t)$ & ~~~~~~~~~~~~~~$1334$~~$(2000)$  & ~~~~~~$1.37\times10^{-6}$~~$(2.84\times10^{-8})$& ~~~~~~$1.90\times10^{-3}$~~$(6.77\times10^{-4})$\\
$f_{1,2}(t)$ & ~~~~~~~~~~~~~~$1334$~~$(2000)$  & ~~~~~~$1.26\times10^{-6}$~~$(1.10\times10^{-8})$ &~~~~~~$2.60\times10^{-3}$~~$(8.70\times10^{-4})$\\
$f_{1,3}(t)$ & ~~~~~~~~~~~~~~$1334$~~$(2000)$  &  ~~~~~~$7.44\times10^{-6}$~~$(3.56\times10^{-8})$ &~~~~~~$1.71\times10^{-2}$~~$(2.30\times10^{-3})$\\
$f_{1,4}(t)$ & ~~~~~~~~~~~~~~$1334$~~$(2000)$  & ~~~~~~$2.59\times10^{-6}$~~$(1.97\times10^{-8})$  & ~~~~~~$1.50\times10^{-3}$~~$(5.44\times10^{-4})$\\
$A_{1,0}(t)$ & ~~~~~~~~~~~~~~$1334$~~$(2000)$  & ~~~~~~~~~~~~~~~~~~~~$\sim$~~~~~~~& ~~~~~~$2.08\times10^{-4}$~~$(1.51\times10^{-4})$\\
\hline
$f_{2,1}(t)$ & ~~~~~~~~~~~~~~$1335$~~$(2002)$  & ~~~~~~$4.90\times10^{-4}$~~$(1.35\times10^{-4})$& ~~~~~~$3.44\times10^{-2}$~~$(8.10\times10^{-3})$\\
$f_{2,2}(t)$ & ~~~~~~~~~~~~~~$1335$~~$(2002)$  & ~~~~~~$6.63\times10^{-4}$~~$(1.71\times10^{-4})$ &~~~~~~$6.92\times10^{-2}$~~$(6.70\times10^{-3})$\\
$A_{2,0}(t)$ & ~~~~~~~~~~~~~~$1335$~~$(2002)$  & ~~~~~~~~~~~~~~~~~~~~$\sim$~~~~~~~& ~~~~~~$6.30\times10^{-4}$~~$(2.57\times10^{-4})$\\
\hline
$f_{3,1}(t)$ & ~~~~~~~~~~~~~~$750$~~~$(1000)$  & ~~~~~~$7.22\times10^{-4}$~~$(7.46\times10^{-5})$& ~~~~~~$3.14\times10^{-2}$~~$(1.80\times10^{-3})$\\
$f_{3,2}(t)$ & ~~~~~~~~~~~~~~$750$~~~$(1000)$  & ~~~~~~$1.97\times10^{-5}$~~$(1.43\times10^{-6})$ &~~~~~~$2.30\times10^{-2}$~~$(3.10\times10^{-3})$\\
$f_{3,3}(t)$ & ~~~~~~~~~~~~~~$750$~~~$(1000)$  & ~~~~~~$2.98\times10^{-5}$~~$(1.80\times10^{-6})$ & ~~~~~~$2.79\times10^{-2}$~~$(1.38\times10^{-3})$\\
$f_{3,4}(t)$ & ~~~~~~~~~~~~~~$750$~~~$(1000)$  & ~~~~~~$7.57\times10^{-5}$~~$(3.56\times10^{-4})$ & ~~~~~~$4.19\times10^{-2}$~~$(6.20\times10^{-3})$\\
$A_{3,0}(t)$ & ~~~~~~~~~~~~~~$750$~~~$(1000)$  & ~~~~~~~~~~~~~~~~~~~~$\sim$~~~~~~~& ~~~~~~$4.54\times10^{-5}$~~$(8.96\times10^{-6})$\\
\hline
\end{tabular}
\end{table}
\qed}
\end{uda}

\begin{uda}\label{uda:splitsignal}
{\rm
\textbf{(Split signal)} Next, we consider a challenging example introduced in \cite{ingrid2011}, defined by
\begin{equation}\label{splitsignal}
\begin{aligned}
f_{I,2}(t)=f_{2,1}(t)+f_{2,2}(t)+A_{2,0}(t),
\end{aligned}
\end{equation}
where
\begin{equation}\label{splitsignalcomp}
\begin{aligned}
&f_{2,1}(t)=\cos(2\pi(10t/\pi)),~~0< t <5\pi/2,\\
&f_{2,2}(t)=\cos(2\pi((2/3\pi)((t-10)^{3}-(2\pi-10)^{3}) + (5/\pi)(t-2\pi))),~~ 2\pi<t <4\pi,\\
&A_{2,0}(t)= t/2, ~~0< t<5\pi/2,\\
\end{aligned}
\end{equation}
 and add a noise with SNR$=10$dB to this signal. 
 This signal does not satisfy the conditions of Theorem~\ref{ssotheo}.
Nevertheless, the quasi-interpolatory spline approximation being highly localized, our techniques are able to recover the IFs and IMFs accurately as indicated in Figure~\ref{fig:splitsignal} and reported on in further detail in Table~\ref{table1}. 

\begin{figure}[ht]
\centering
\begin{tabular}{ccc}
\includegraphics[width=6cm]{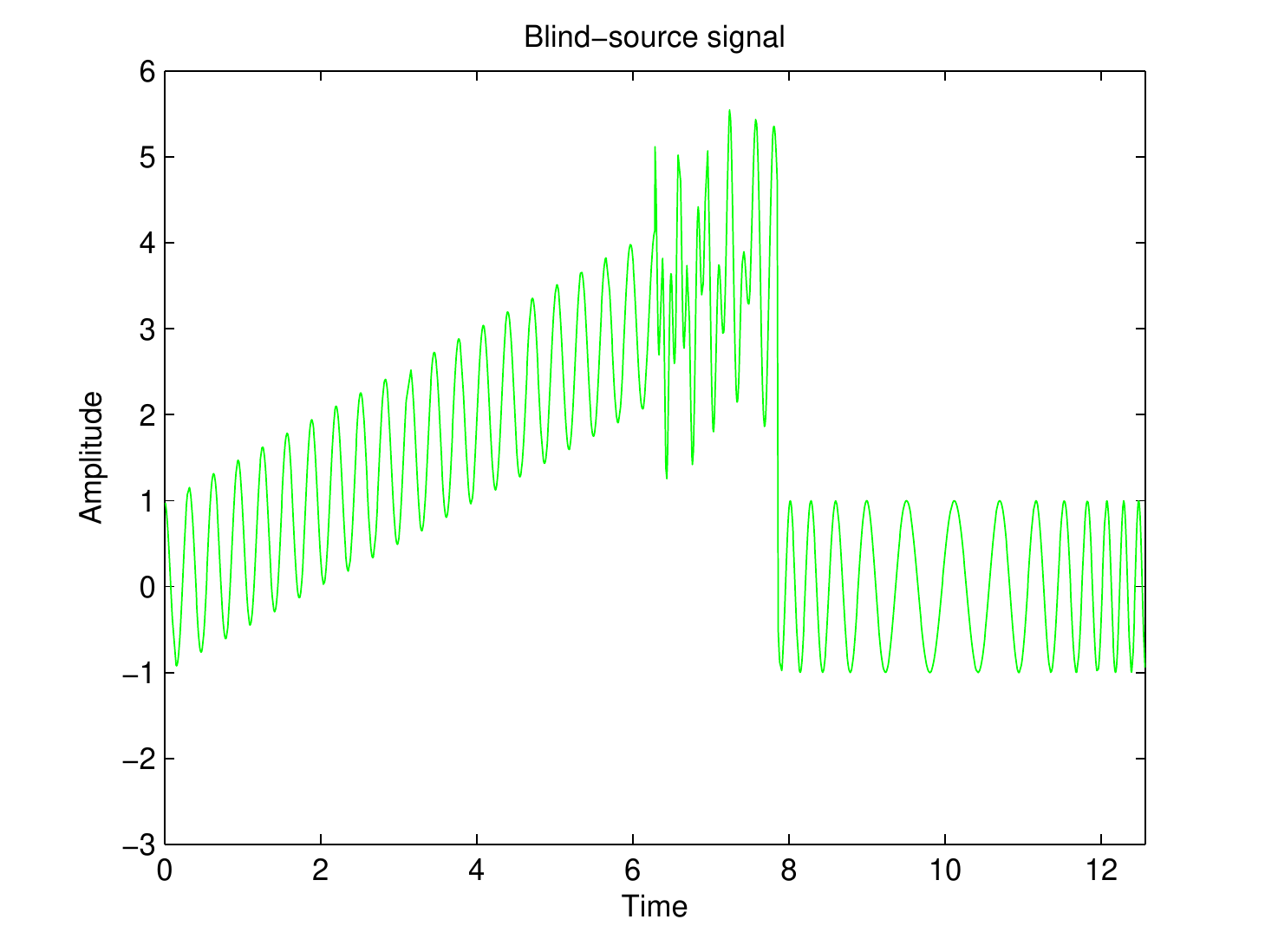}
 & \includegraphics[width=6cm]{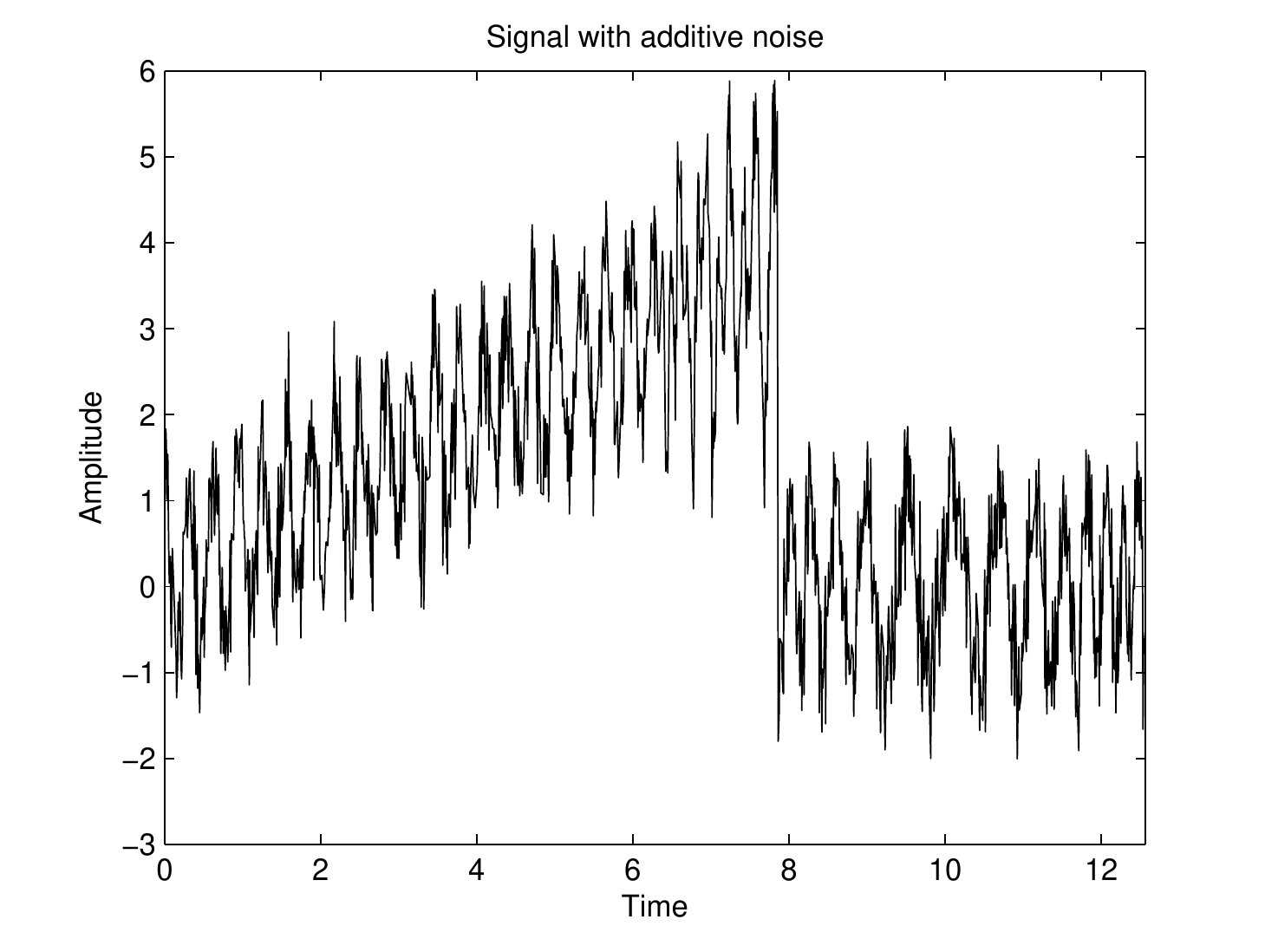}
 & \includegraphics[width=6cm]{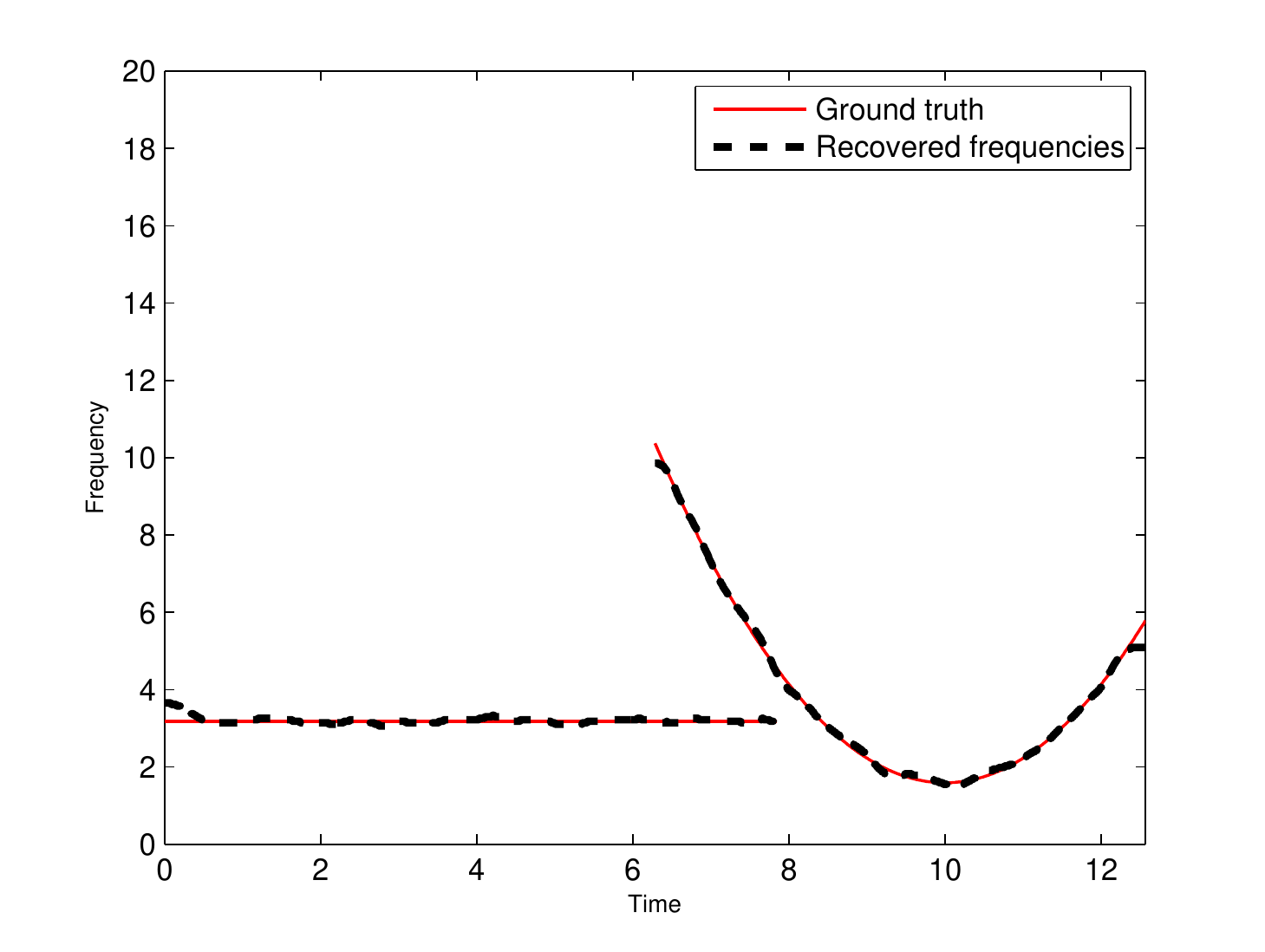}\\
  \includegraphics[width=6cm]{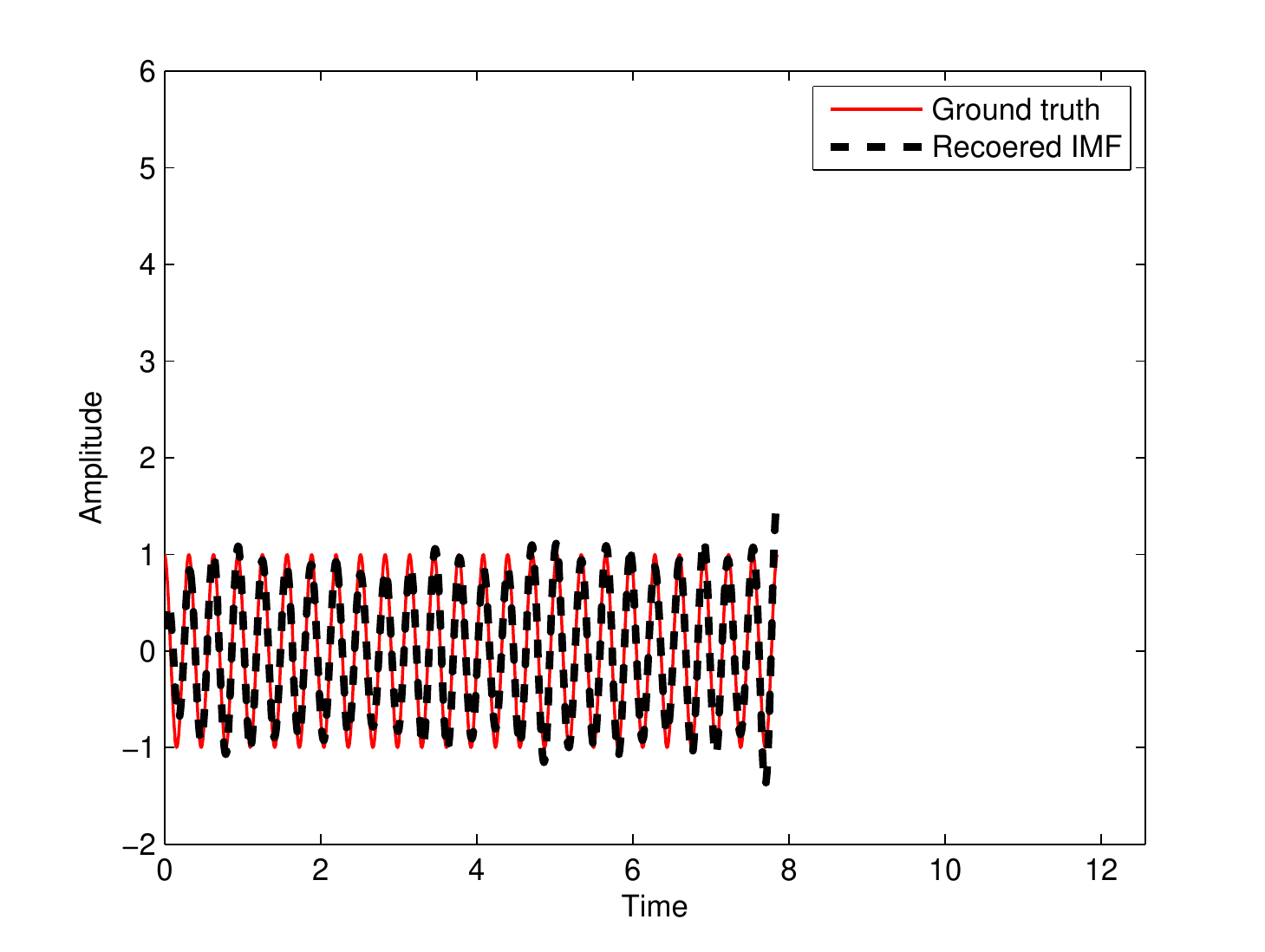}
&\includegraphics[width=6cm]{all_figures/splitsignal_rec_imf1.pdf}
 &\includegraphics[width=6cm]{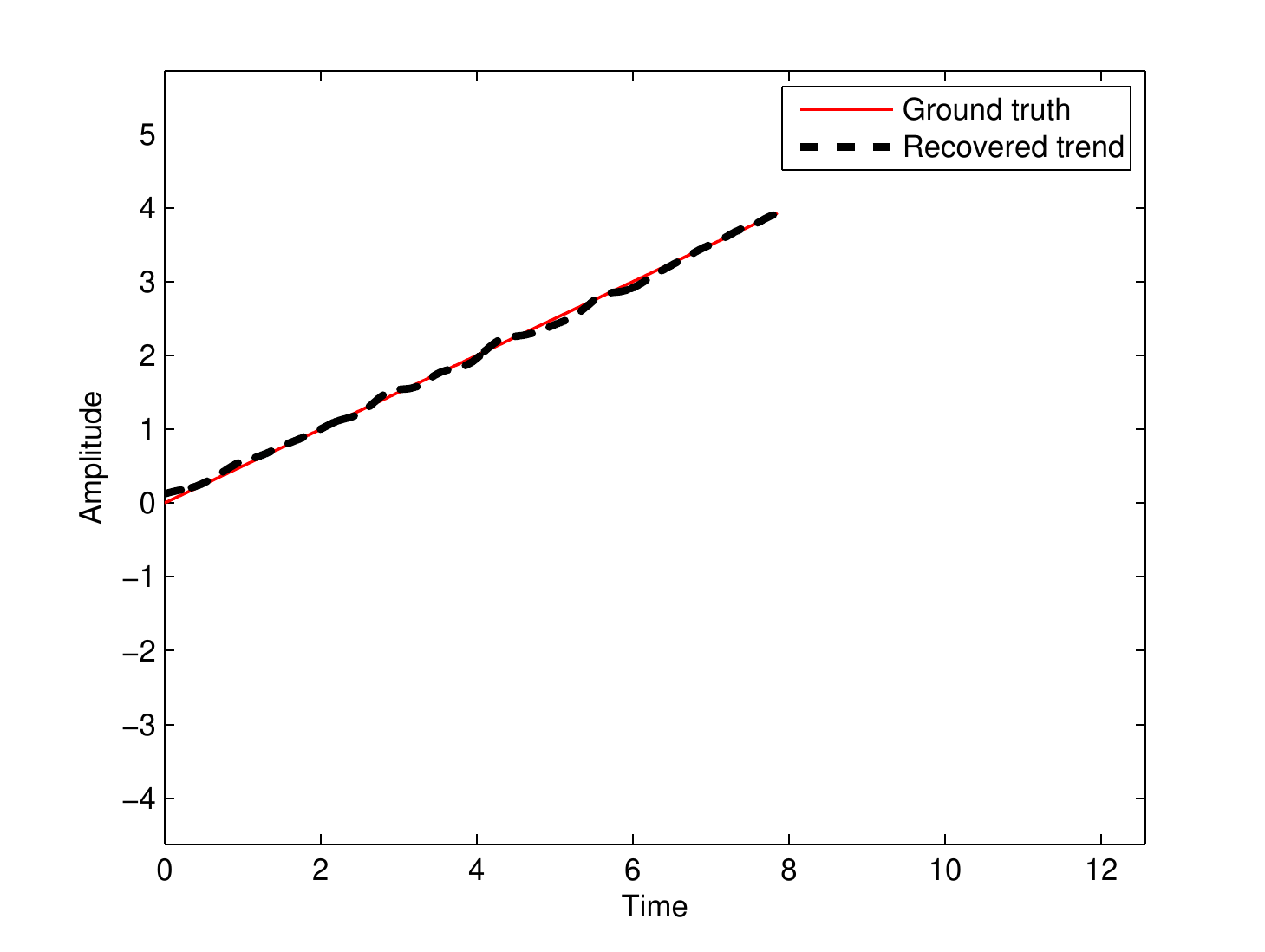}\\
\end{tabular}
\caption{Top (left to right): blind-source signal $f_{I,2}(t)$, $f_{I,2}(t)$ with additive noise, recovered frequencies. Middle (left to right): recovered $1$st IMF, recovered $2$nd IMF,  Bottom: recovered trend.}
\label{fig:splitsignal}
\end{figure}
\qed}
\end{uda}

\begin{uda}\label{uda:varamplitude}
{\rm
\textbf{(Variable amplitudes)} In this example, we illustrate the  proposed method when applied to a non-stationary signal; i.e., a signal where the amplitudes are also time-dependent in addition to the frequencies, defined by
\begin{equation}\label{varamplitude}
\begin{aligned}
f_{I,3}(t)=f_{3,1}(t)+f_{3,2}(t)+f_{3,3}(t)+f_{3,4}(t)+A_{3,0}(t), ~~0< t<30,\\
\end{aligned}
\end{equation}
where
\begin{equation}\label{varamplitudecomp}
\begin{aligned}
&f_{3,1}(t)=\exp((1-t/30)^2+(t/30)^3)\cos(2\pi(t+\cos(t)/10)),\\
&f_{3,2}(t)=(3/2)\cos(2\pi(2t+t^2/100)),\\
&f_{3,3}(t)=(t/10+13/10)\cos(2\pi(7t/2+\sin(t/2)/5)),\\
&f_{3,4}(t)=(1+\cos(\pi t/50)/2)\cos(2\pi(5t+t^2/50)),\\
&A_{3,0}(t)= (64/5)((t-15)/16)^{4}-(64/5)((t-15)/12)^{2}+4t/5+16/5.\\
\end{aligned}
\end{equation}
The signal $f_{I,3}(t)$ is added with a white noise with SNR$=25$dB.
The results are illustrated graphically in Figure~\ref{fig:varamplitude}, with details given in Table~\ref{table1}.
\begin{figure}[ht]
\centering
\begin{tabular}{ccc}
\includegraphics[width=6cm]{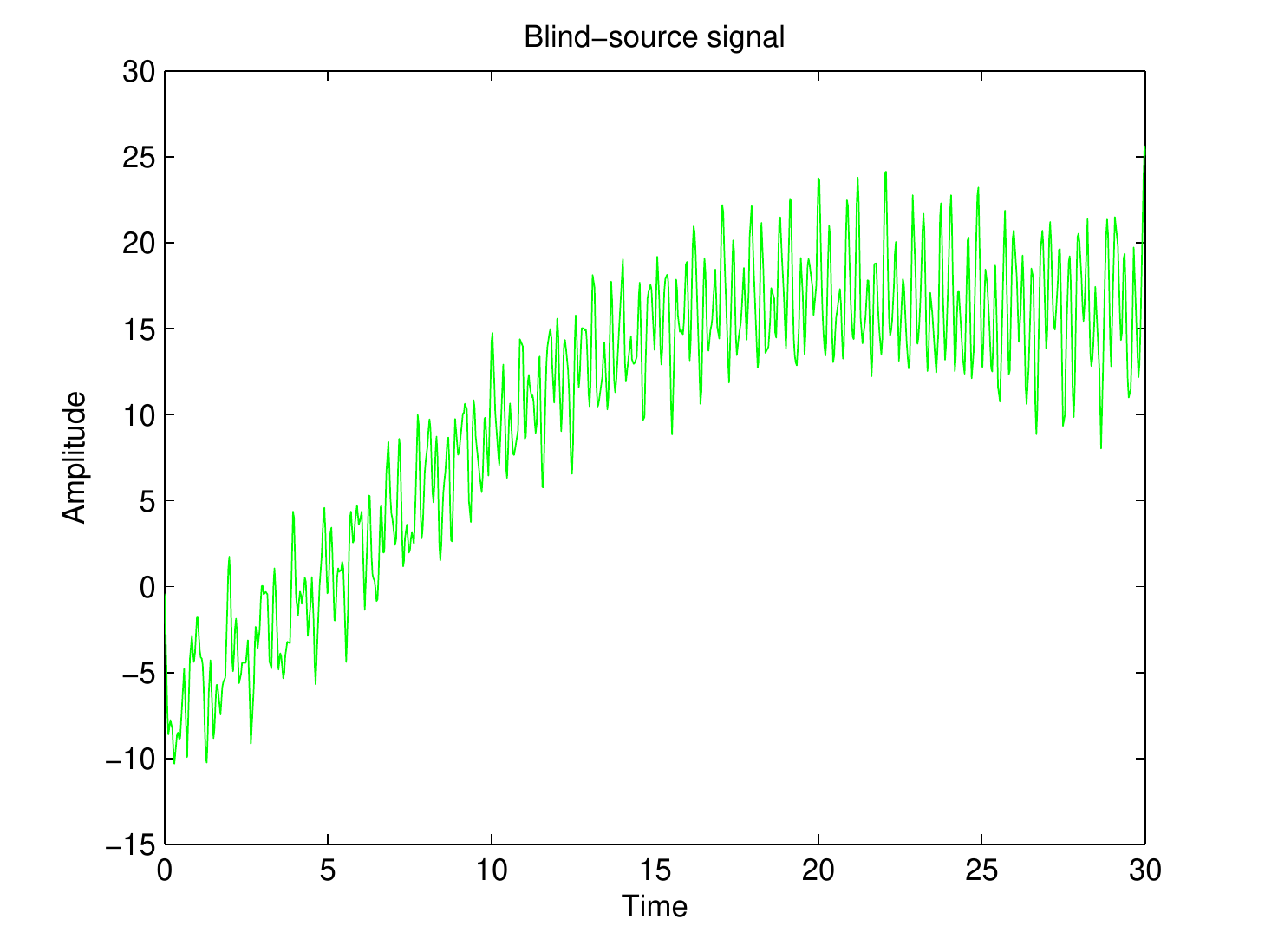}
 & \includegraphics[width=6cm]{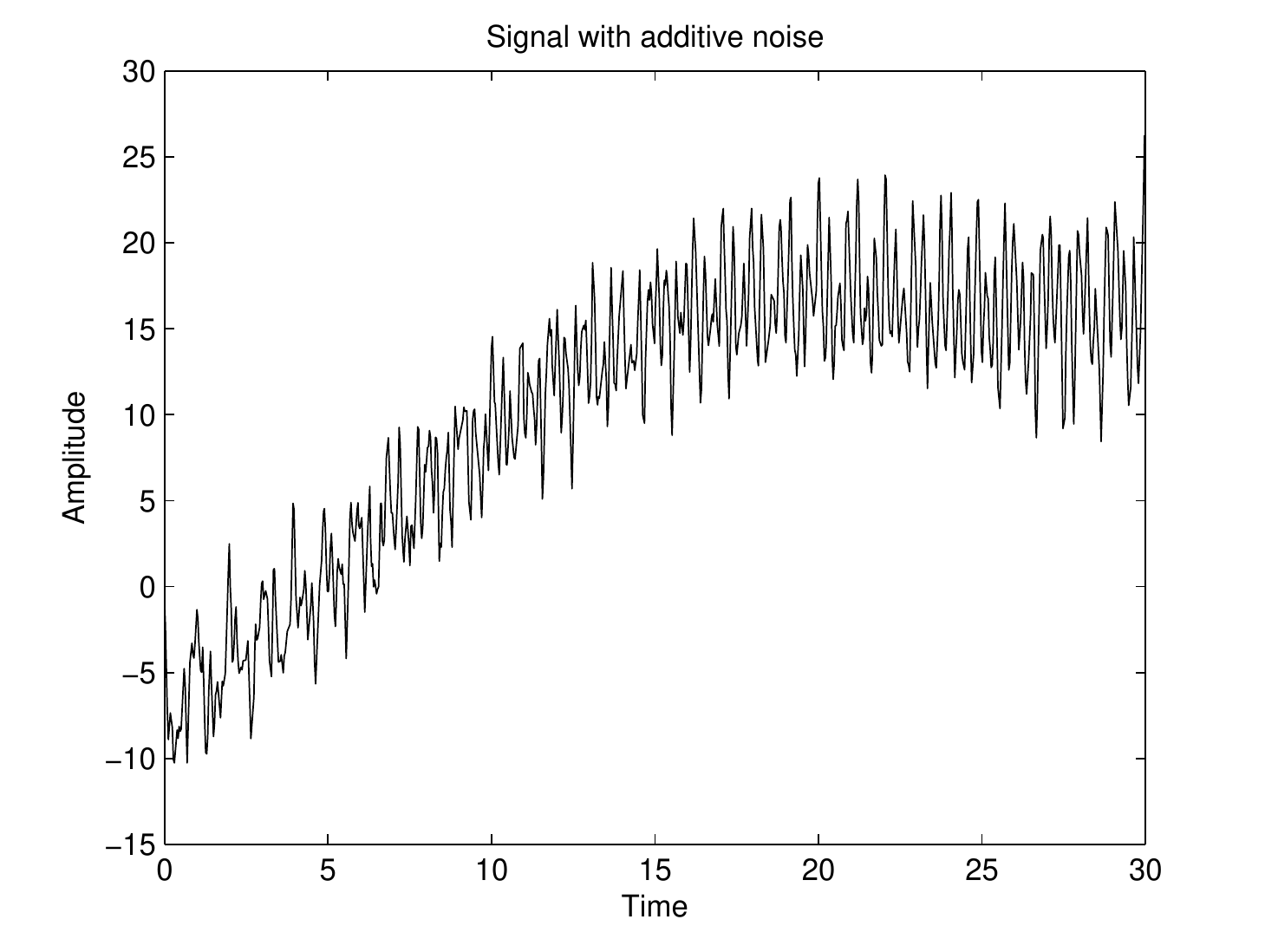}
 & \includegraphics[width=6cm]{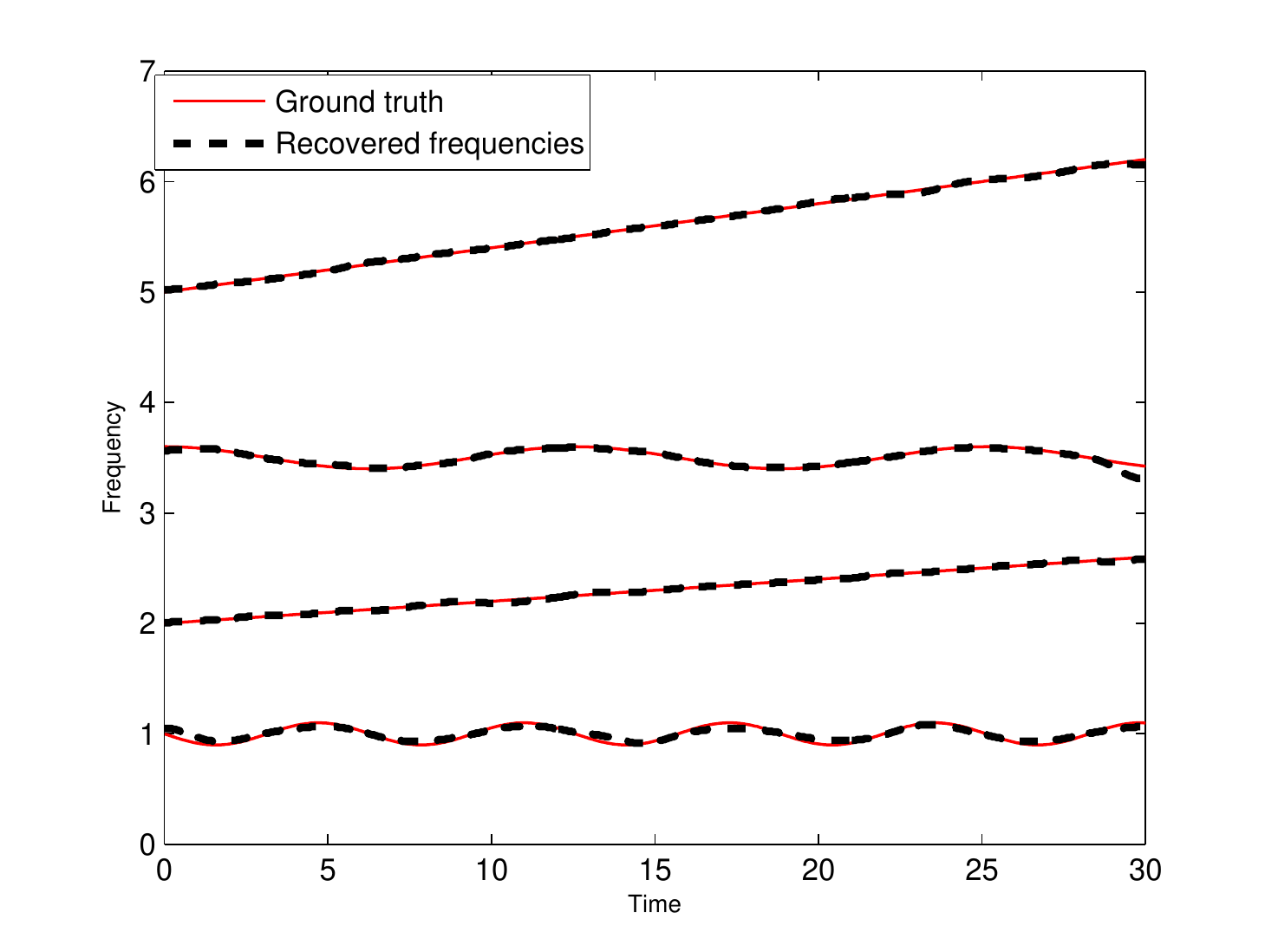}\\
  \includegraphics[width=6cm]{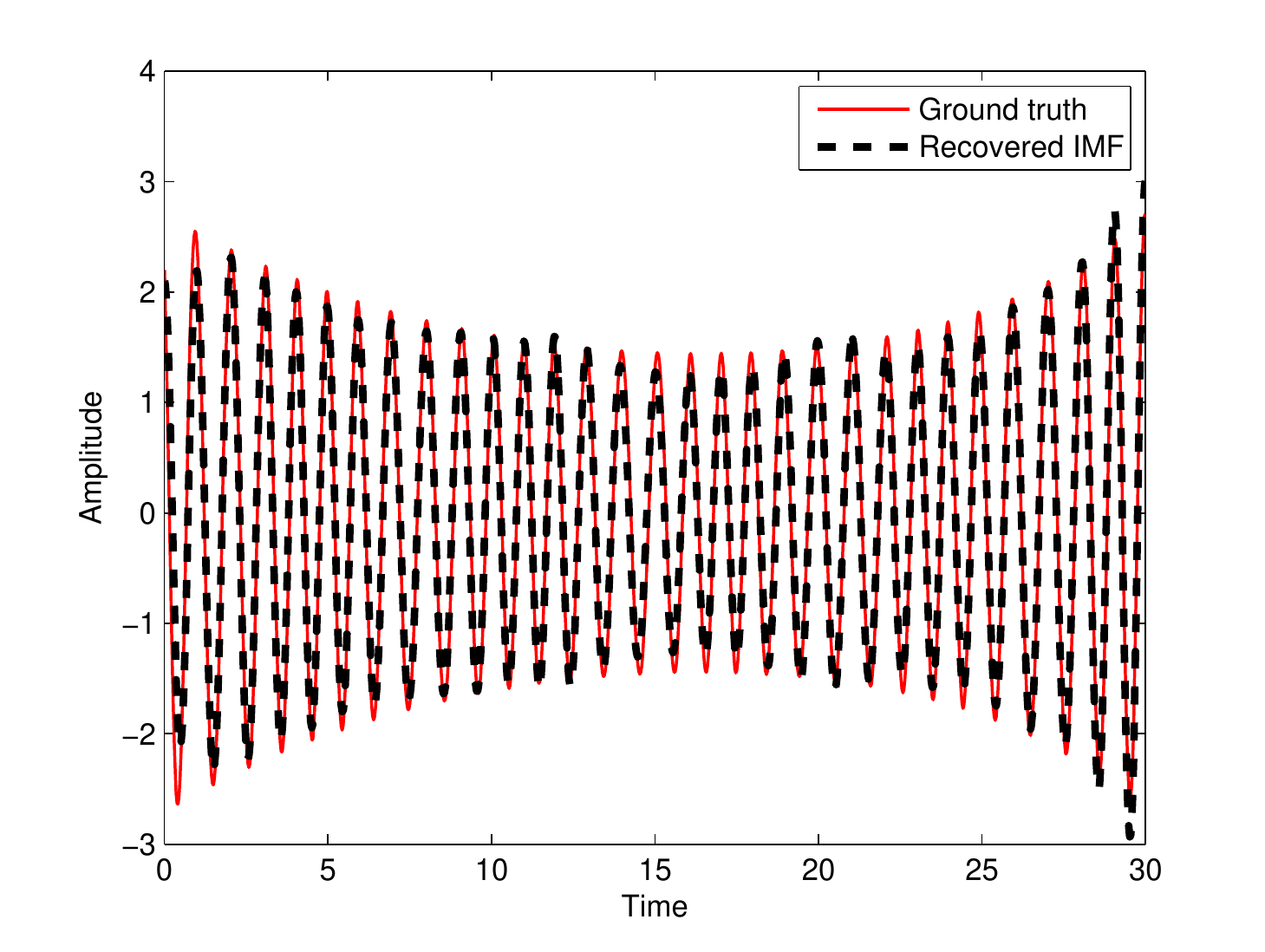}
&\includegraphics[width=6cm]{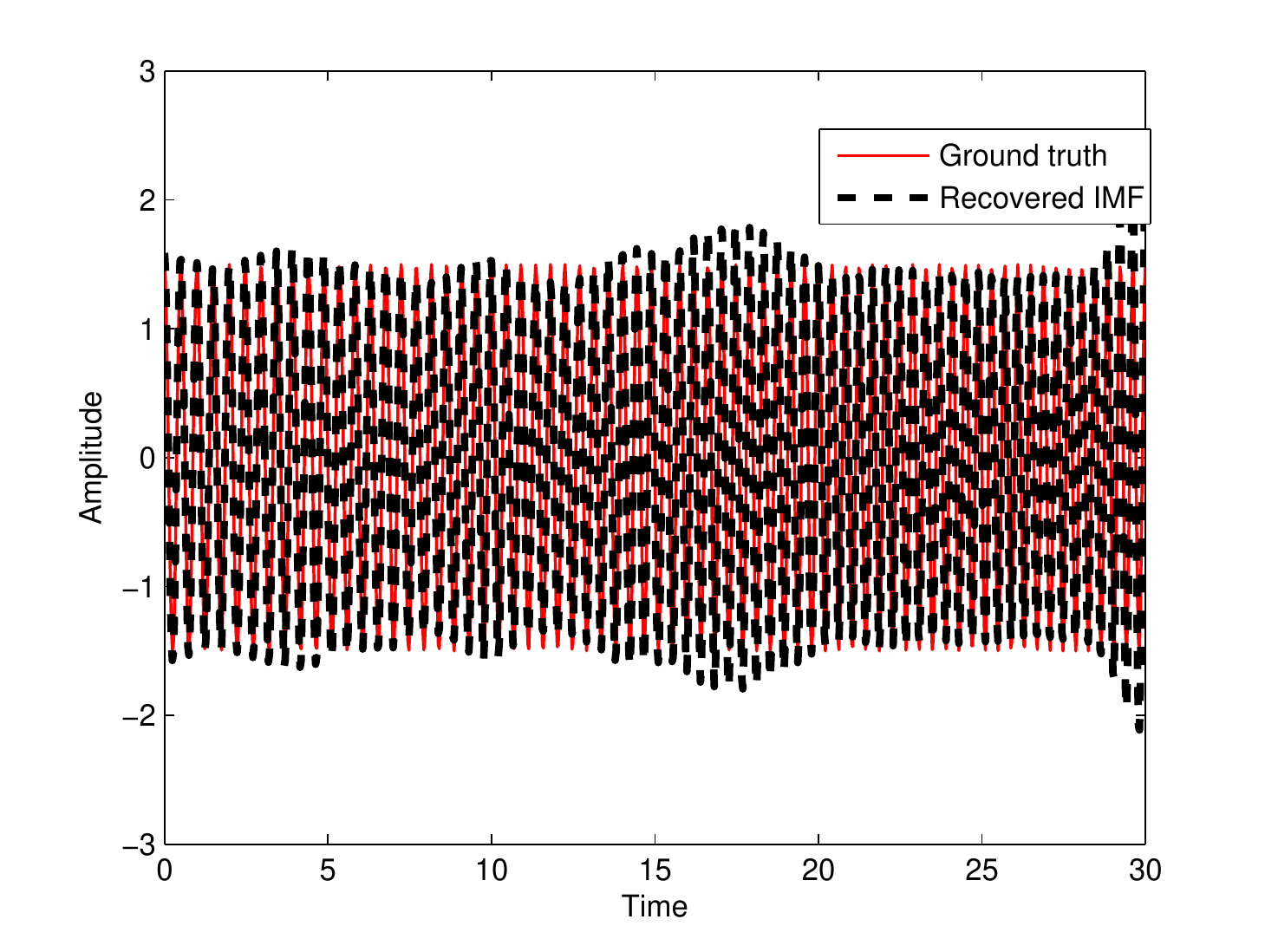}
 &\includegraphics[width=6cm]{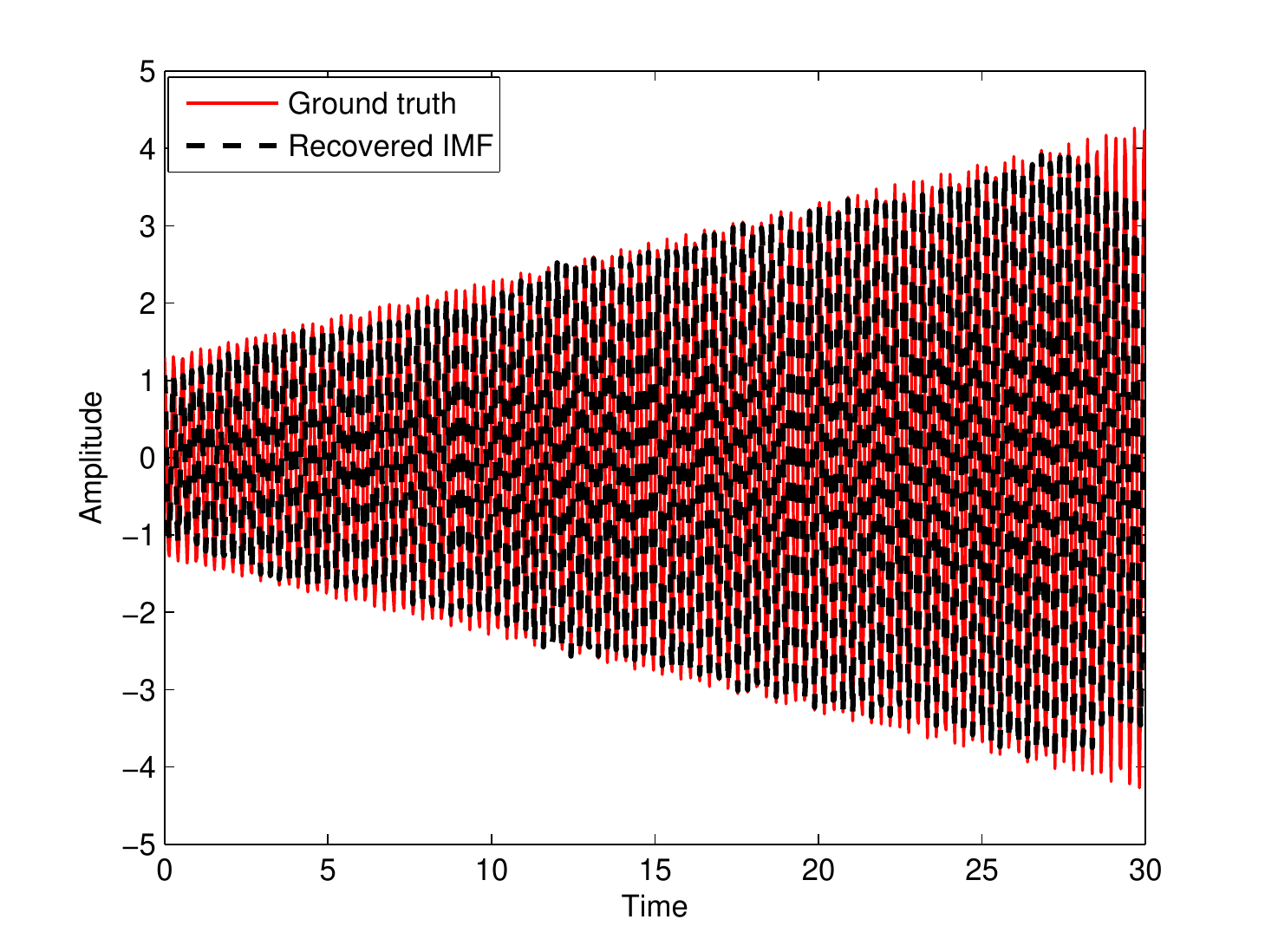}\\
 \includegraphics[width=6cm]{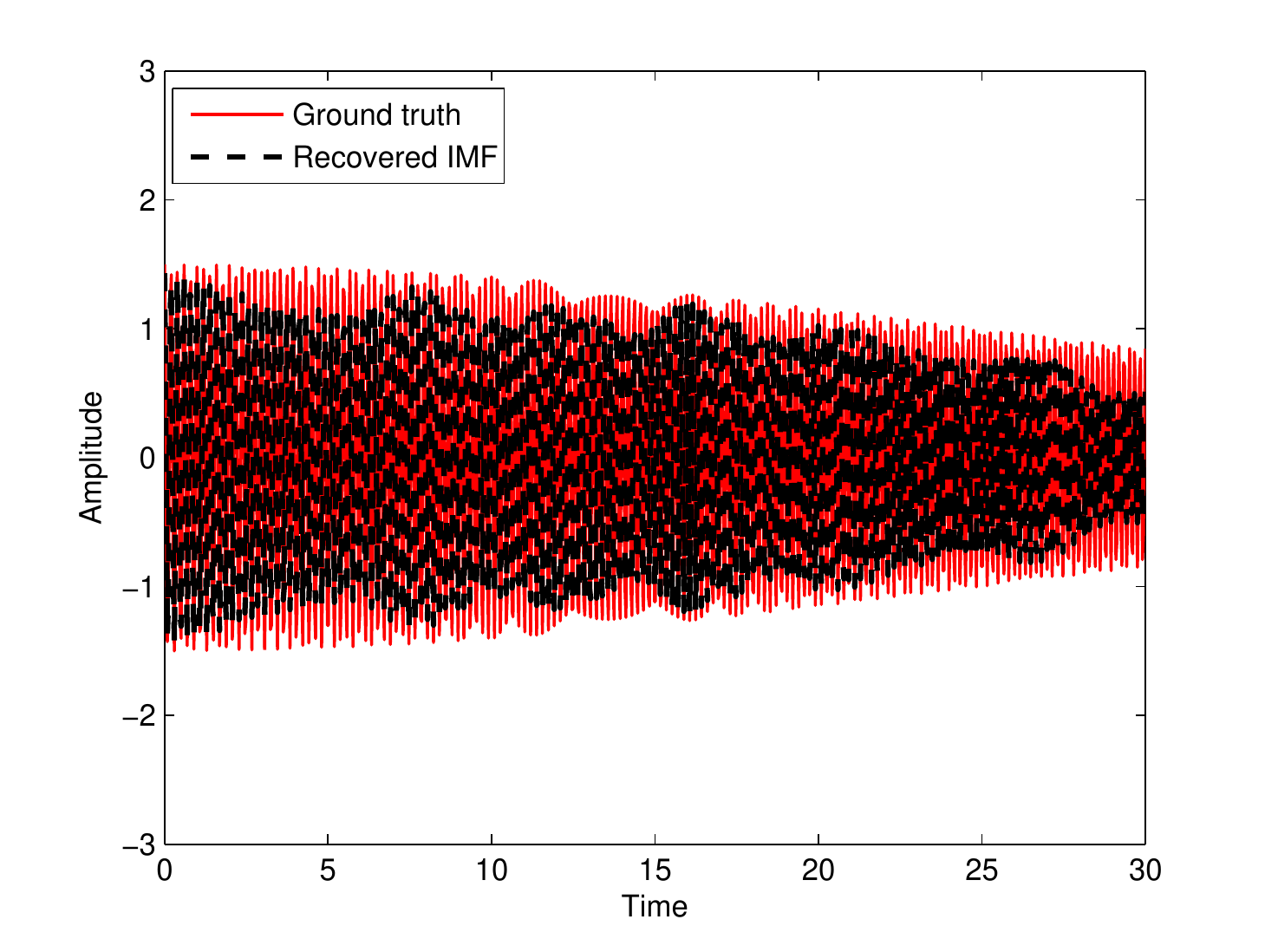}
&\includegraphics[width=6cm]{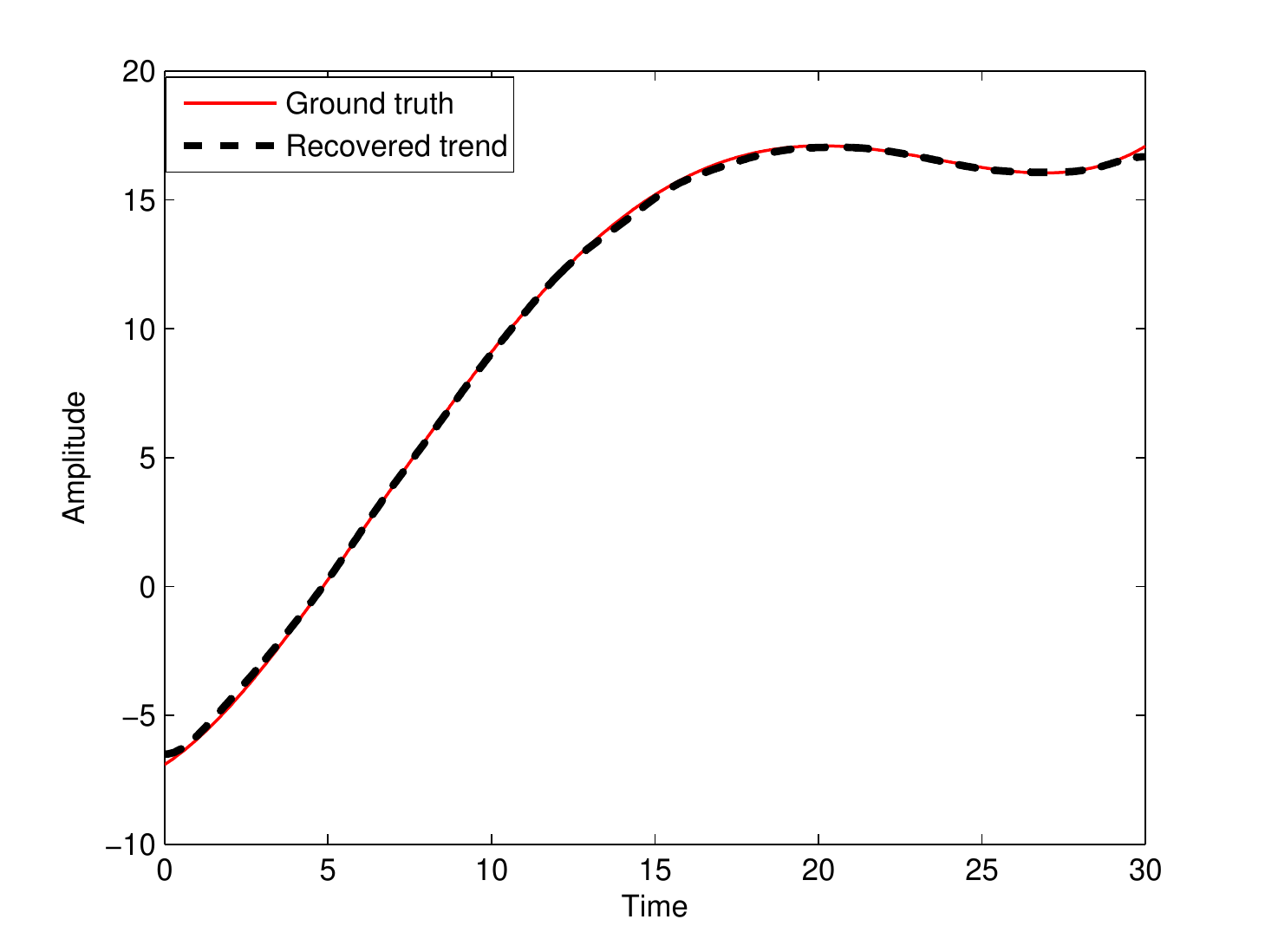}\\
\end{tabular}
\caption{Top (left to right): blind-source signal $f_{I,3}(t)$, $f_{I,3}(t)$ with additive noise, recovered frequencies. Middle (left to right): recovered $1$st IMF, recovered $2$nd IMF,  recovered $3$rd IMF. Bottom: recovered $4$th IMF, recovered trend.}
\label{fig:varamplitude}
\end{figure}
\qed}
\end{uda}

\begin{uda}\label{uda:robustness}
{\rm
\textbf{(Robustness)} 
The following example demonstrates the robustness of our method.
\begin{equation}\label{robustsignal}
\begin{aligned}
f_{I,4}(t)=f_{4,1}(t)+f_{4,2}(t)+A_{4,0}(t), ~~0< t<1,\\
\end{aligned}
\end{equation}
where
\begin{equation}\label{robustcomp}
\begin{aligned}
&f_{4,1}(t)=(1/2)\exp(-(t-1/2)^{2}/50)\cos(2\pi(20t+25t^{2})),\\
&f_{4,2}(t)=(1/4)(1+\cos(\pi t/2))\cos(2\pi(40t+25t^{2})),\\
&A_{4,0}(t)= t^{2}.\\
\end{aligned}
\end{equation}
To sufficiently analyze the robustness, we use our method with the pure  blind source signal $f_{I,4}(t)$ and again the signal added with various levels of white noise (SNR$=0,-5,-8$). The reconstructed results of signals with different noise levels are shown graphically in Figure~\ref{fig:robustness} and and numerically in Table~\ref{table2}.
\begin{figure}[ht]
\centering
\begin{tabular}{cccc}
\includegraphics[width=4.5cm]{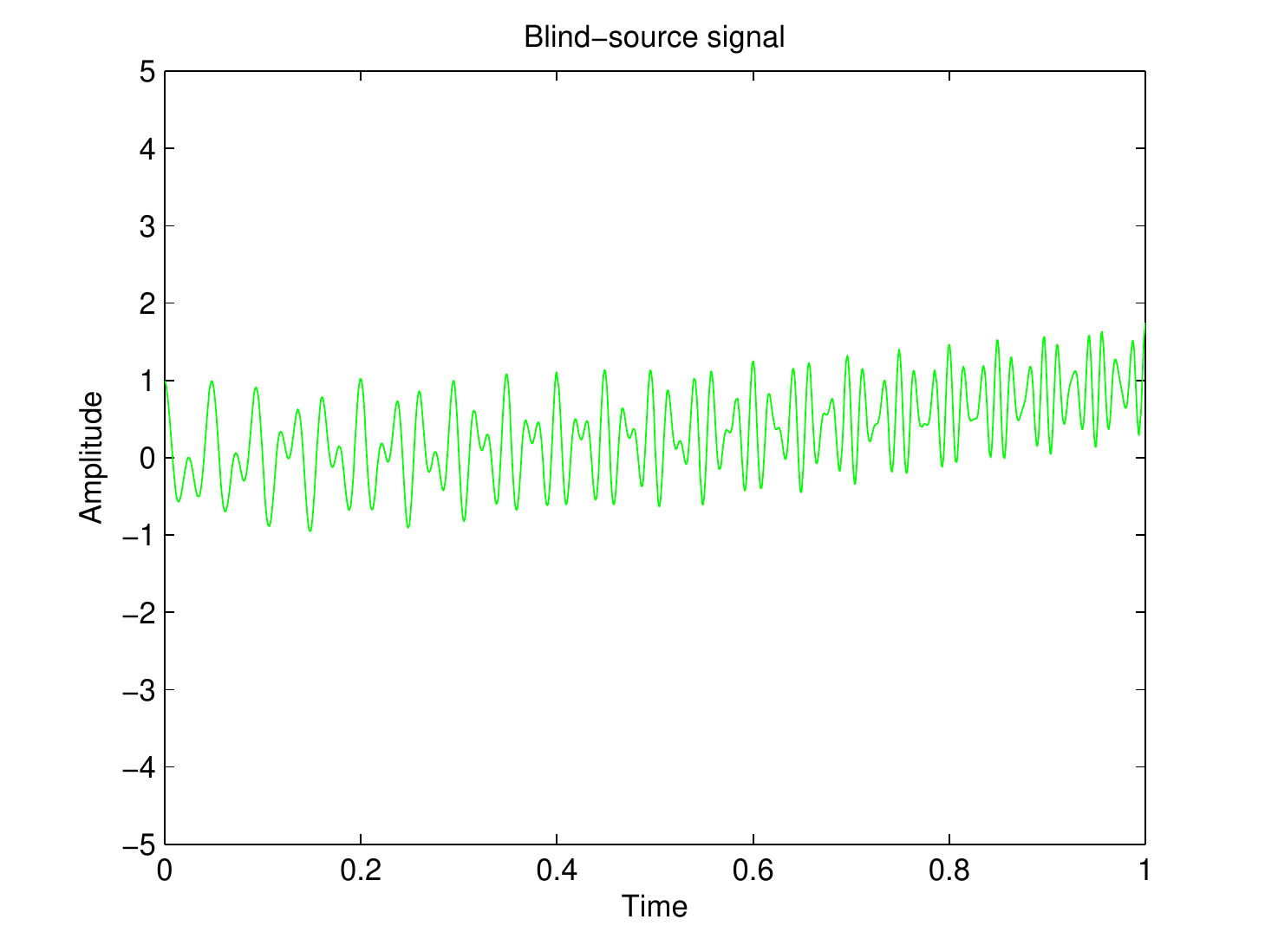}
 & \includegraphics[width=4.5cm]{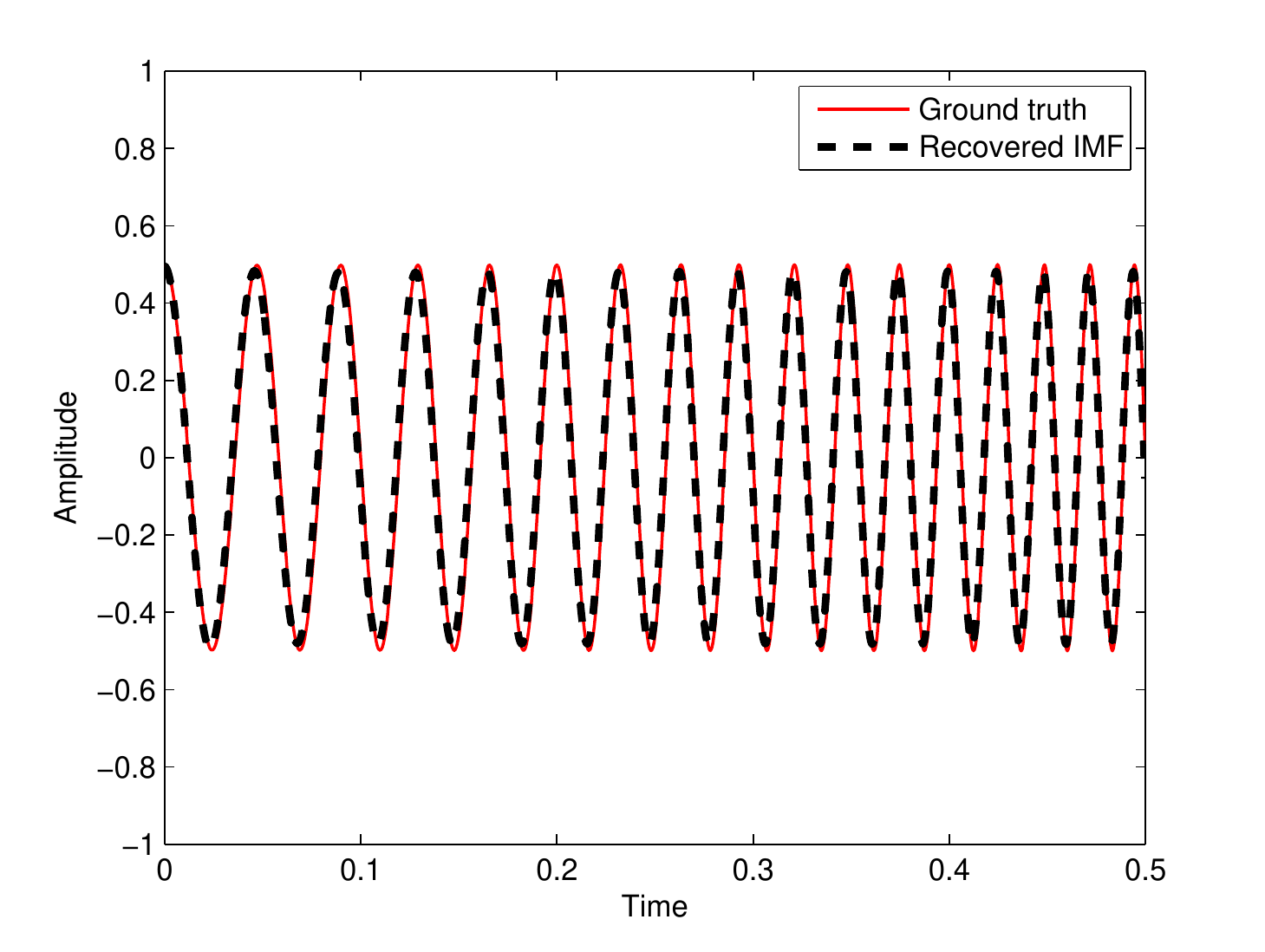}
 & \includegraphics[width=4.5cm]{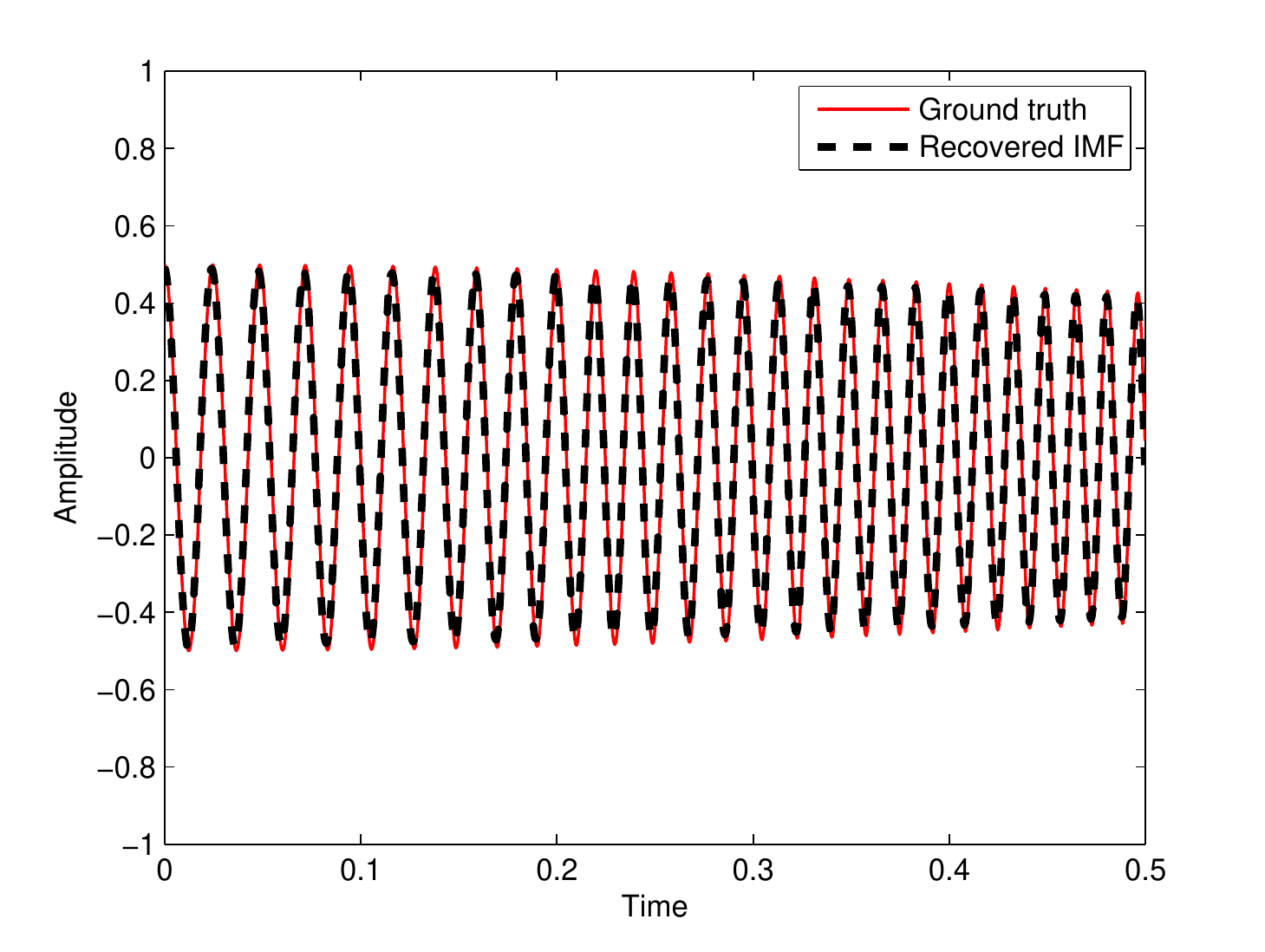}
&\includegraphics[width=4.5cm]{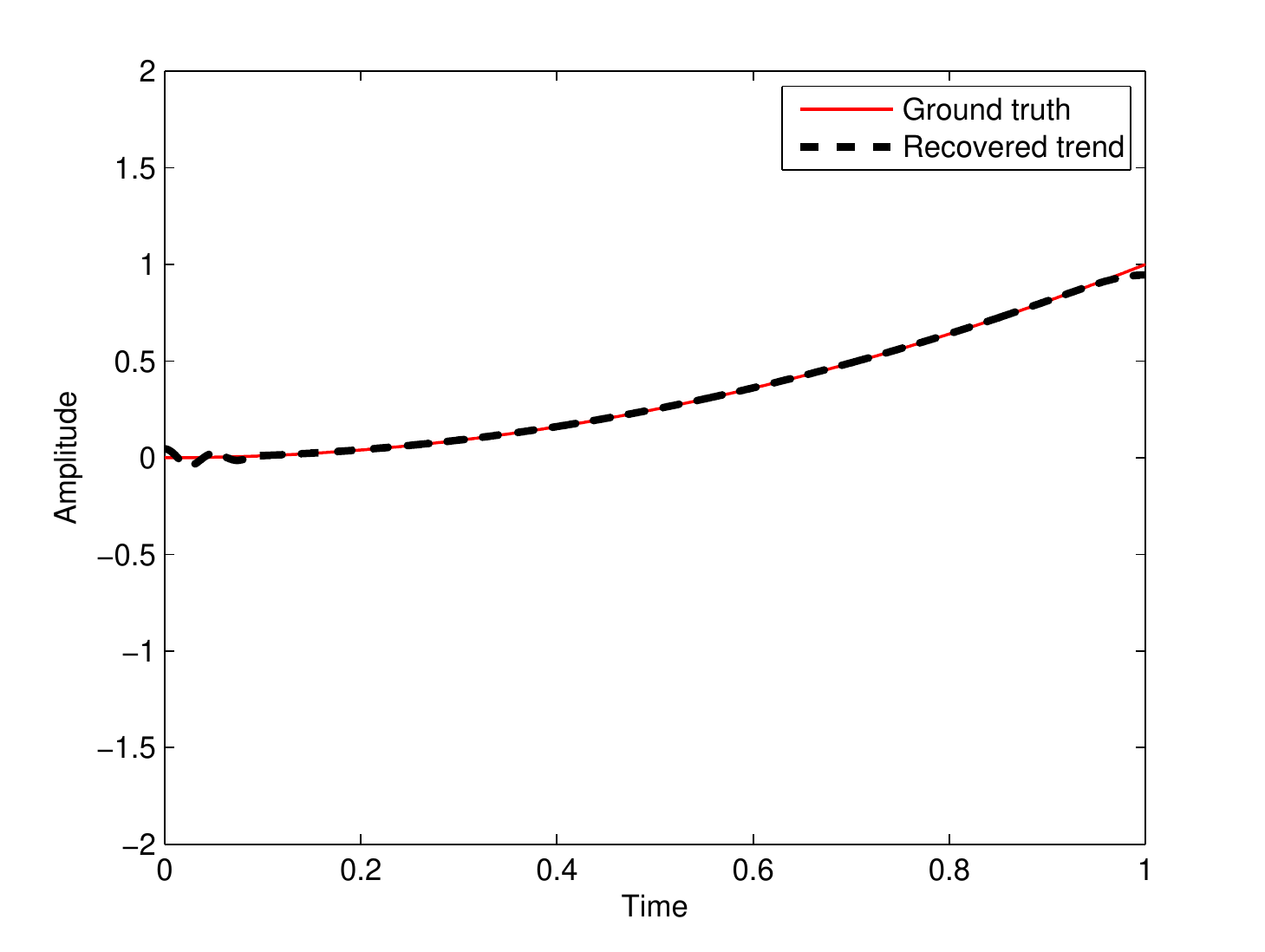}\\
 \includegraphics[width=4.5cm]{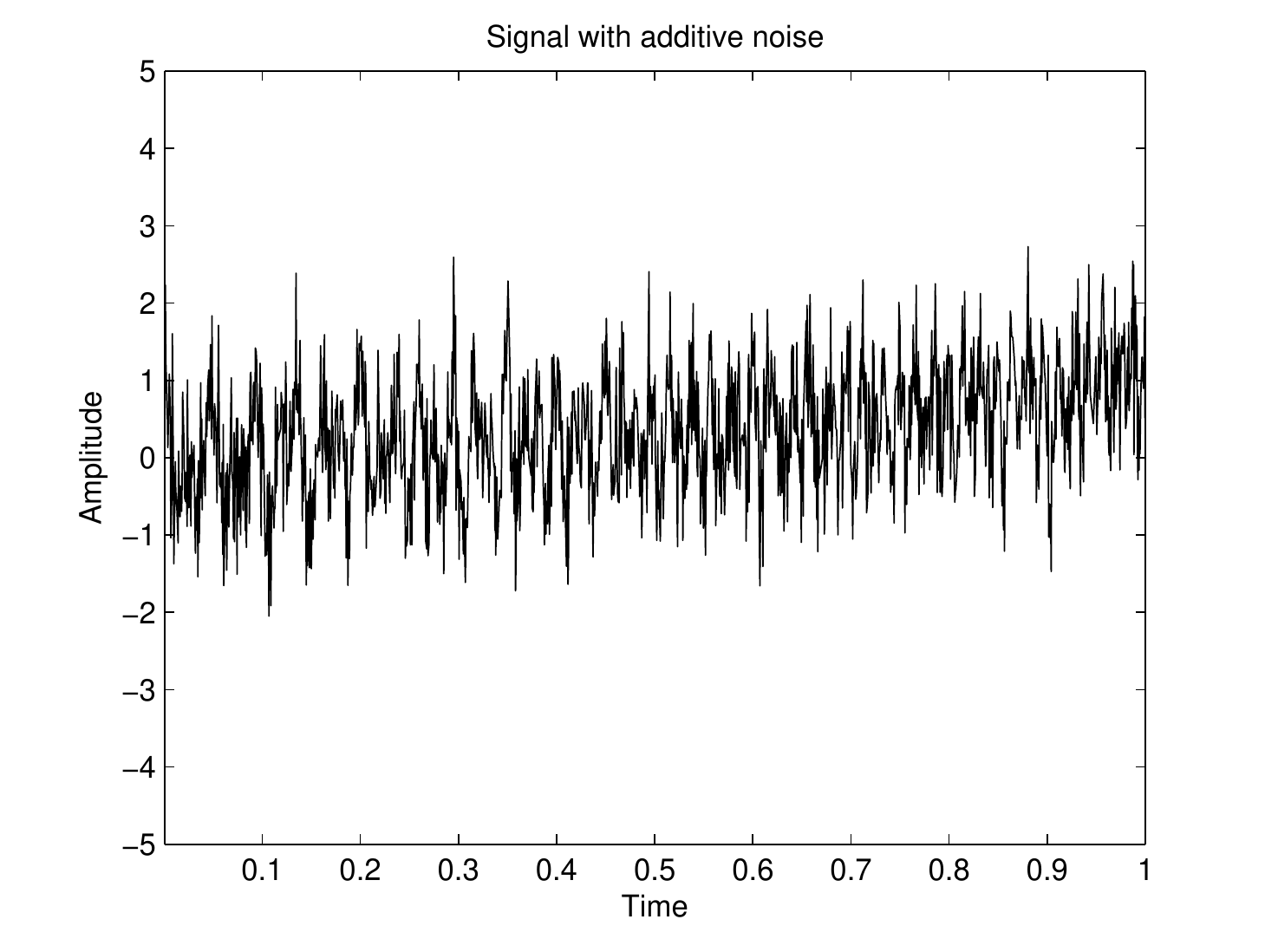}
 & \includegraphics[width=4.5cm]{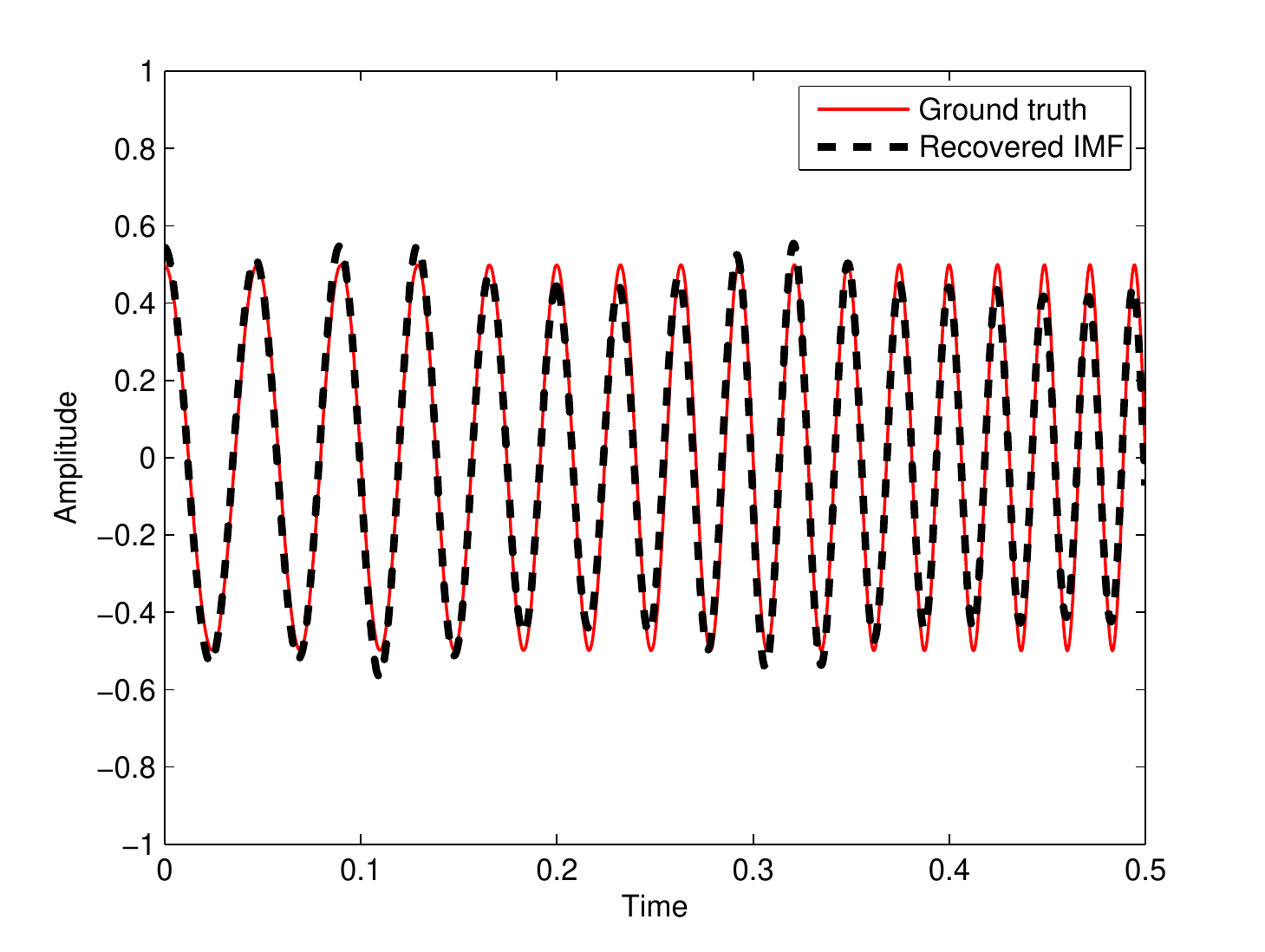}
 & \includegraphics[width=4.5cm]{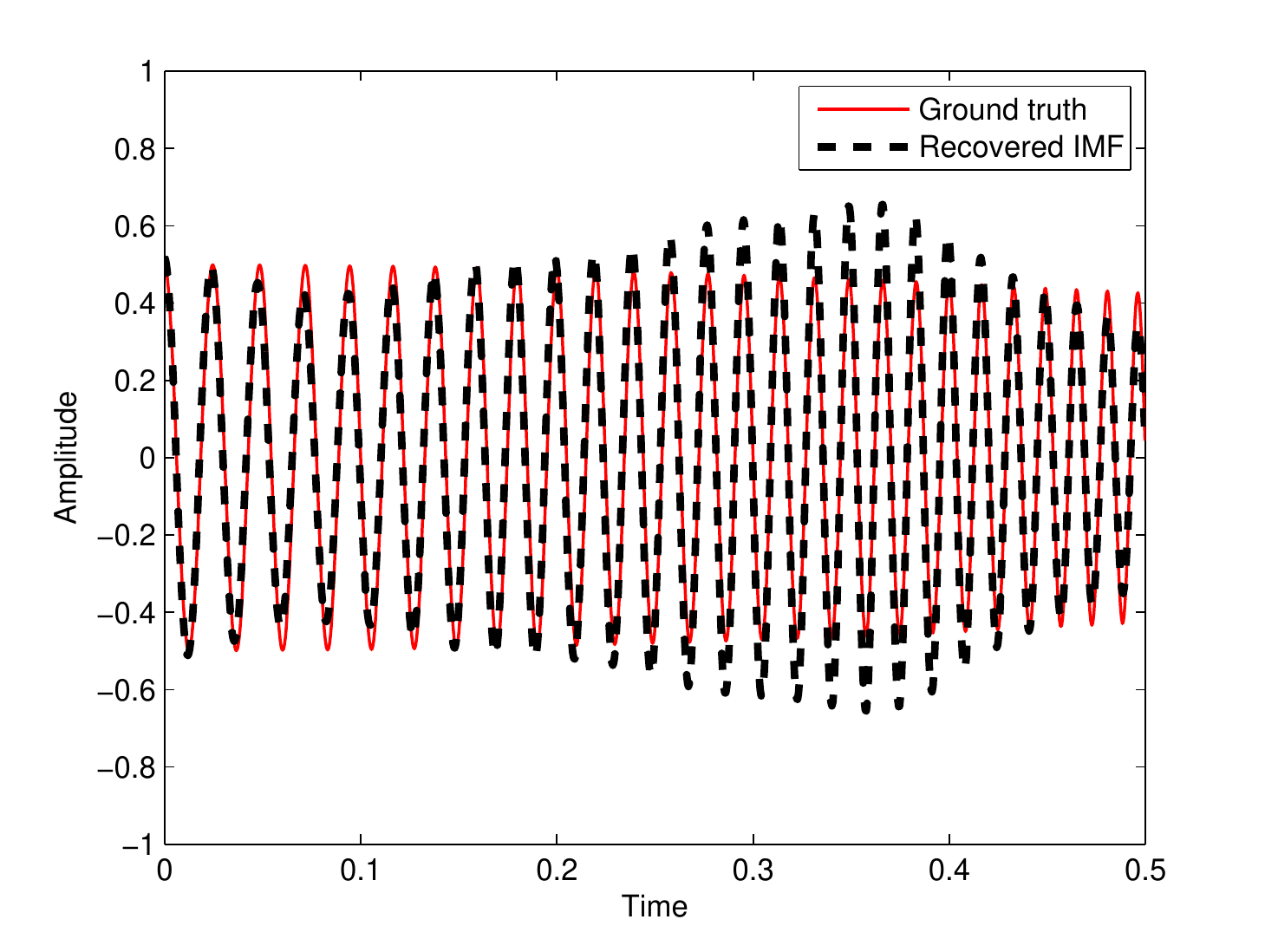}
&\includegraphics[width=4.5cm]{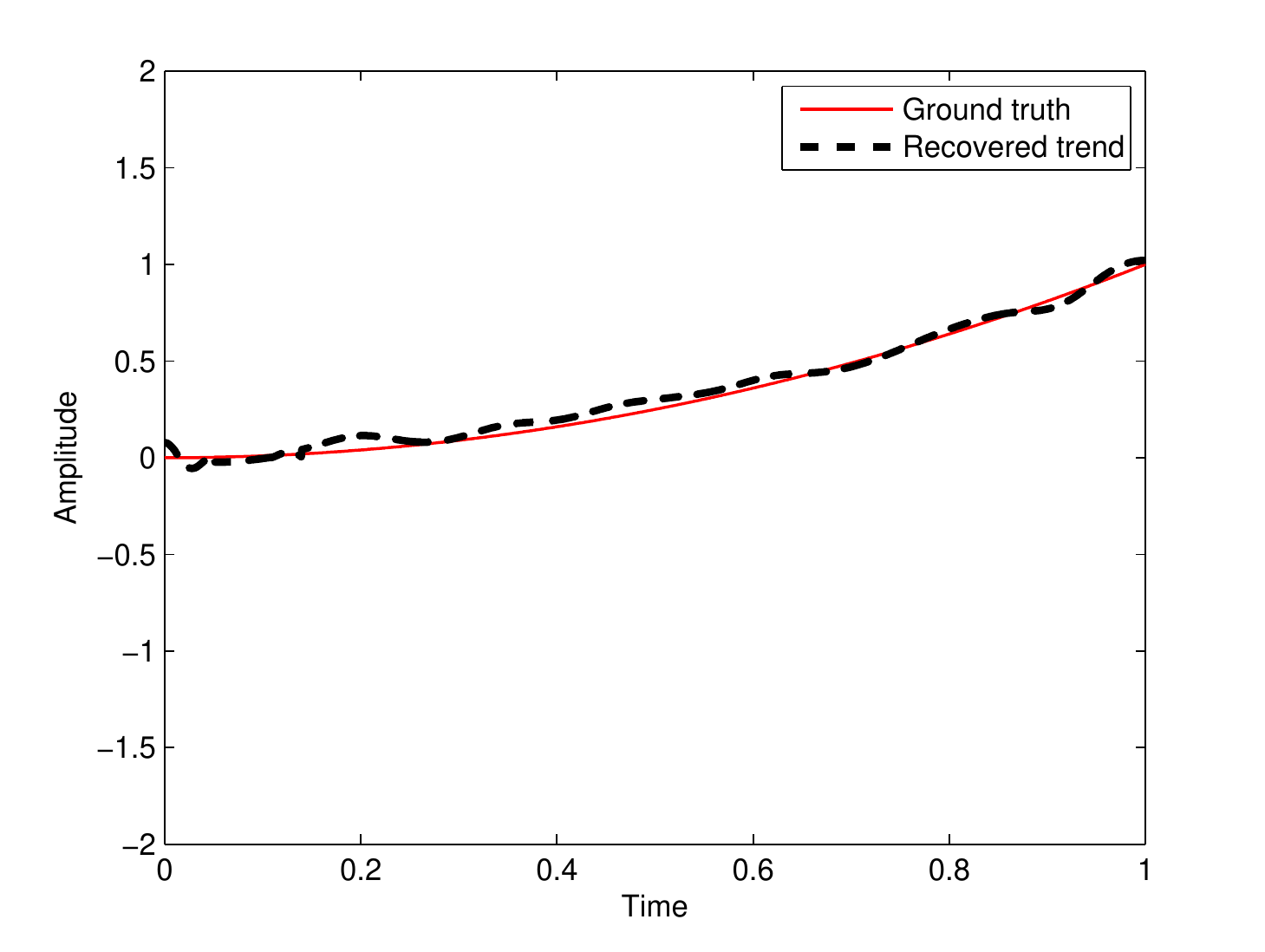}\\
\includegraphics[width=4.5cm]{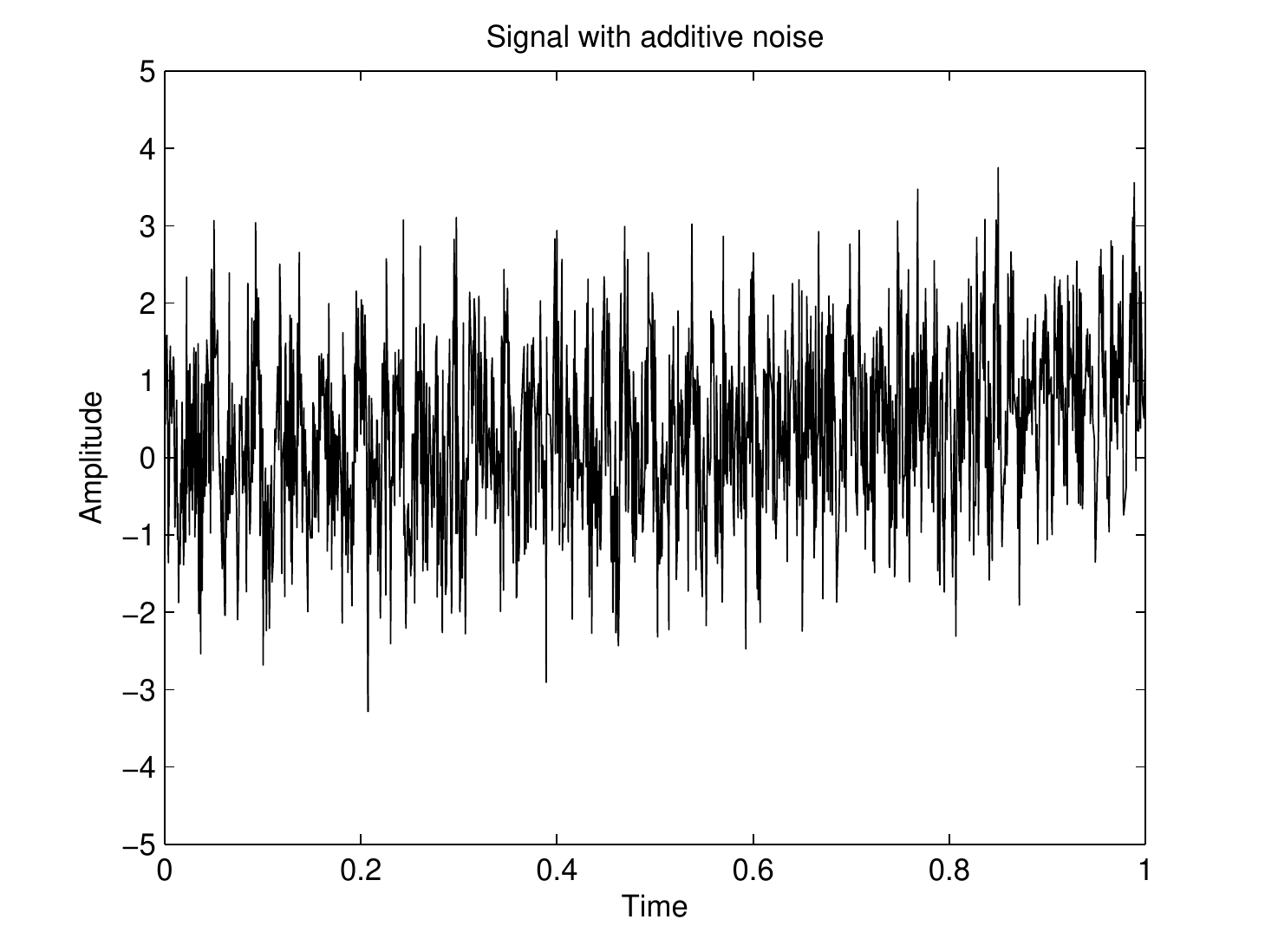}
 & \includegraphics[width=4.5cm]{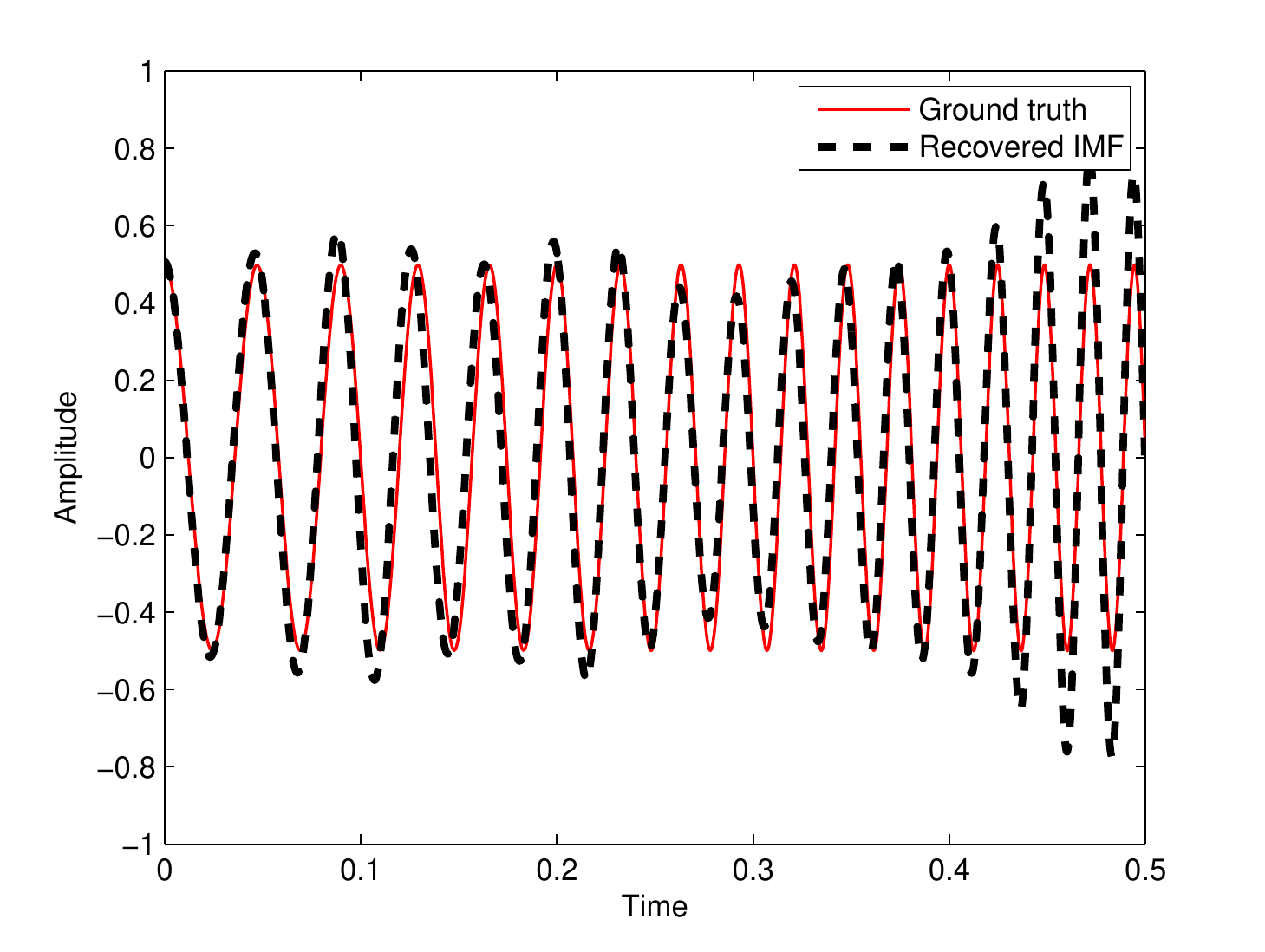}
 & \includegraphics[width=4.5cm]{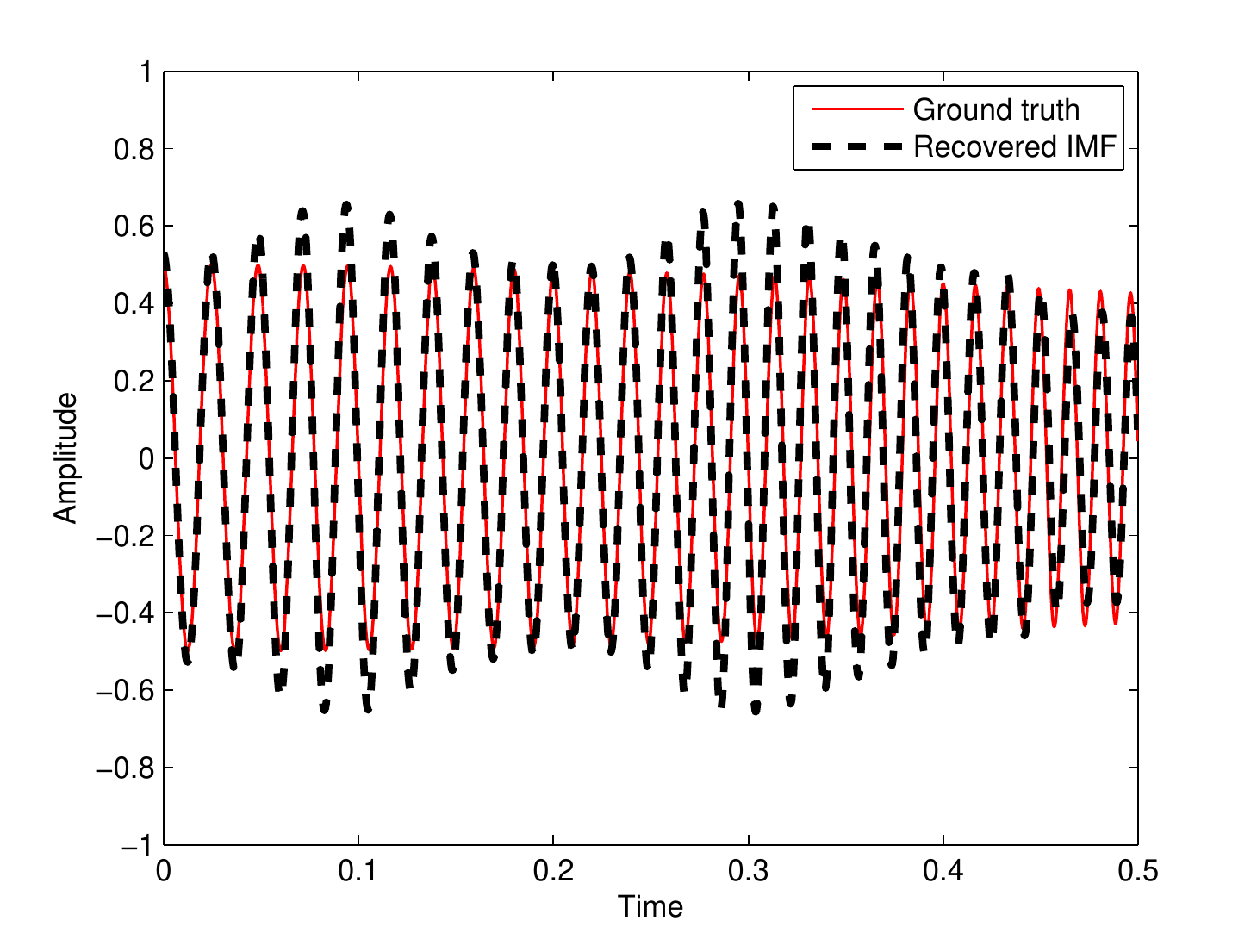}
&\includegraphics[width=4.5cm]{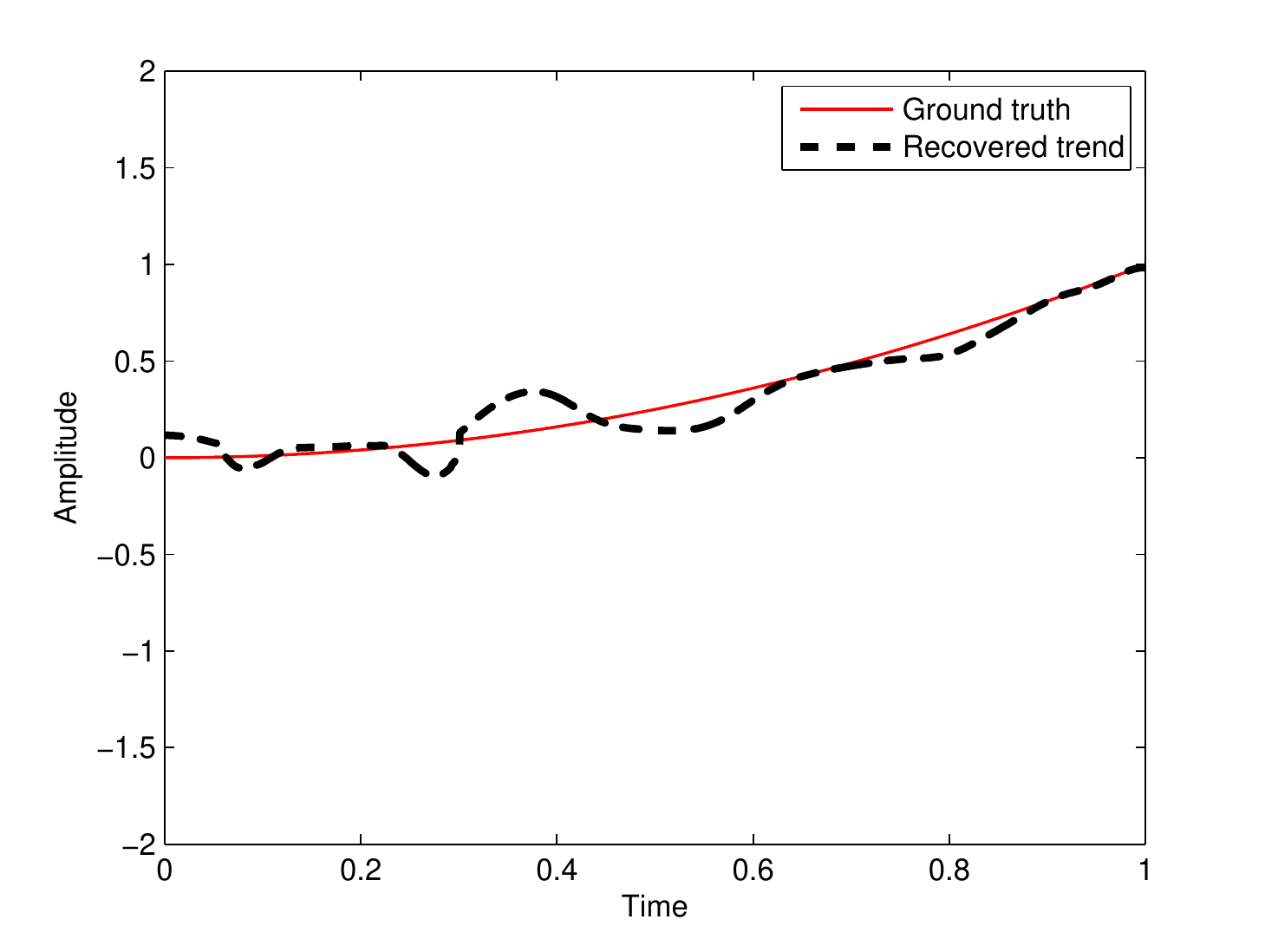}\\
\includegraphics[width=4.5cm]{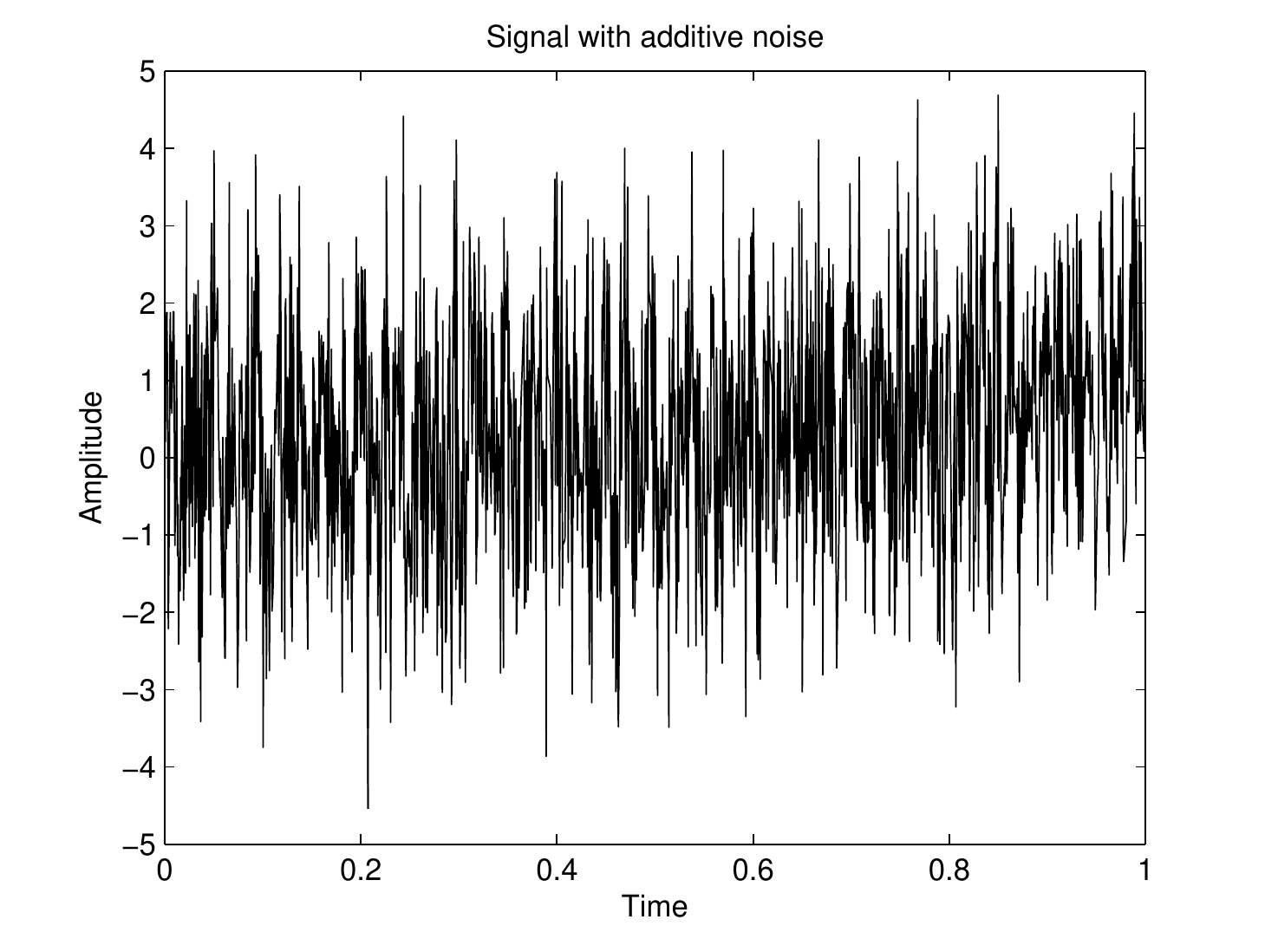}
 & \includegraphics[width=4.5cm]{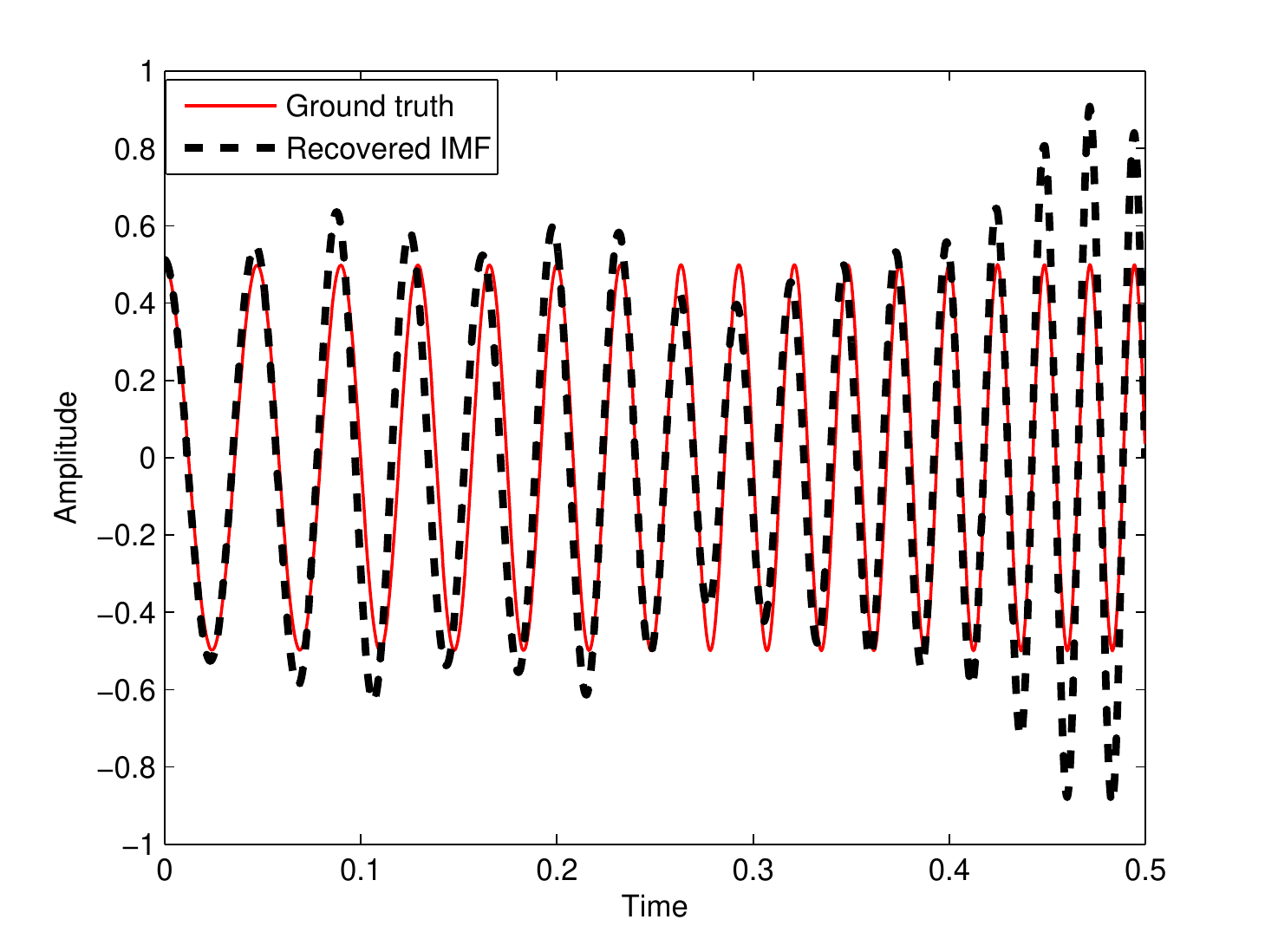}
 & \includegraphics[width=4.5cm]{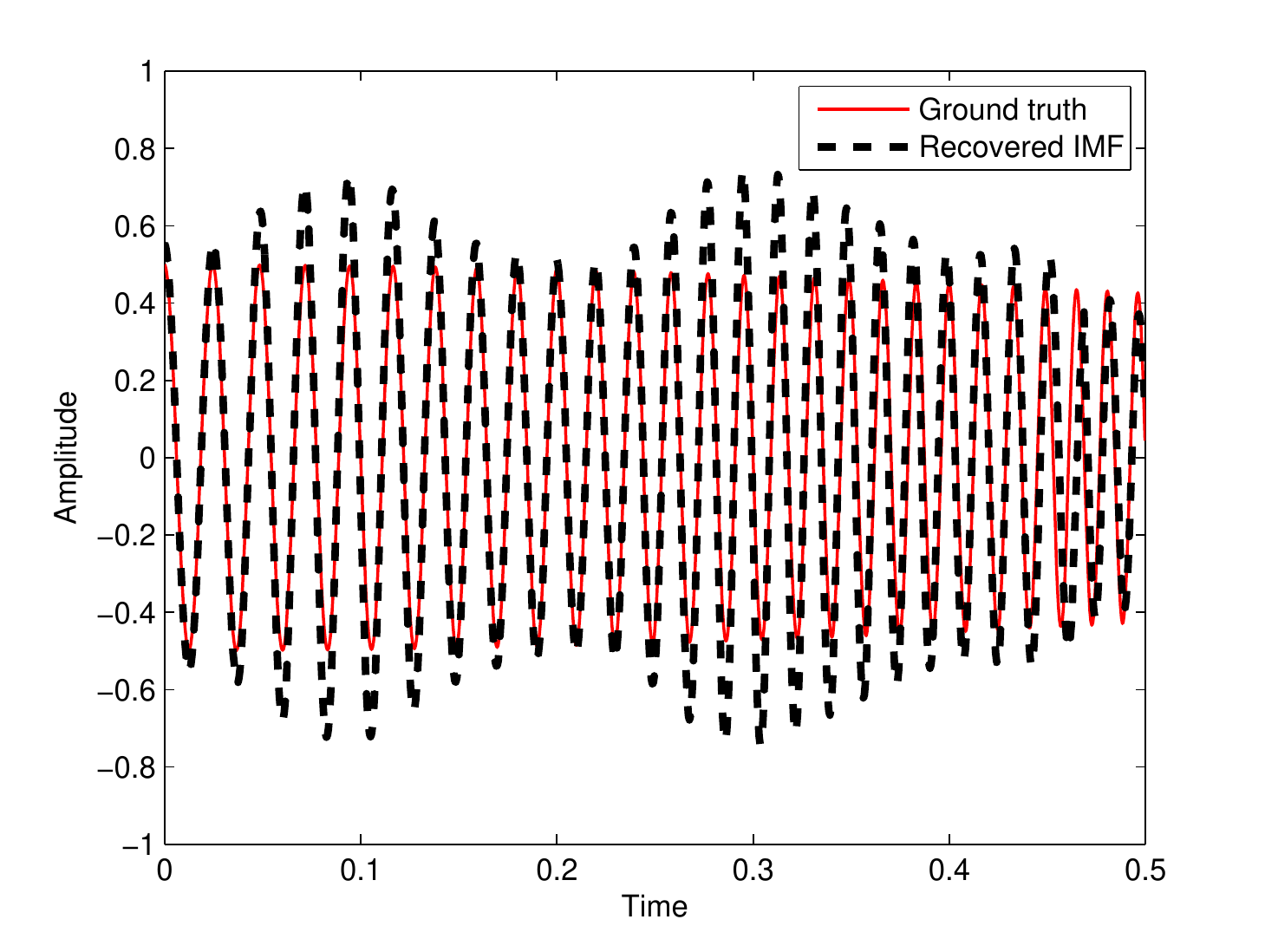}
&\includegraphics[width=4.5cm]{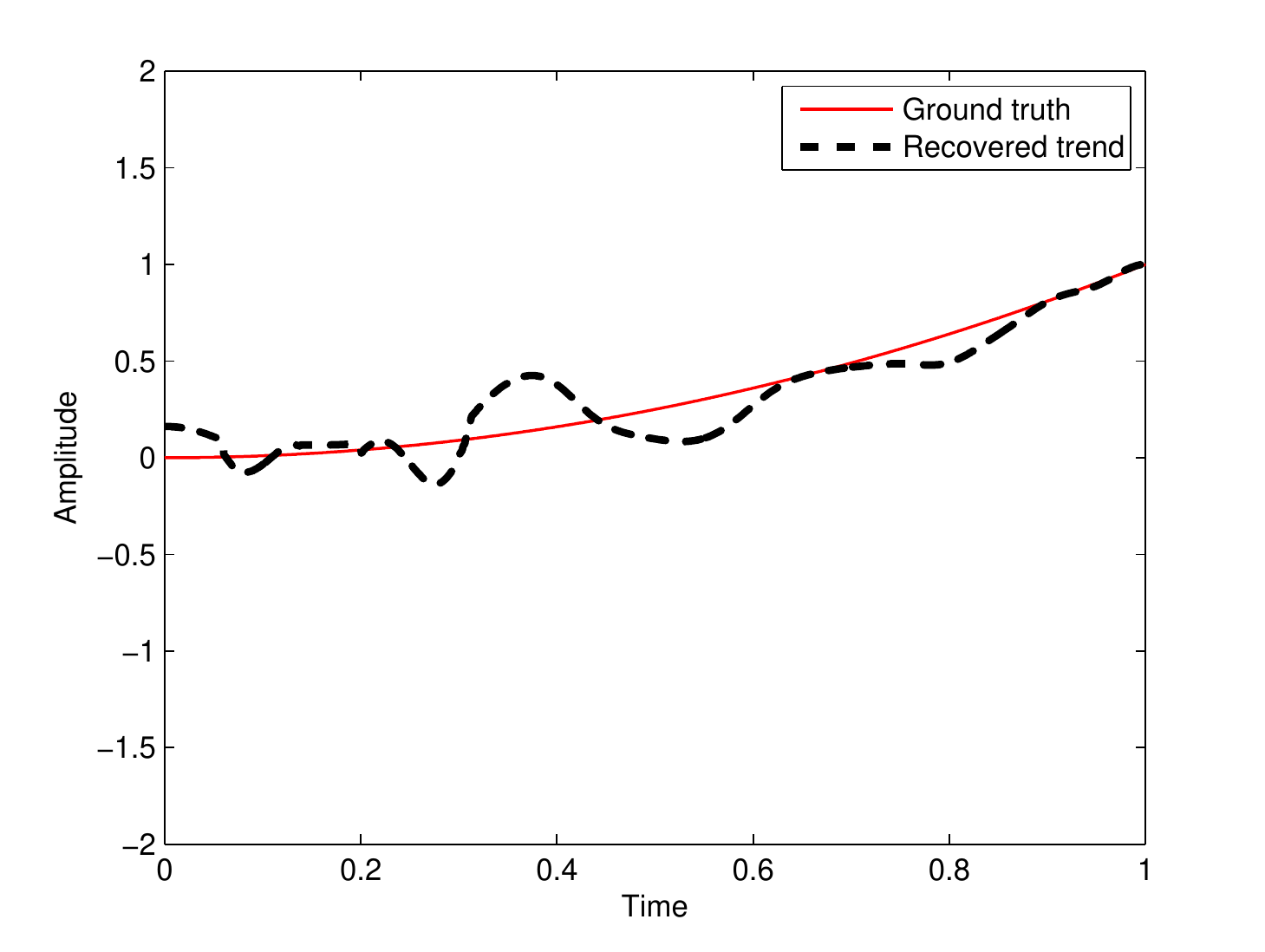}\\
\end{tabular}
\caption{Top (left to right): blind-source signal $f_{I,4}(t)$, recovered 1st IMF, recovered 2nd IMF, recovered trend. Second row (left to right): $f_{I,4}(t)$ with additive noise (SNR=0), recovered 1st IMF, recovered 2nd IMF, recovered trend. Third row (left to right): $f_{I,4}(t)$ with additive noise (SNR=-5), recovered 1st IMF, recovered 2nd IMF, recovered trend. Bottom: (left to right): $f_{I,4}(t)$ with additive noise (SNR=-8), recovered 1st IMF, recovered 2nd IMF , recovered trend.}
\label{fig:robustness}
\end{figure}
\begin{table}[h]
\centering
\scriptsize
\caption{Reconstructed results of Examples $4$}
\label{table2}       
\begin{tabular}{|l|l|l|l|}
\hline
& Size of non-uniform samples (test data) & NMSE (STD) of reconstructed IFs &NMSE (STD) of reconstructed IMFs\\
\hline
$f_{4,1}(t)$ & ~~~~~~~~~~~~~~$1500$~~$(2000)$  & ~~~~~~$5.31\times10^{-5}$~~$(3.84\times10^{-10})$& ~~~~~~$3.18\times10^{-2}$~~$(1.42\times10^{-5})$\\
$f_{4,2}(t)$ & ~~~~~~~~~~~~~~$1500$~~$(2000)$  & ~~~~~~$2.82\times10^{-5}$~~$(7.88\times10^{-8})$ &~~~~~~$3.19\times10^{-2}$~~$(7.11\times10^{-5})$\\
$A_{4,0}(t)$ & ~~~~~~~~~~~~~~$1500$~~$(2000)$  & ~~~~~~~~~~~~~~~~~~~~$\sim$~~~~~~~& ~~~~~~$4.15\times10^{-4}$~~$(5.26\times10^{-6})$\\
\hline
$f_{4,1}(t)$ & ~~~~~~~~~~~~~~$1500$~~$(2000)$  & ~~~~~~$1.60\times10^{-4}$~~$(1.70\times10^{-5})$& ~~~~~~$6.24\times10^{-2}$~~$(1.12\times10^{-2})$\\
$f_{4,2}(t)$ & ~~~~~~~~~~~~~~$1500$~~$(2000)$  & ~~~~~~$1.81\times10^{-4}$~~$(1.60\times10^{-5})$ &~~~~~~$7.84\times10^{-2}$~~$(1.37\times10^{-2})$\\
$A_{4,0}(t)$ & ~~~~~~~~~~~~~~$1500$~~$(2000)$  &~~~~~~~~~~~~~~~~~~~~$\sim$~~~~~~~& ~~~~~~$1.06\times10^{-2}$~~$(1.55\times10^{-3})$\\
\hline
$f_{4,1}(t)$ & ~~~~~~~~~~~~~~$1500$~~$(2000)$  & ~~~~~~$1.61\times10^{-2}$~~$(5.00\times10^{-3})$& ~~~~~~$1.62\times10^{-1}$~~$(2.63\times10^{-2})$\\
$f_{4,2}(t)$ & ~~~~~~~~~~~~~~$1500$~~$(2000)$  & ~~~~~~$1.14\times10^{-2}$~~$(4.90\times10^{-3})$ &~~~~~~$3.11\times10^{-1}$~~$(8.29\times10^{-2})$\\
$A_{4,0}(t)$ & ~~~~~~~~~~~~~~$1500$~~$(2000)$  & ~~~~~~~~~~~~~~~~~~~~$\sim$~~~~~~~& ~~~~~~$2.37\times10^{-2}$~~$(6.10\times10^{-3})$\\
\hline
$f_{4,1}(t)$ & ~~~~~~~~~~~~~~$1500$~~$(2000)$  & ~~~~~~$3.15\times10^{-2}$~~$(9.00\times10^{-3})$& ~~~~~~$4.09\times10^{-1}$~~$(5.47\times10^{-2})$\\
$f_{4,2}(t)$ & ~~~~~~~~~~~~~~$1500$~~$(2000)$  & ~~~~~~$1.72\times10^{-2}$~~$(5.70\times10^{-3})$ &~~~~~~$6.54\times10^{-1}$~~$(1.49\times10^{-1})$\\
$A_{4,0}(t)$ & ~~~~~~~~~~~~~~$1500$~~$(2000)$  & ~~~~~~~~~~~~~~~~~~~~$\sim$~~~~~~~& ~~~~~~$4.73\times10^{-2}$~~$(1.25\times10^{-2})$\\
\hline
\end{tabular}
\end{table}
\qed}
\end{uda}

\begin{uda}\label{uda:extrapolation}
{\rm
\textbf{(Extrapolation)} Our aim in this example is to examine the utility of our methods in extrapolating the signal a little bit beyond the time interval where its values are available, both in the past and in the future.  
We consider 
\begin{equation}\label{extapolation}
\begin{aligned}
f_{I,5}(t)=f_{5,1}(t)+f_{5,2}(t)+A_{5,0}(t), ~~0< t<30,\\
\end{aligned}
\end{equation}
where
\begin{equation}\label{extrapolationcomp}
\begin{aligned}
&f_{5,1}(t)=\exp(2(t/30)^{2}-t/15+1)\cos(2\pi(2t+0.01t^{2})),\\
&f_{5,2}(t)=(1+4(t/30)^{2}+3(1-t/30)^{3})\cos(2\pi(4t+0.01t^{2})),\\
&A_{5,0}(t)= t^{3}/100-47t^{2}/100+27t/5-9.\\
\end{aligned}
\end{equation}
To verify that our approach applies to data extrapolation, we process the data on the  time-interval $[0,30]$, and predict the value of the IFs, IMFs, and the trend on the interval  $[-0.1,30.1]$; i.e., for $0.1$ time units in the past and the same in the future. 
To demonstrate that our extrapolated graphs are good approximation of the unknown ground truth, we zoom-in to the intervals $[-0.1,0.7]$ and $[29.3,30.1]$. 
The results are shown graphically in Figure~\ref{fig:extrapolation}.
\begin{figure}[ht]
\centering
\begin{tabular}{ccc}
\includegraphics[width=6cm]{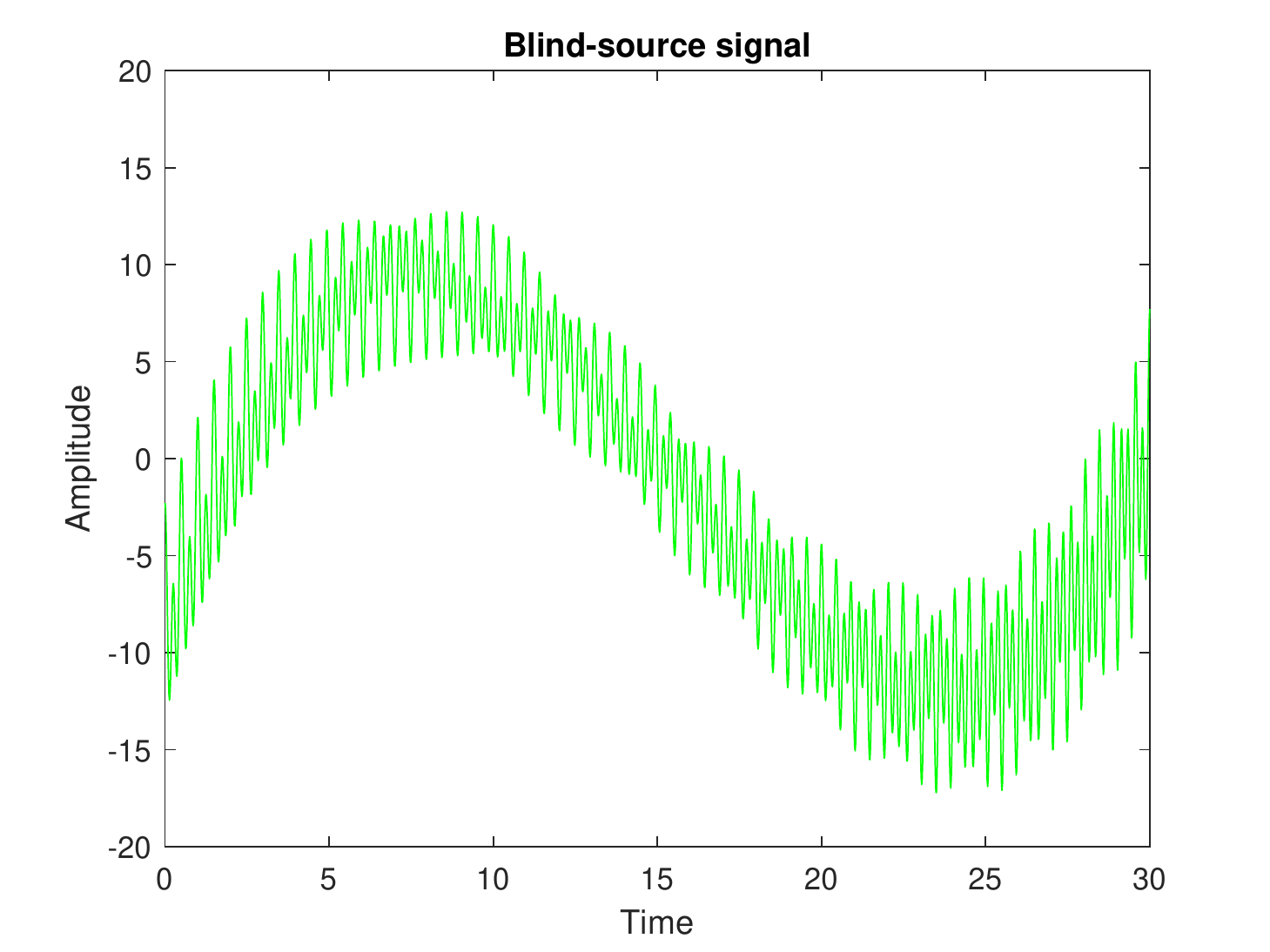}
 & \includegraphics[width=6cm]{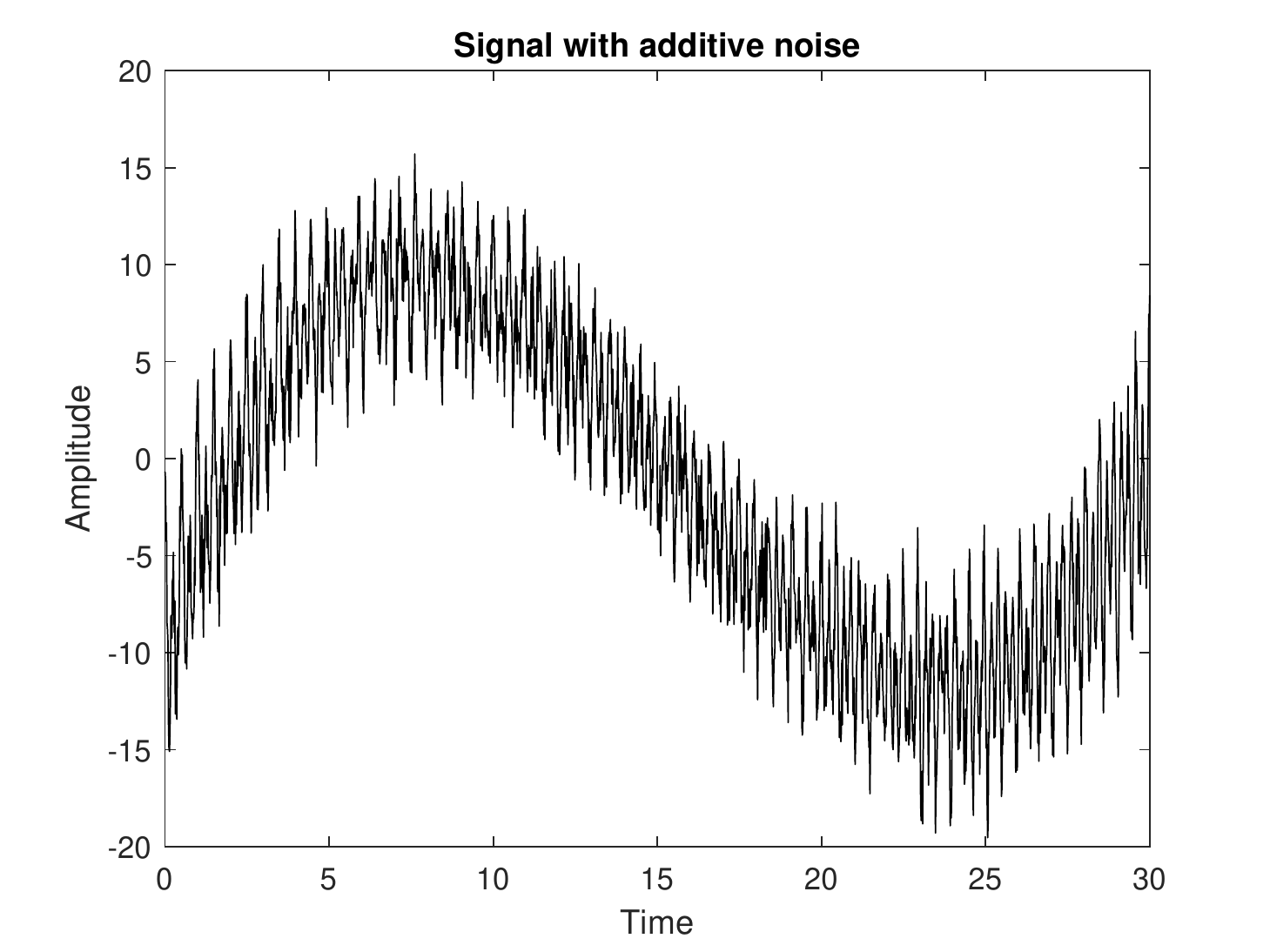}
 &\includegraphics[width=6cm]{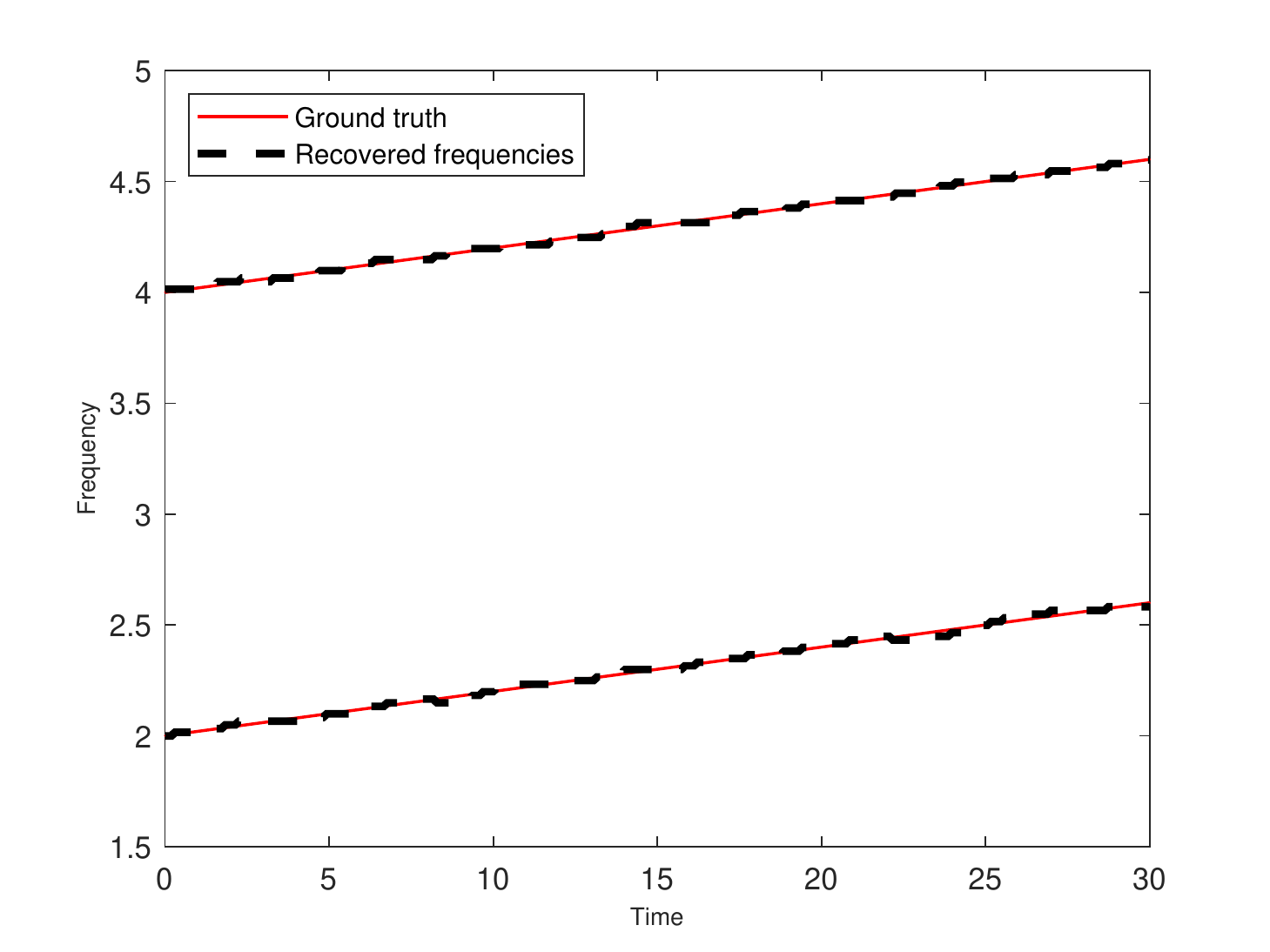}\\
\includegraphics[width=6cm]{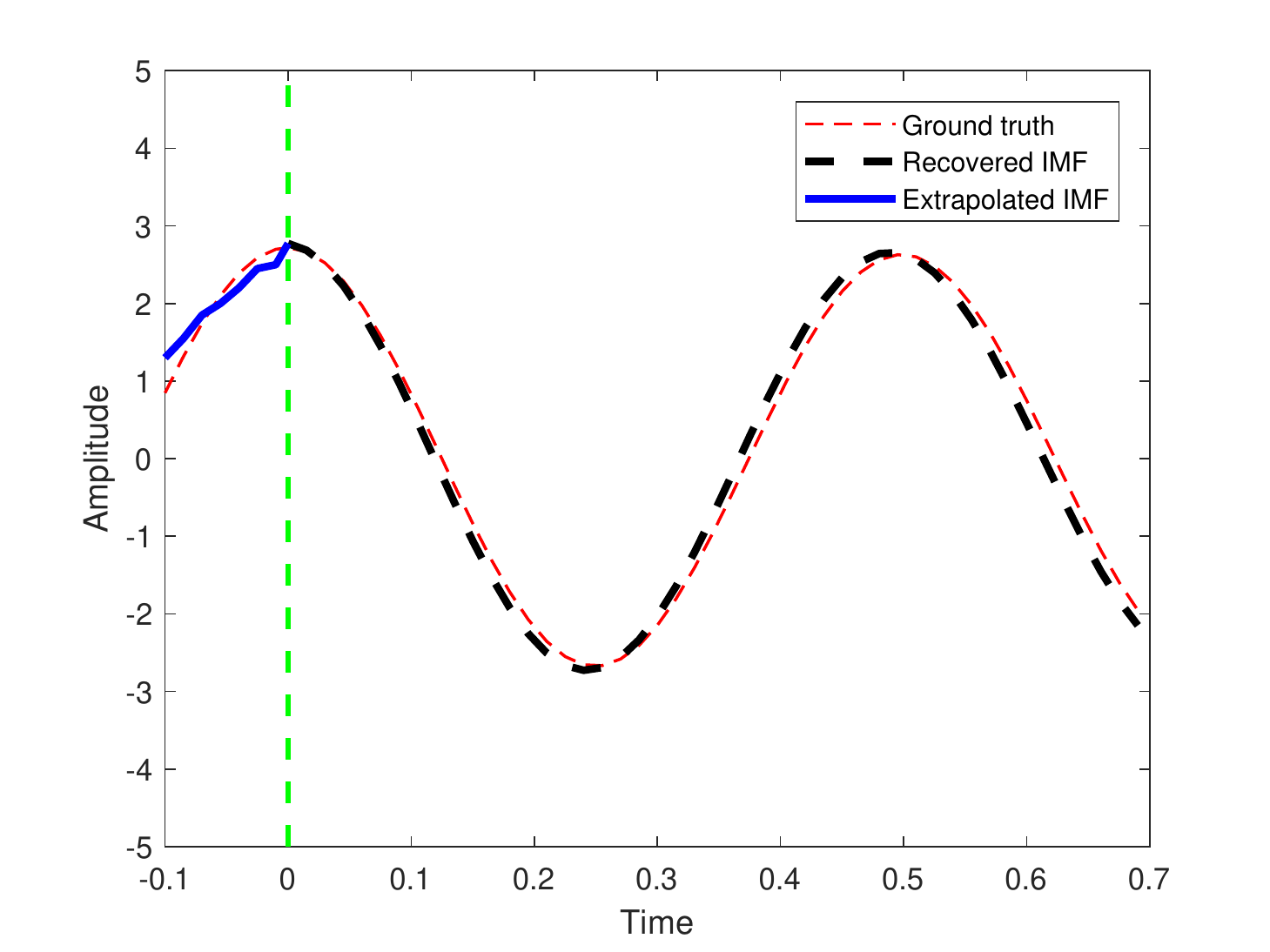}
& \includegraphics[width=6cm]{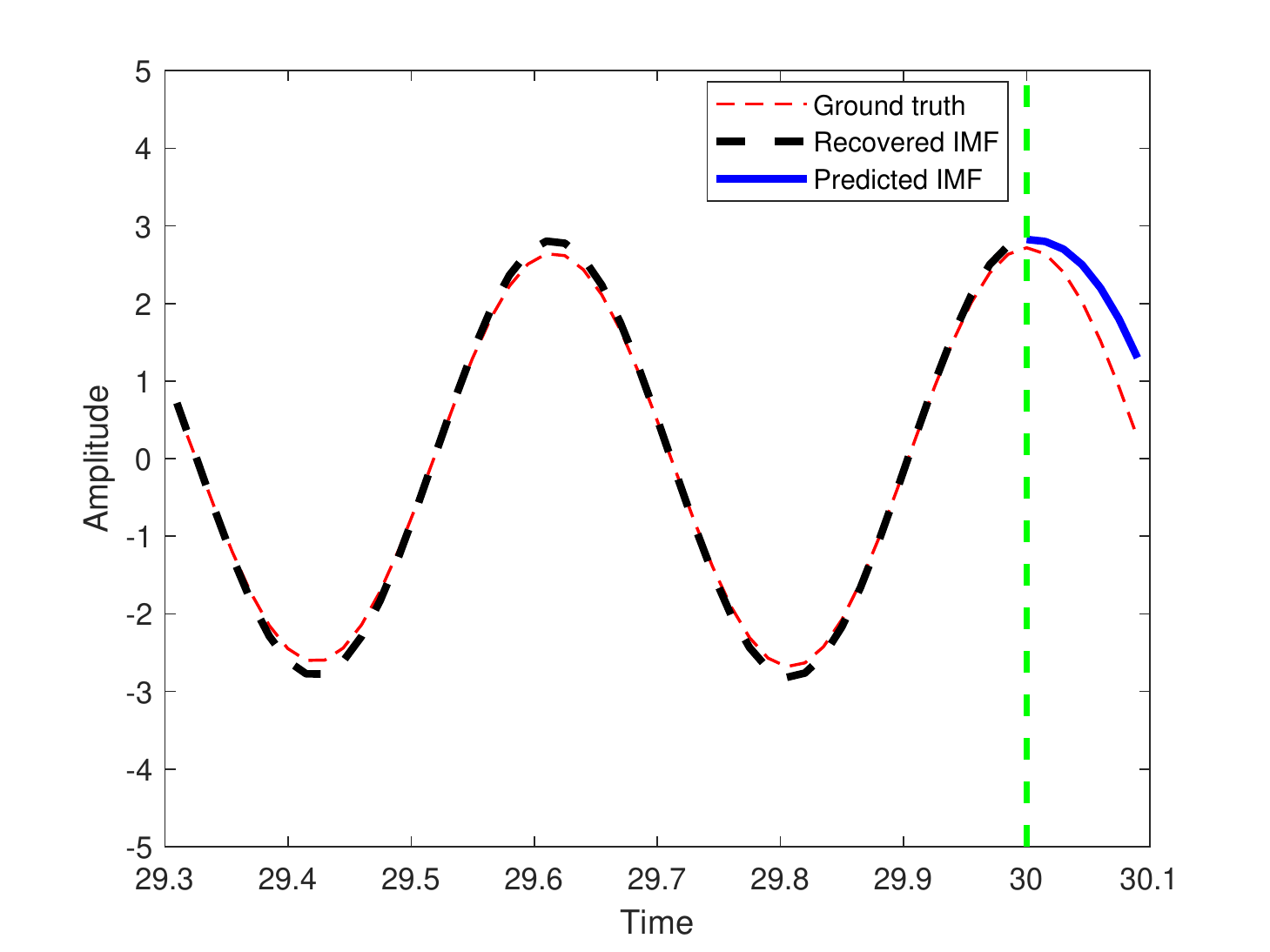}
 &\includegraphics[width=6cm]{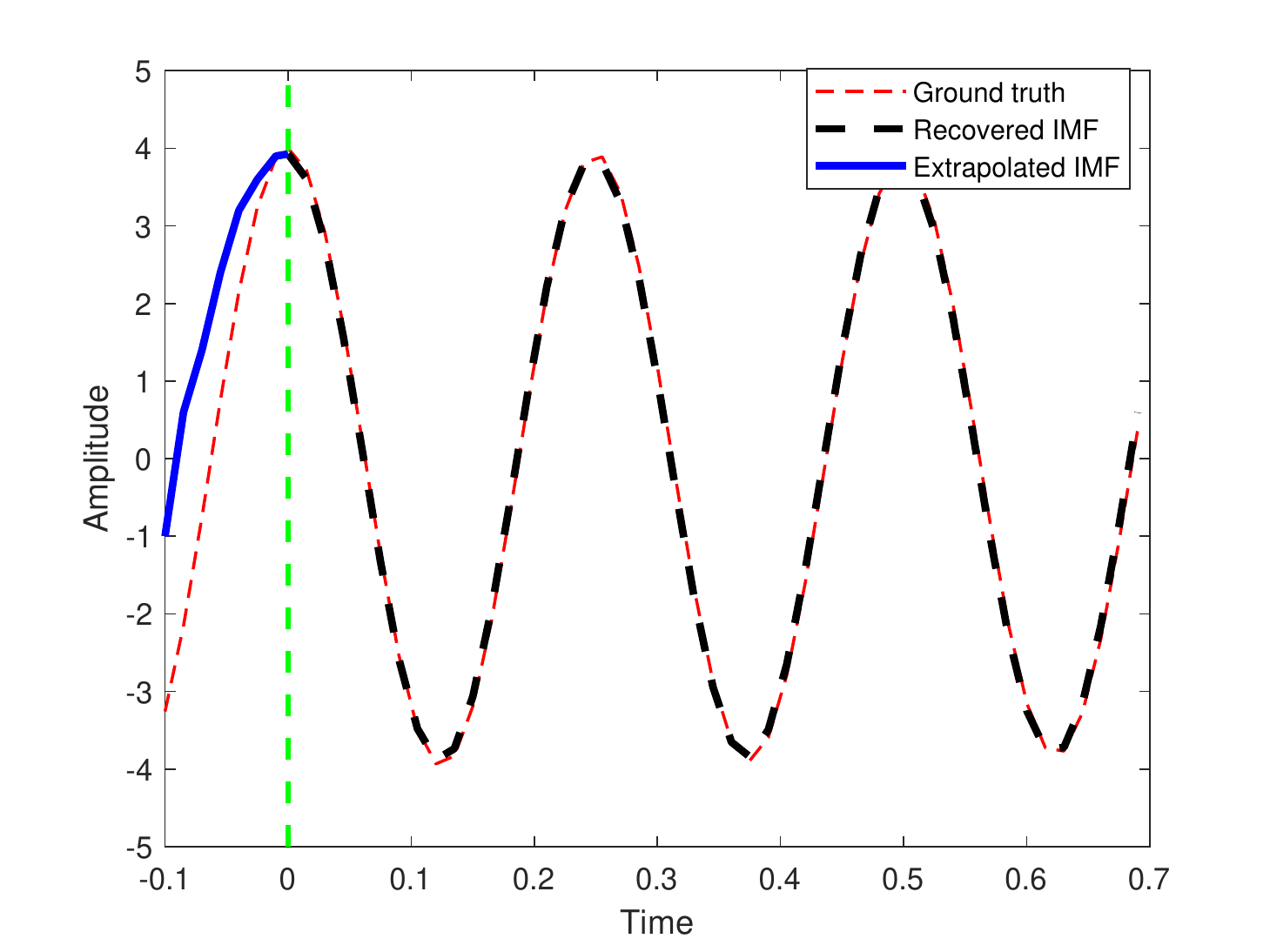}\\
   \includegraphics[width=6cm]{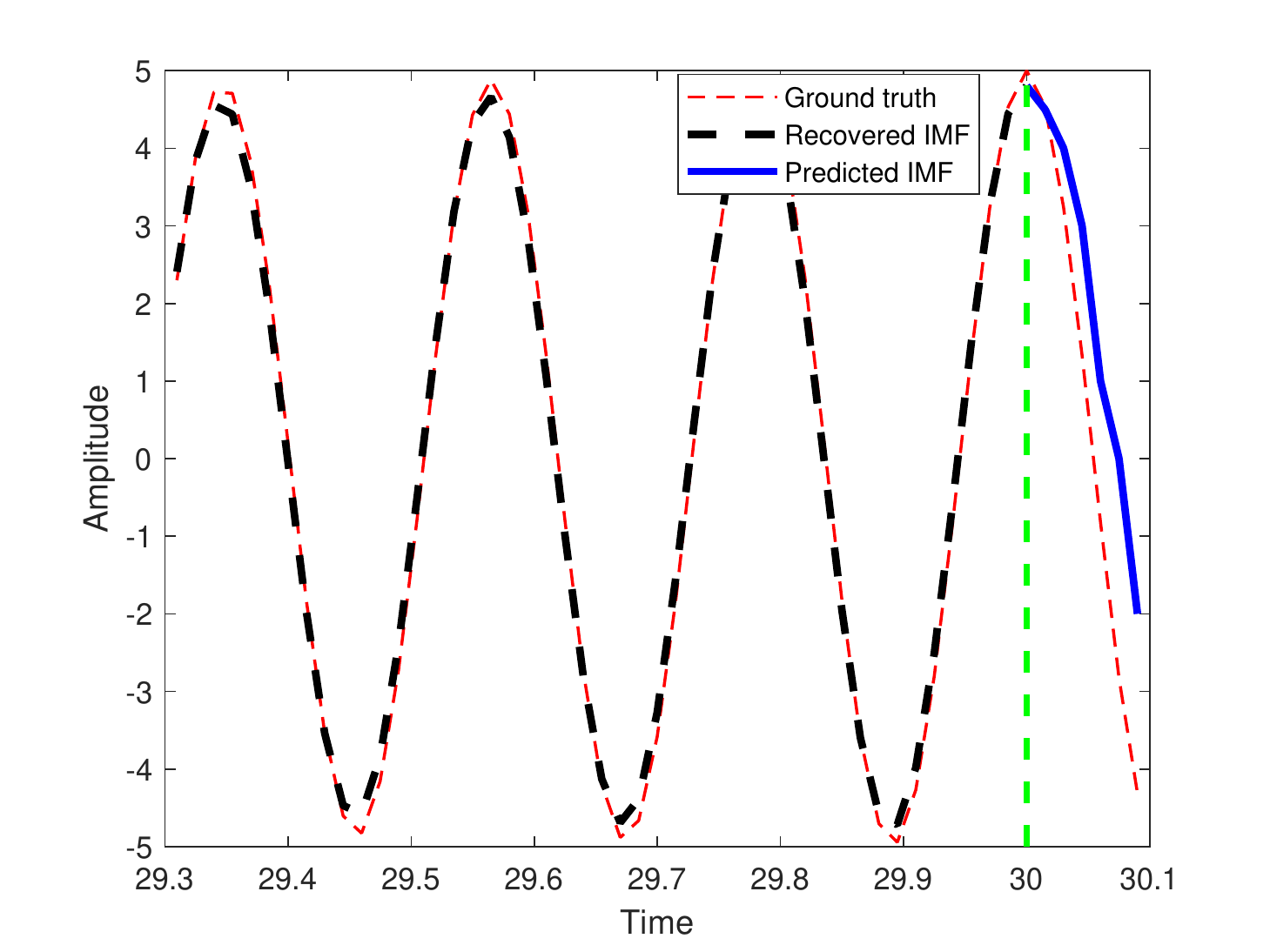}
&\includegraphics[width=6cm]{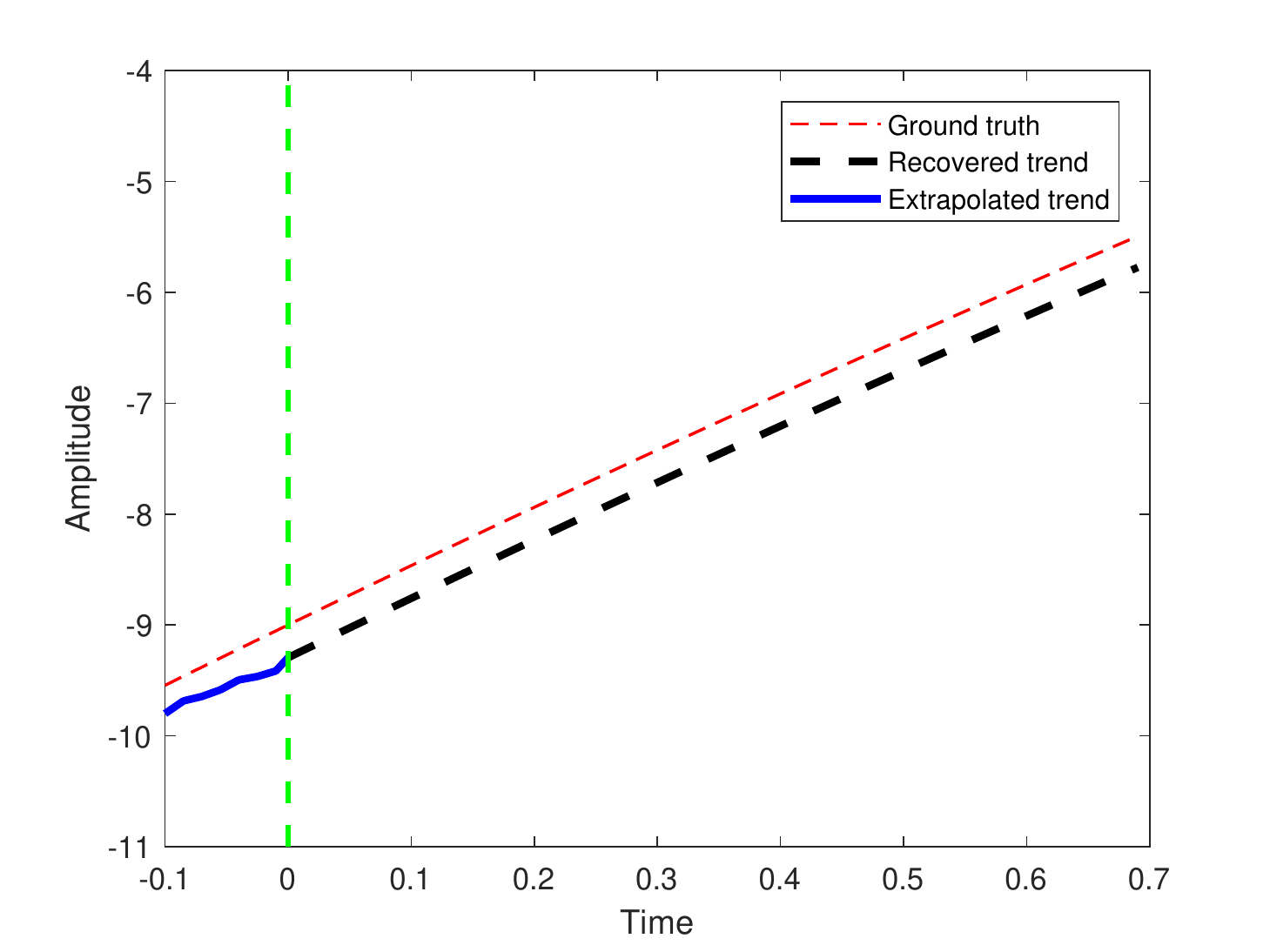}
&\includegraphics[width=6cm]{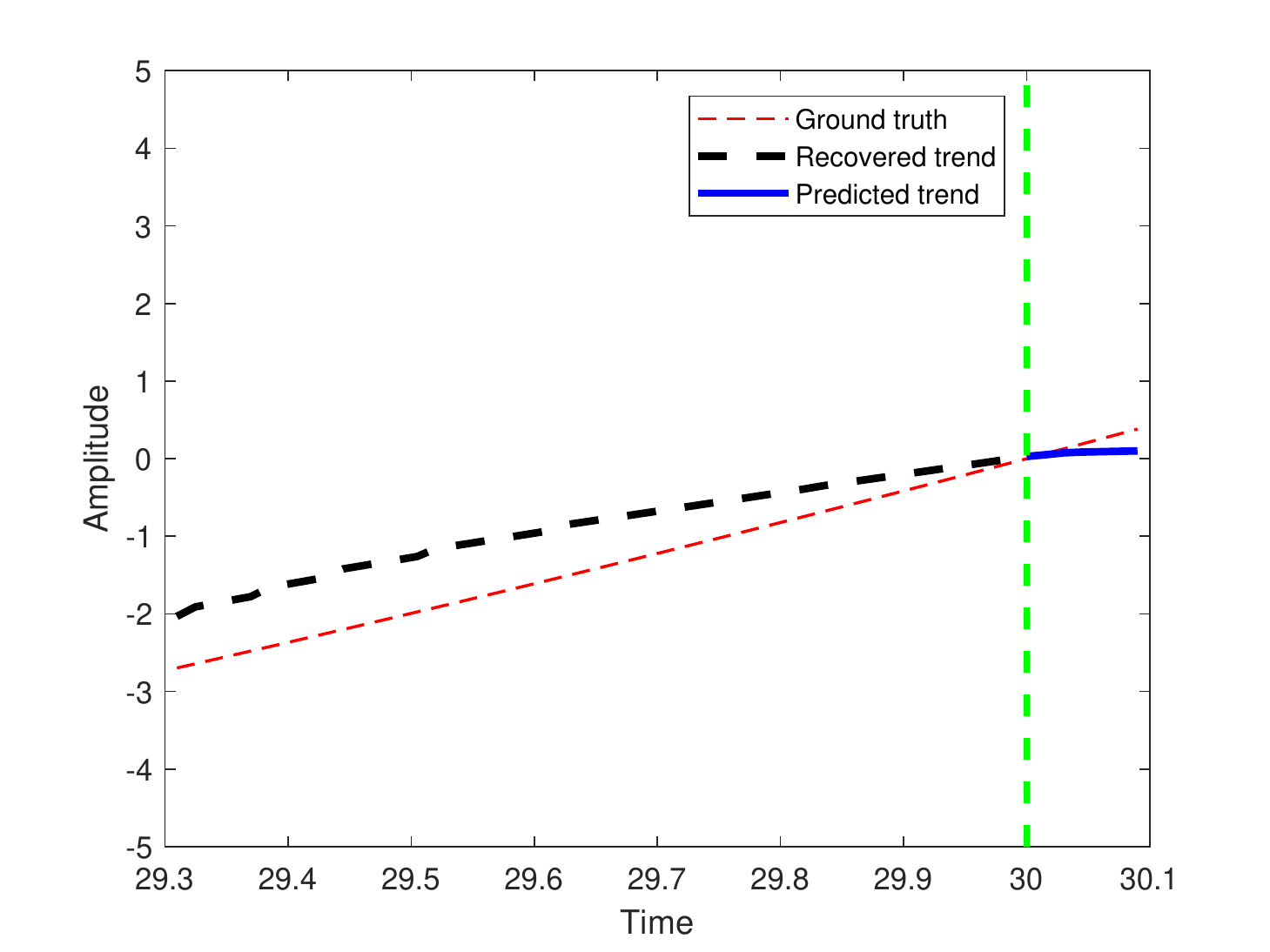}\\
\end{tabular}
\caption{Top (left to right): blind-source signal $f_{I,5}(t)$, $f_{I,5}(t)$ with additive noise, recovered frequencies. In the second and third rows, we zoom-in to the sub-intervals $[0,0.7]$ and $[29.3,30]$ and extrapolate the data outside the time domain to $[-0.1,0]$ and $[30,30.1]$, respectively. Middle (left to right): Extrapolated $1$st IMF, predicted $1$st IMF, extrapolated $2$nd IMF. Bottom (left to right): Predicted $2$nd IMF, extrapolated trend, predicted trend.}
\label{fig:extrapolation}
\end{figure}

\qed}
\end{uda}

 \begin{uda}\label{uda:batecho}
{\rm \textbf{(Bat Echo-location)}
\noindent We consider a real-word signal, namely ``a bat echolocation signal" $f_{{\rm bat}}$ \footnote{http://dsp.rice.edu/software/bat-echolocation-chirp} emitted by a large brown bat, and discover that it consists of four IMFs,
\begin{equation*}
\begin{aligned}
 f_{I,{\rm bat}}=f_{{\rm bat},1}+f_{{\rm bat},2}+f_{{\rm bat},3}+f_{{\rm bat},4}.
\end{aligned}
\end{equation*}
To convince ourselves that this decomposition makes sense, we add an unknown signal $f_{{\rm bat},5}$, given by
\begin{equation*}
\begin{aligned}
 f_{{\rm bat},5}(t)=(4/125-(7/254)\cos(2\pi t))\cos(2\pi(30t-13t^{2})),
\end{aligned}
\end{equation*}
to $f_{I,{\rm bat}}$ and decompose the combined blind source signal $f_{I,{\rm bat}}+f_{{\rm bat},5}$, and discover that it consists of five IMFs, $\widetilde{f}_{{\rm bat},1}$, $\widetilde{f}_{{\rm bat},2}$,\ldots, $\widetilde{f}_{{\rm bat},5}$. The MSEs of the five IMFs are $2.26\times10^{-4}$, $1.10\times10^{-4}$, $1.49\times10^{-4}$, $7.67\times10^{-4}$, and $1.74\times10^{-2}$, respectively, where MSE is measured by $\| f_{{\rm bat},i}-\widetilde{f}_{{\rm bat},i}\|_{2}^{2}/\|f_{{\rm bat},i}\|_{2}^{2}.$ Through the comparison of recovery, it demonstrates that the proposed computational scheme can recover the IMFs effectively. Furthermore, it enables us to identity and quantify the blind-source signal through efficient extraction of information from the individual components.

\begin{figure}[ht]
\centering
\begin{tabular}{ccc}
\includegraphics[width=6cm]{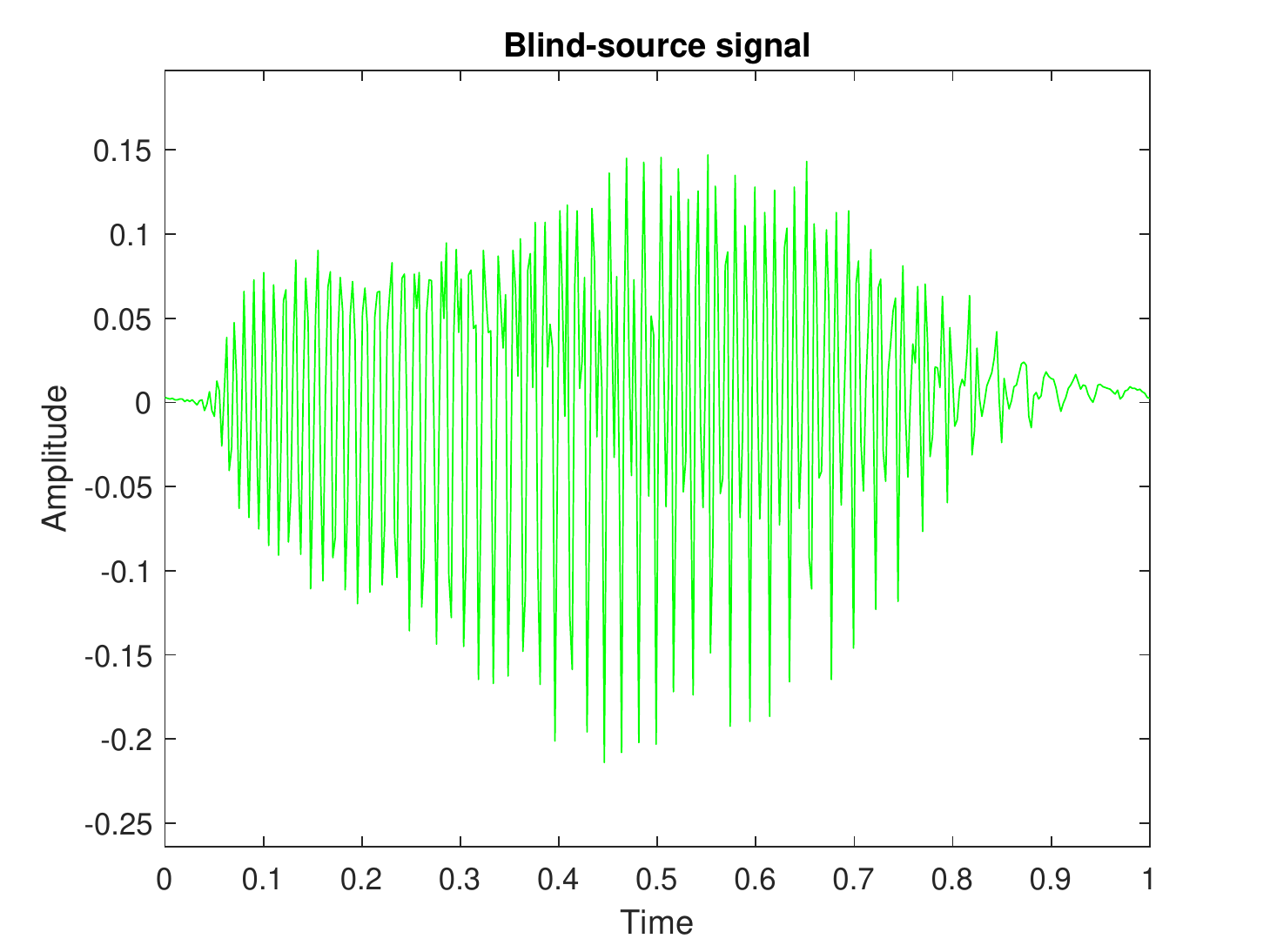}
 & \includegraphics[width=6cm]{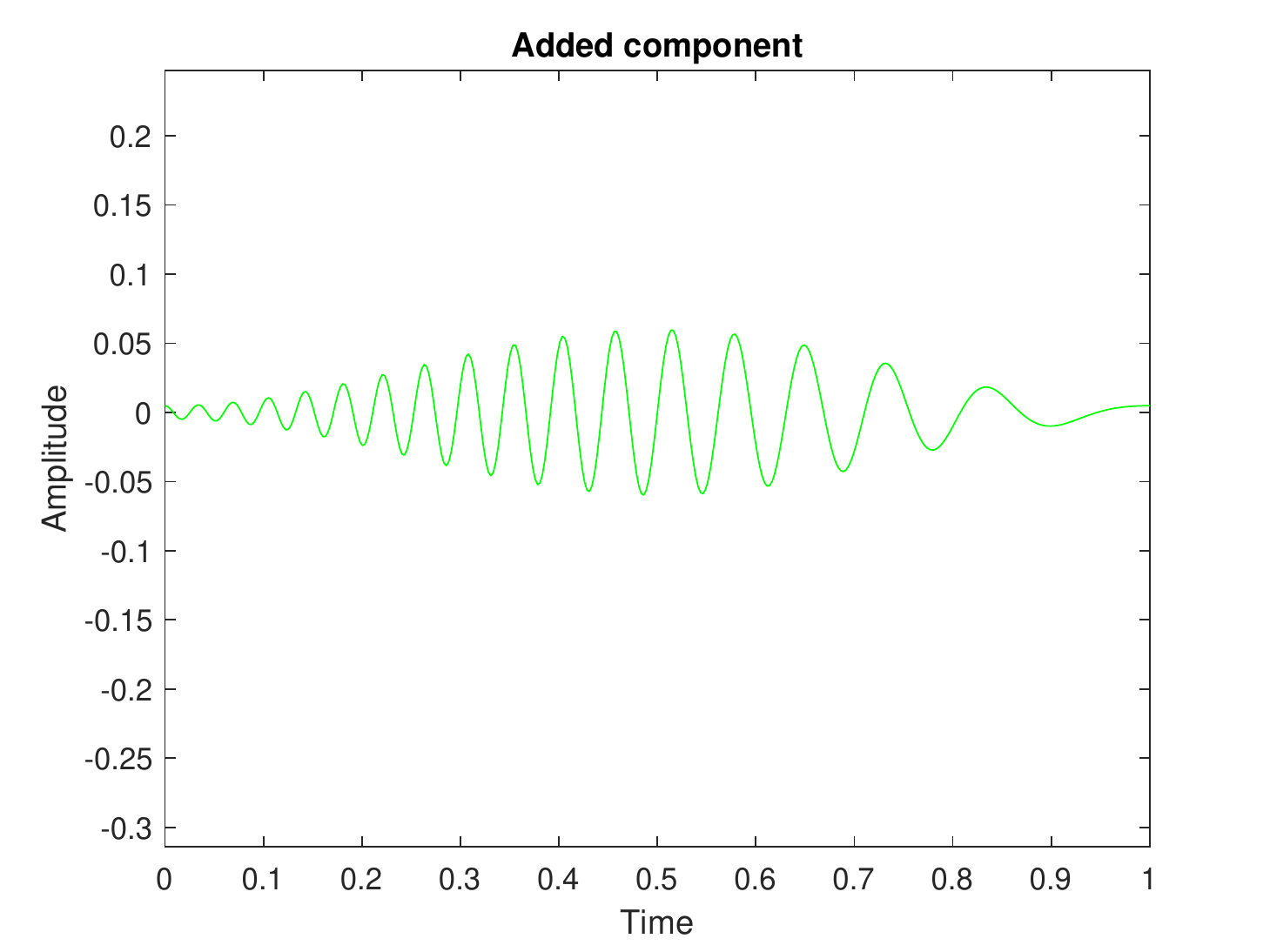}
 & \includegraphics[width=6cm]{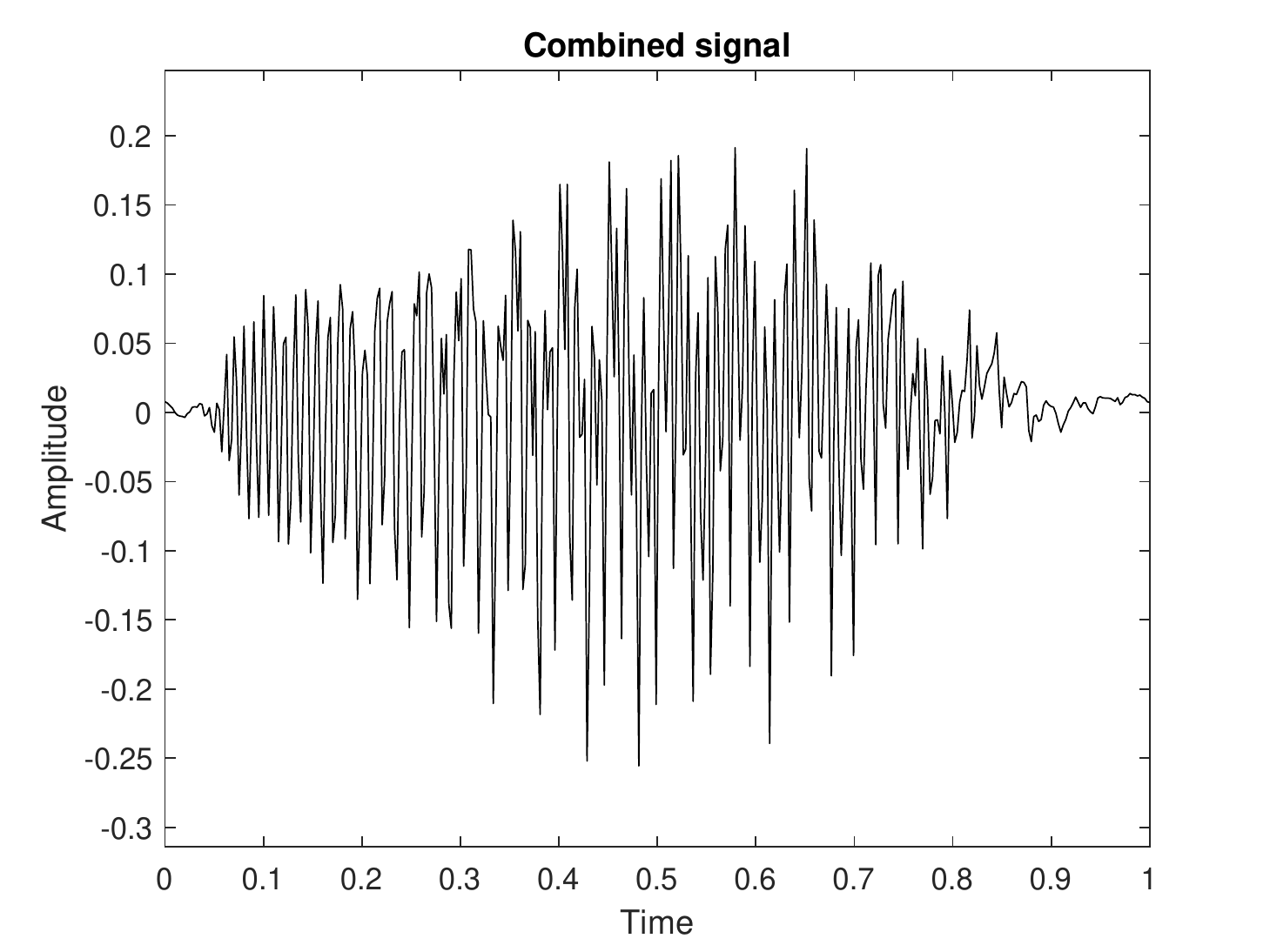} \\
\includegraphics[width=6cm]{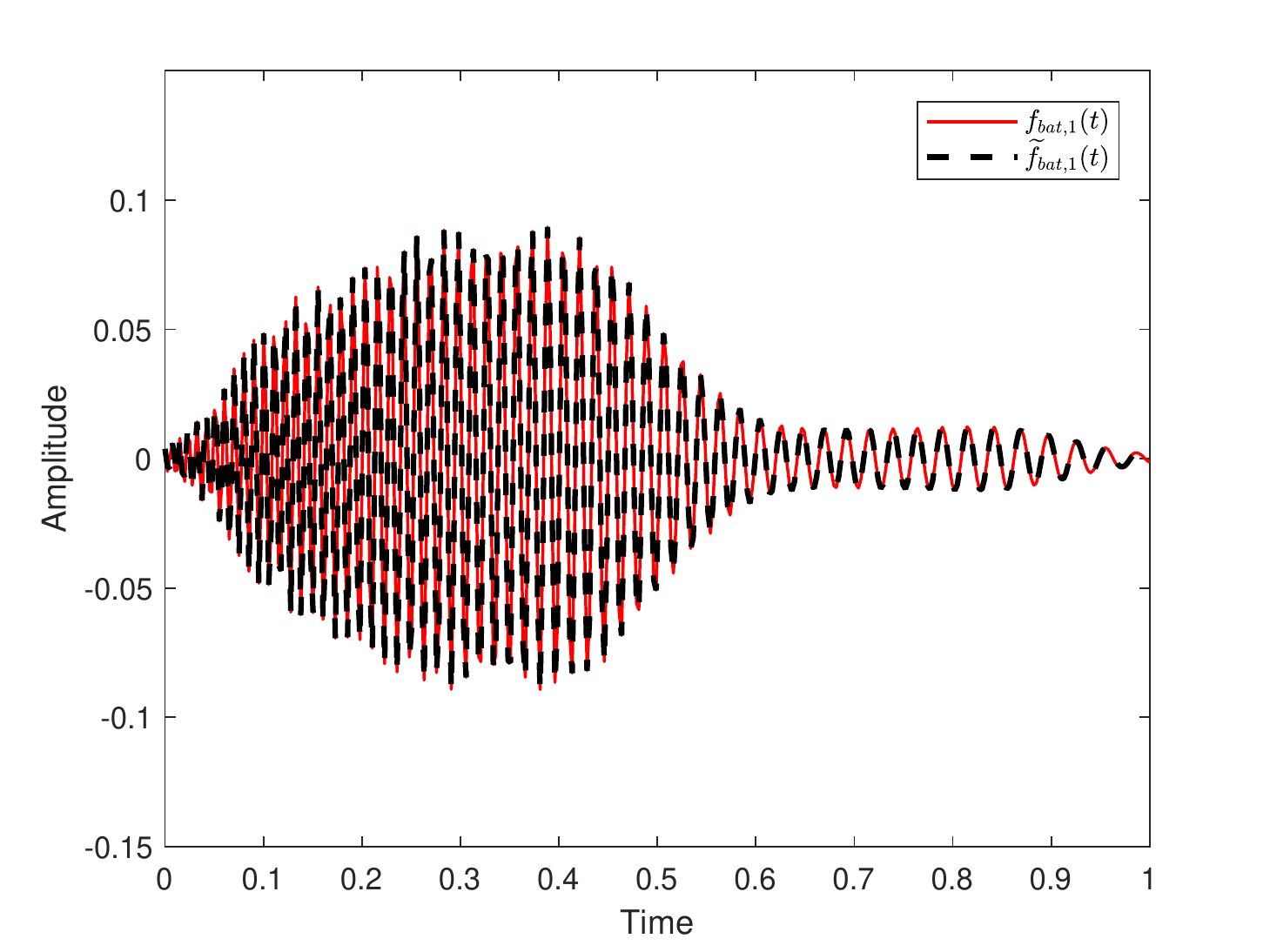}
 & \includegraphics[width=6cm]{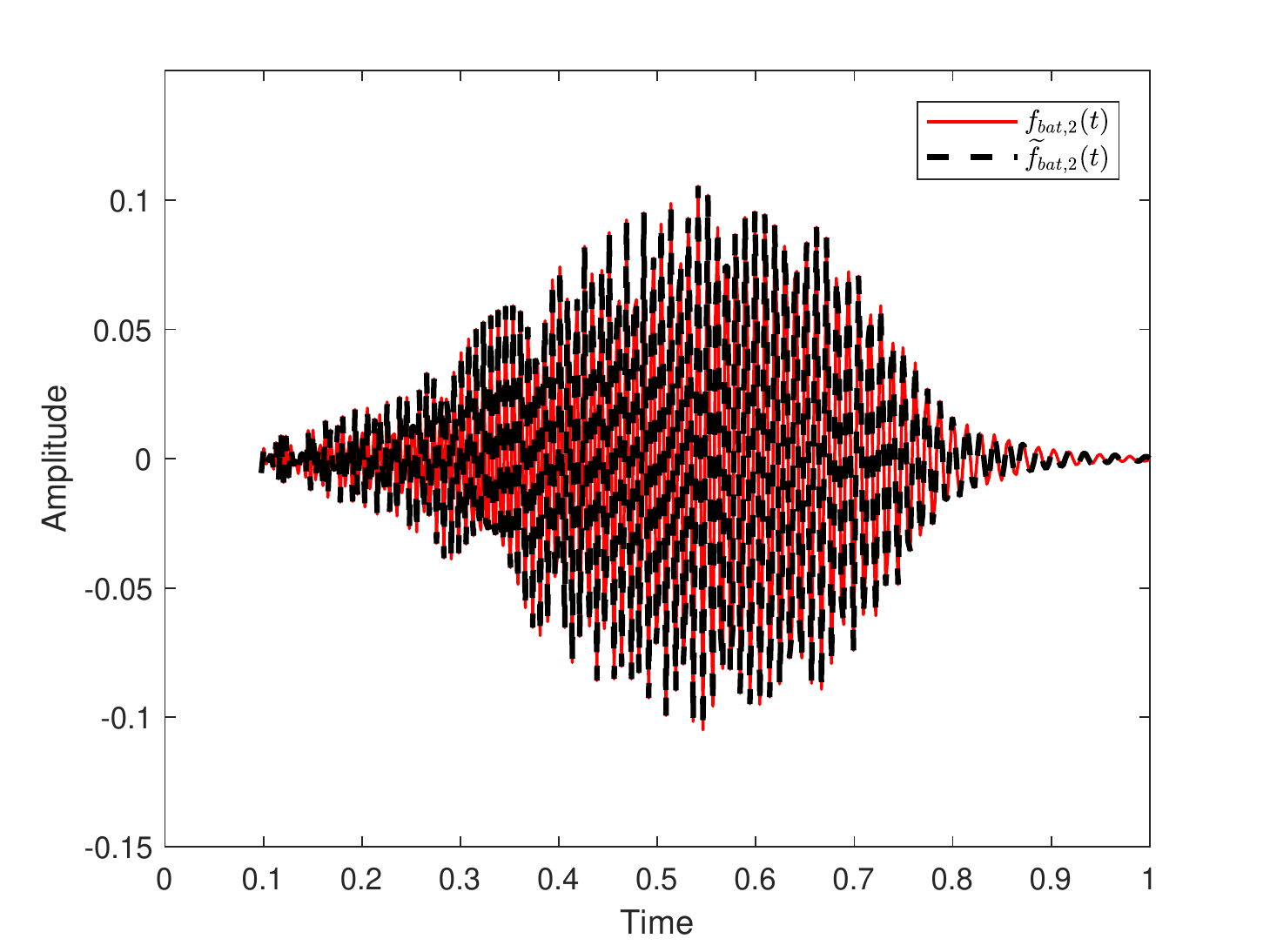}
 &\includegraphics[width=6cm]{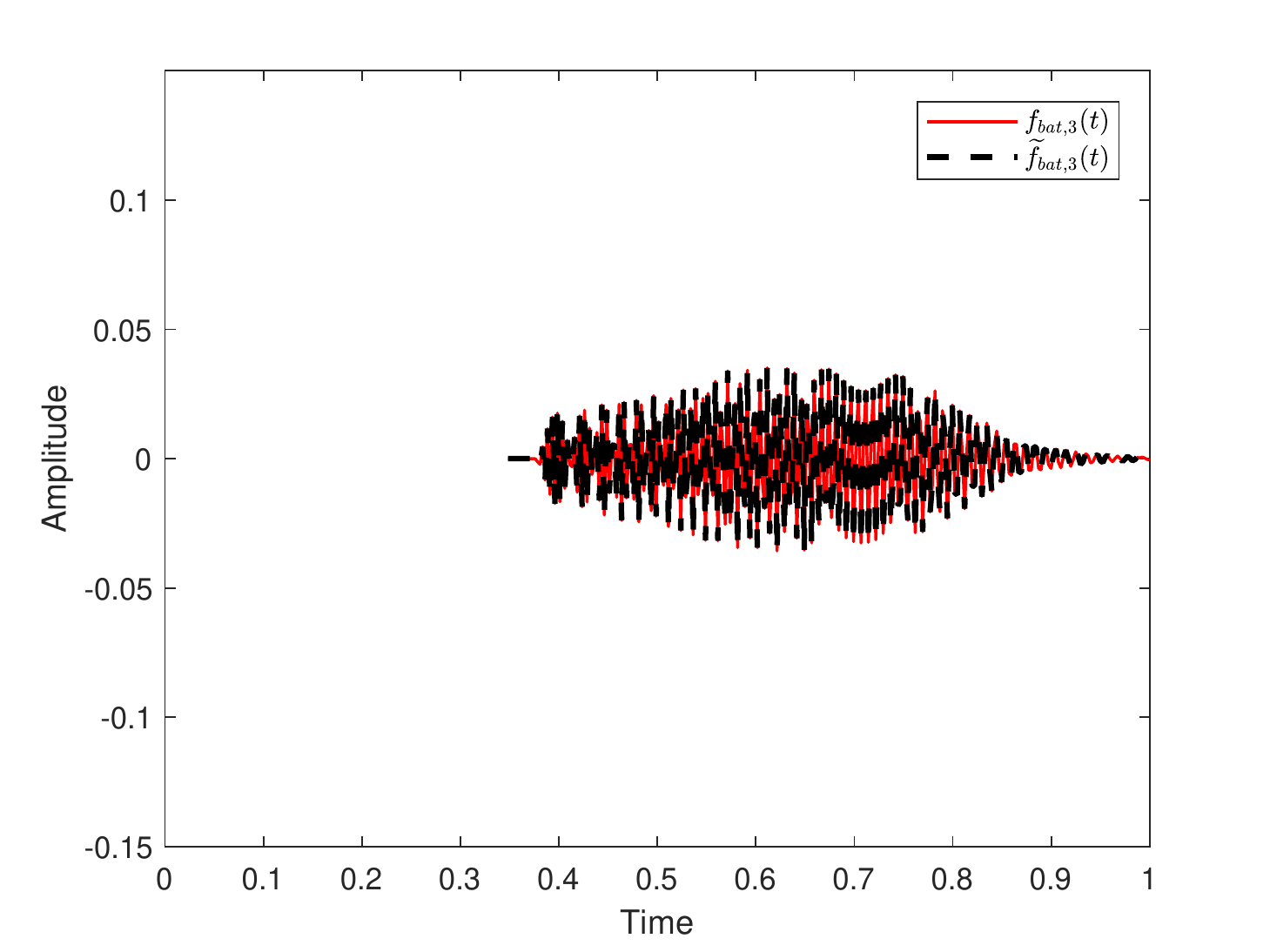} \\
 \includegraphics[width=6cm]{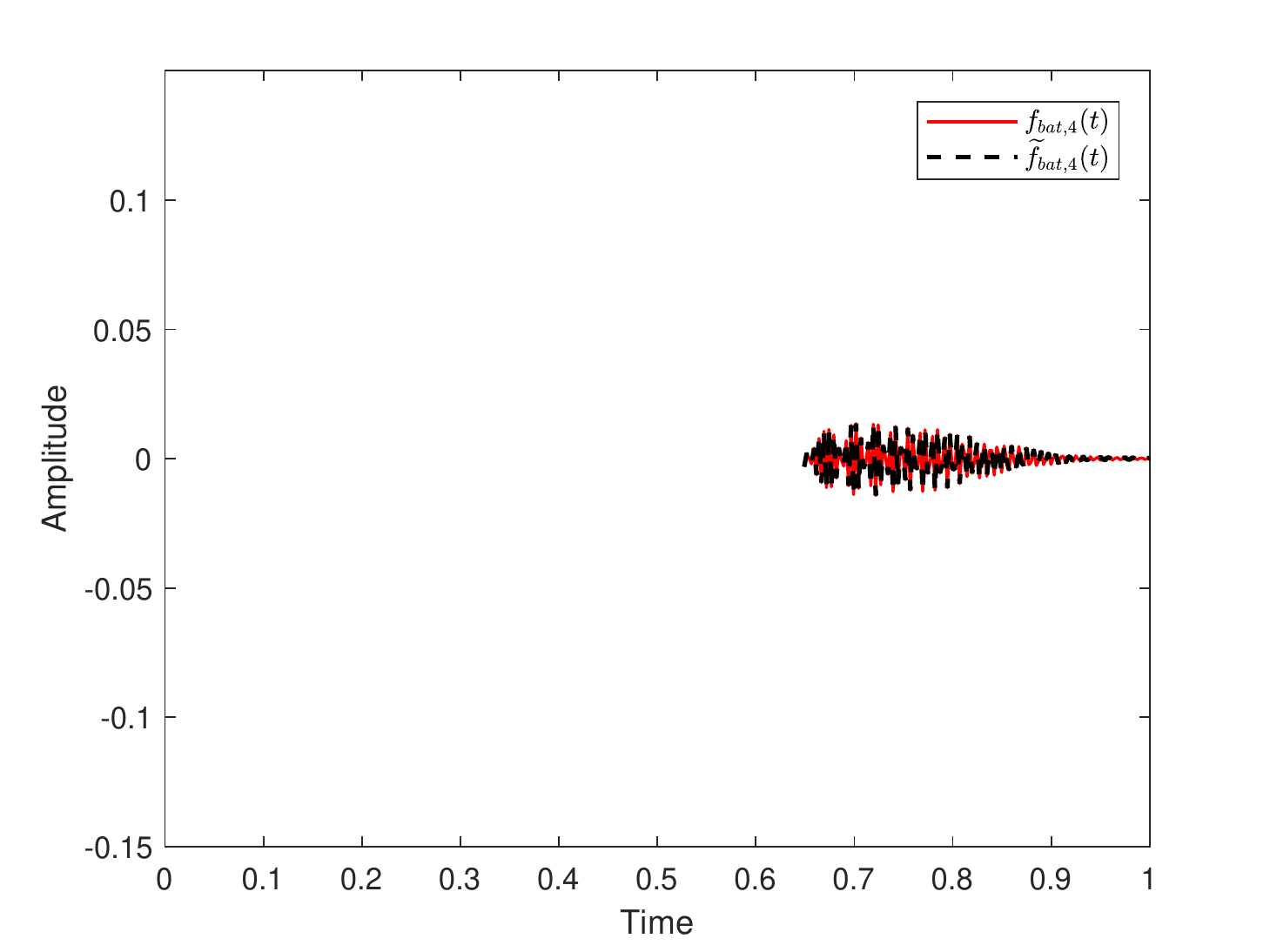}
 & \includegraphics[width=6cm]{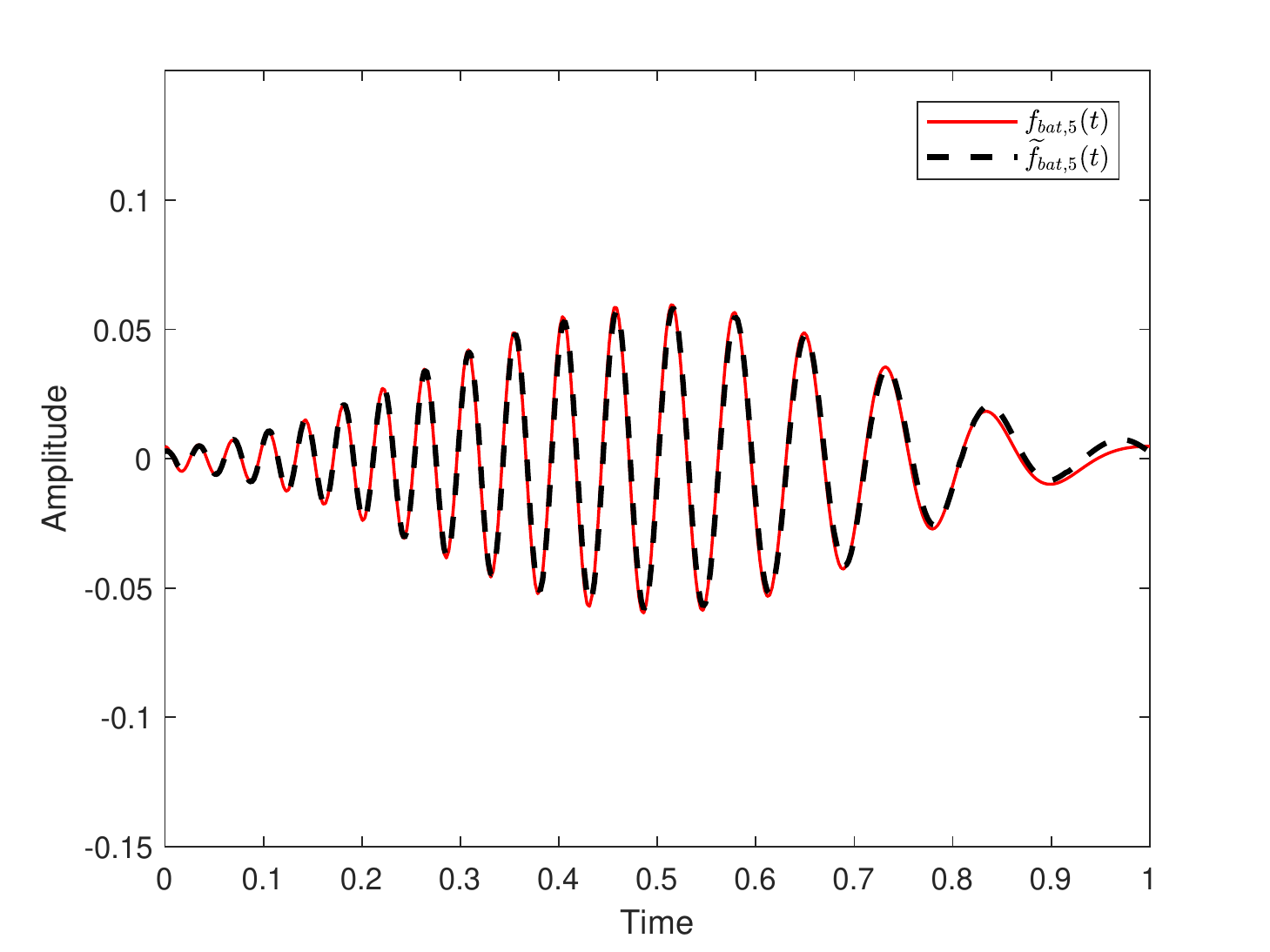} \\
\end{tabular}
\caption{Top (left to right): blind-source signal, added component $f_{bat,5}(t)$, combined signal. Middle (left to right): recovered $1$st IMF, recovered $2$nd IMF, recovered $3$rd IMF. Bottom (left to right): recovered $4$th IMF, recovered $5$th IMF.}
\label{challengefig6}
\end{figure}
\qed}
\end{uda}

\bhag{Conclusions}\label{concludesect}
In this paper, we have developed a rigorous theory, along with an effective algorithm and efficient computational scheme, for the construction of deep networks to resolving the problem of  separation of signal components and extraction of their instantaneous frequencies, based on random, non-uniform samples of a blind-source composite signal on a bounded time interval. This provides an effective solution of an important problem in signal processing, that none of the existing methods, including the popular empirical mode decomposition (EMD), synchrosqueezed wavelet transform (SST), as well as our previous signal separation operation (SSO), could work satisfactorily in general. A highlight of our computational procedure is that it precisely determines the exact number of signal components without an a priori knowledge of the composite signal. In addition, our deep networks are ``theory inspired''; i.e., constructed on the basis of a solid mathematical theory which allows the use of pre-fabricated networks that do not require training in the classical sense. Our results are proved mathematically as well as demonstrated experimentally, including an example of real-world bat echo-location signal.

\bibliographystyle{abbrv}

\begin{thebibliography}{10}

\bibitem{chen1988construction}
G.~Chen, C.~K. Chui, and M.~Lai.
\newblock Construction of real-time spline quasiinterpolation schemes.
\newblock {\em Approx. Theory Appl}, 4(4):61--75, 1988.

\bibitem{chui1988multivariate}
C.~K. Chui.
\newblock {\em Multivariate splines}.
\newblock SIAM, 1988.

\bibitem{chuiwaveletbk}
C.~K. Chui.
\newblock {\em An introduction to wavelets}.
\newblock Academic press, San Diego, 1992.

\bibitem{chuidiamond90}
C.~K. Chui and H.~Diamond.
\newblock A general framework for local interpolation.
\newblock {\em Numerische Mathematik}, 58(1):569--581, 1990.

\bibitem{chuilinwu2014}
C.~K. Chui, Y.-T. Lin, and H.-T. Wu.
\newblock Real-time dynamics acquisition from irregular samples—with
  application to anesthesia evaluation.
\newblock {\em Analysis and Applications}, 14(04):537--590, 2016.

\bibitem{bspaper}
C.~K. Chui and H.~N. Mhaskar.
\newblock Signal decomposition and analysis via extraction of frequencies.
\newblock {\em Applied and Computational Harmonic Analysis}, 40(1):97--136,
  2016.

\bibitem{chuimhasmaryke16}
C.~K. Chui, H.~N. Mhaskar, and M.~D. van~der Walt.
\newblock Data-driven atomic decomposition via frequency extraction of
  intrinsic mode functions.
\newblock {\em GEM-International Journal on Geomathematics}, 7(1):117--146,
  2016.

\bibitem{cdw}
C.~K. Chui and M.~D. van~der Walt.
\newblock Signal analysis via instantaneous frequency estimation of signal
  components.
\newblock {\em GEM-International Journal on Geomathematics}, 6(1):1--42, 2015.

\bibitem{ingrid2011}
I.~Daubechies, J.~Lu, and H.~T. Wu.
\newblock Synchrosqueezed wavelet transforms: an empirical mode
  decomposition-like tool.
\newblock {\em Applied and computational harmonic analysis}, 30(2):243--261,
  2011.

\bibitem{daubechies1996nonlinear}
I.~Daubechies and S.~Maes.
\newblock A nonlinear squeezing of the continuous wavelet transform based on
  auditory nerve models.
\newblock {\em Wavelets in medicine and biology}, pages 527--546, 1996.

\bibitem{deboorbk}
C.~{de Boor}.
\newblock {\em A practical guide to splines}.
\newblock Springer Verlag., 1978.

\bibitem{de1973spline}
C.~de~Boor and G.~Fix.
\newblock Spline approximation by quasiinterpolants.
\newblock {\em Journal of Approximation Theory}, 8(1):19--45, 1973.

\bibitem{prony_original}
B.~G.~R. De~Prony.
\newblock Essai {\'e}xperimental et analytique: sur les lois de la
  dilatabilit{\'e} de fluides {\'e}lastique et sur celles de la force expansive
  de la vapeur de l’alkool,a diff{\'e}rentes temp{\'e}ratures.
\newblock {\em Journal de l’{\'e}cole polytechnique}, 1(22):24--76, 1795.

\bibitem{huang1998empirical}
N.~E. Huang, Z.~Shen, S.~R. Long, M.~C. Wu, H.~H. Shih, Q.~Zheng, N.~Yen, C.~C.
  Tung, and H.~H. Liu.
\newblock The empirical mode decomposition and the {H}ilbert spectrum for
  nonlinear and non-stationary time series analysis.
\newblock {\em Proceedings of the Royal Society of London. Series A:
  Mathematical, Physical and Engineering Sciences}, 454(1971):903--995, 1998.

\bibitem{multilayer}
H.~N. Mhaskar.
\newblock Approximation properties of a multilayered feedforward artificial
  neural network.
\newblock {\em Advances in Computational Mathematics}, 1(1):61--80, 1993.

\bibitem{mhaskar1995degree}
H.~N. Mhaskar and C.~A. Micchelli.
\newblock Degree of approximation by neural and translation networks with a
  single hidden layer.
\newblock {\em Advances in Applied Mathematics}, 16(2):151--183, 1995.

\bibitem{wuthakur}
G.~Thakur and H.~T. Wu.
\newblock Synchrosqueezing-based recovery of instantaneous frequency from
  nonuniform samples.
\newblock {\em SIAM Journal on Mathematical Analysis}, 43(5):2078--2095, 2011.

\bibitem{maryke_thesis2015}
M.~D. van~der Walt.
\newblock {\em Wavelet Analysis of Non-stationary Signals with Applications}.
\newblock PhD thesis, University of Missouri, St. Louis, 2015.

\bibitem{hautieng_thesis2012}
H.-T. Wu.
\newblock {\em Adaptive Analysis of Complex Data Sets}.
\newblock PhD thesis, Princeton University, 2012.

\bibitem{wu2011one}
H.~T. Wu, P.~Flandrin, and I.~Daubechies.
\newblock One or two frequencies? the synchrosqueezing answers.
\newblock {\em Advances in Adaptive Data Analysis}, 3(01n02):29--39, 2011.

\end{thebibliography}

\end{document}